\documentclass[twoside,11pt]{article}

\usepackage{jmlr2e}
\usepackage{booktabs}       
\usepackage{multirow}
\usepackage{multicol}
\usepackage{hyperref}
\usepackage{amsmath}

\DeclareMathOperator*{\argmax}{argmax}   
\hypersetup{
    colorlinks=true,
    citecolor=black,
    linkcolor=blue,
    }
\usepackage{amssymb}
\usepackage{pifont}
\newcommand{\cmark}{\ding{51}}%
\newcommand{\xmark}{\ding{55}}%
\usepackage{float}
\usepackage{xcolor, soul}        
\usepackage[normalem]{ulem}
\useunder{\uline}{\ul}{}
\usepackage{wrapfig}

\jmlrheading{1}{2021}{1-48}{4/00}{10/00}{Madotto et al.}

\ShortHeadings{Few-Shot Bot}{Madotto et al.}
\firstpageno{1}

\begin{document}

\title{Few-Shot Bot: Prompt-Based Learning for Dialogue Systems}

\author{\name Andrea Madotto, \name Zhaojiang Lin, \name Genta Indra Winata, \name Pascale Fung \\
\email amadotto@connect.ust.hk \\
       Department of Electronics and Computer Engineering\\
       The Hong Kong University of Science and Technology
       }

\editor{}

\maketitle

\begin{abstract}
Learning to converse using only a few examples is a great challenge in conversational AI. The current best conversational models, which are either good chit-chatters (e.g., BlenderBot) or goal-oriented systems (e.g., MinTL), are language models (LMs) fine-tuned on large conversational datasets. Training these models is expensive, both in terms of computational resources and time, and it is hard to keep them up to date with new conversational skills. A simple yet unexplored solution is prompt-based few-shot learning~\citep{brown2020language} which does not require gradient-based fine-tuning but instead uses a few examples in the LM context as the only source of learning. In this paper, we explore prompt-based few-shot learning in dialogue tasks. We benchmark LMs of different sizes in nine response generation tasks, which include four knowledge-grounded tasks, a task-oriented generations task, three open-chat tasks, and controlled stylistic generation, and five conversational parsing tasks, which include dialogue state tracking, graph path generation, persona information extraction, document retrieval, and internet query generation. The current largest released LM (GPT-J-6B) using prompt-based few-shot learning, and thus requiring no training, achieves competitive performance to fully trained state-of-the-art models. Moreover, we propose a novel prompt-based few-shot classifier, that also does not require any fine-tuning, to select the most appropriate prompt given a dialogue history. Finally, by combining the power of prompt-based few-shot learning and a Skill Selector, we create an end-to-end chatbot named the \textbf{Few-Shot Bot} (FSB), which automatically selects the most appropriate conversational skill, queries different knowledge bases or the internet, and uses the retrieved knowledge to generate a human-like response, all using only few dialogue examples per skill. 
\end{abstract}
\begin{takehome}
Comparable results to those of the state-of-the-art fully trained dialogue model can be achieved by prompting (no gradient fine-tuning involved) large LMs with few examples, as in~\cite{brown2020language}. 
\end{takehome}

\begin{code}
\href{https://github.com/andreamad8/FSB}{https://github.com/andreamad8/FSB}
\end{code}

\newpage
\tableofcontents
\newpage

\begin{figure}[t]
    \centering
    \includegraphics[width=\linewidth]{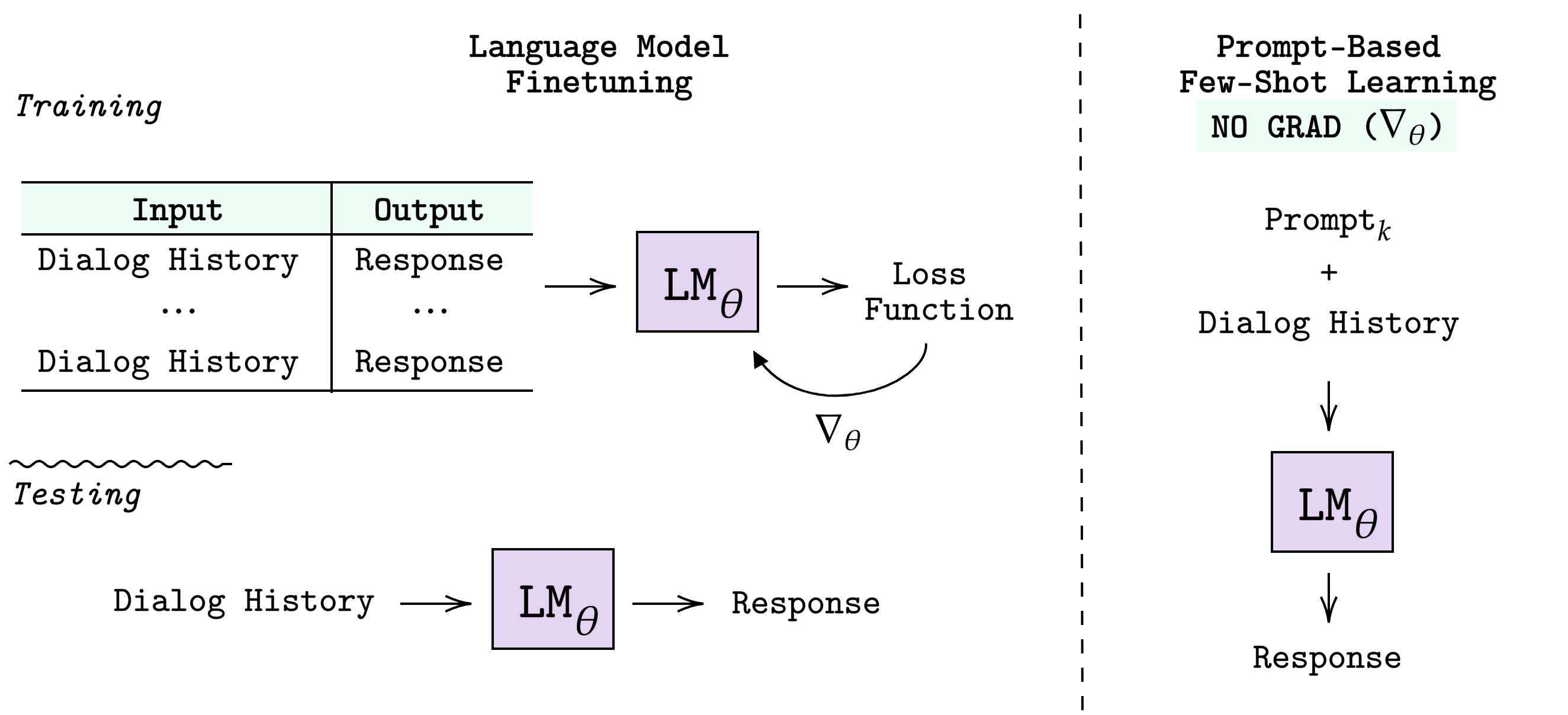}
    \caption{Fine-tuning vs. prompt-based few-shot learning.}
    \label{fig:promptvsfine}
\end{figure}
\section{Introduction}
Prompt-based few-shot learning~\citep{radford2019language,brown2020language} is a gradient-free technique for performing natural language processing (NLP) tasks using only a few examples (or demonstration) in the language model (LM) context. Prompt-based learning relies \textbf{only} on these textual prompts, provided as input for every sample or query, to teach LMs how to perform the desired task. \cite{brown2020language} showed that prompt-based few-shot learning on large LMs achieves comparable results to state-of-the-art full-shots models in a large variety of language understanding tasks (e.g., question answering, sentiment analysis). Figure~\ref{fig:promptvsfine} highlights the difference between classical training/fine-tuning schema and prompt-based learning. In conversational AI, few-shot learning is extremely important given the high annotation cost of datasets, and thus prompt-based learning is the perfect candidate for the job. Recent blogs\footnote{\url{shorturl.at/nrJL9}} and YouTube videos\footnote{\url{https://youtu.be/PqbB07n_uQ4}} show the ability of GPT3~\citep{brown2020language} to chat fluently with humans. 

\begin{figure}[t]
    \centering
    \includegraphics[width=\linewidth]{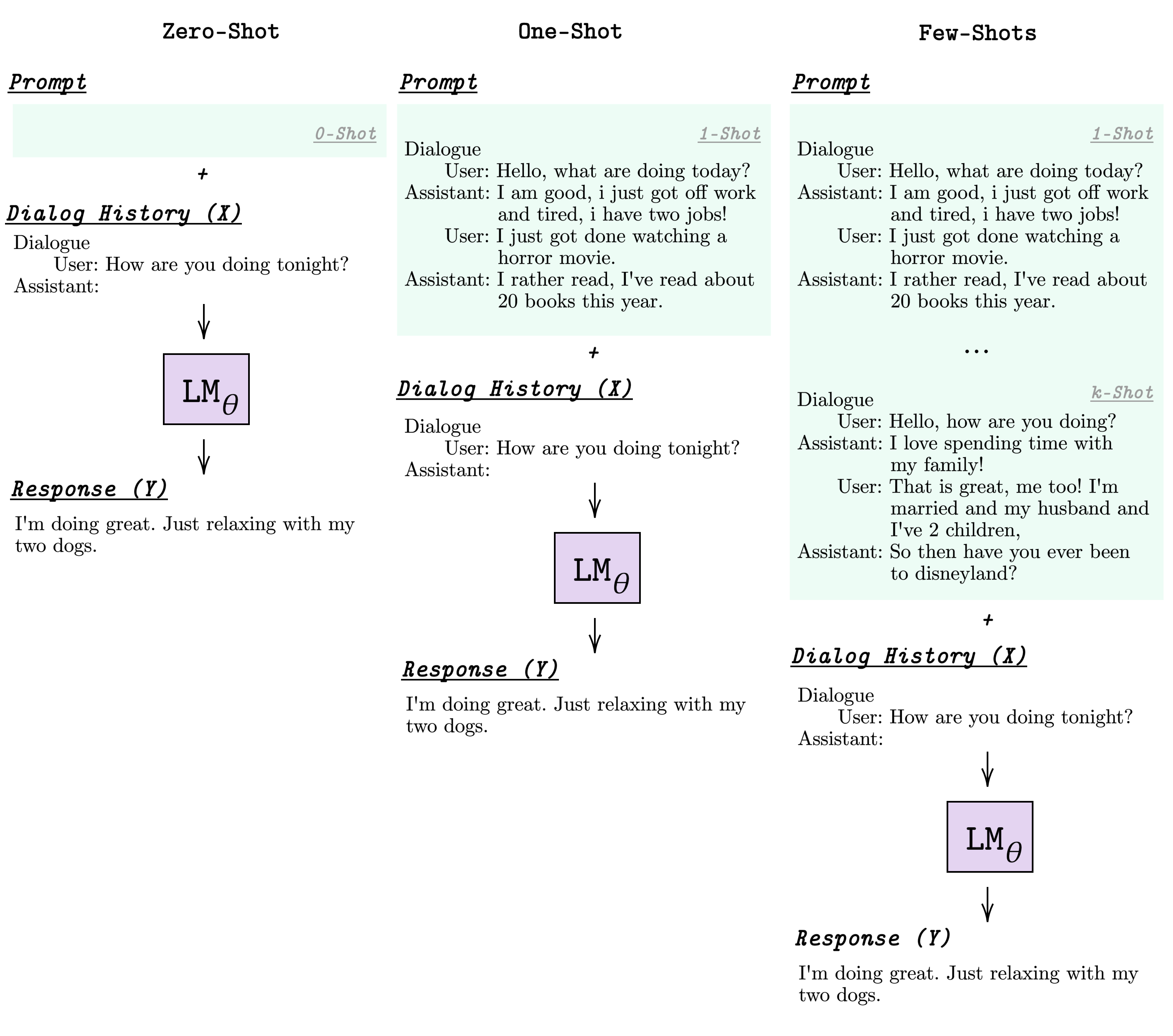}
    \caption{Example of prompt-based few-shot learning for zero, one and few-shots.}
    \label{fig:examplefewshot}
\end{figure}

To the best of our knowledge, there is no formal benchmark for prompt-based few-shot learning in conversational AI datasets. Therefore, we benchmark prompt-based few-shot learning on a large variety of dialogue-related datasets -- a total of 11 -- which include a large variety of tasks such as chit-chat, knowledge-grounded response generation, controlled stylistic generation, dialogue parsing (e.g., dialogue state tracking, document retrievers, etc).  Figure~\ref{fig:examplefewshot} showcases an example of (zero/one/few)-shot prompts in dialogue response generation. We design a prompt for each task using $k$-examples from the training set. The results in Section~\ref{sec:results}, show a promising trend in all the tasks and, in some cases, an improvement over state-of-the-art models trained on full shots.

Aside from the experimental results on each task, we propose a meta-task named \textit{Skill-Selection}, which maps dialogue histories into the most likely prompt or model that can solve a certain task. Also, in this meta-task, we leverage prompt-based few-shot learning, but instead of directly classifying the skill, which greatly limits the number of shots in the LM context, we propose a simple perplexity-based classifier (Section~\ref{sec:skillselector}). Finally, we put everything together to create the first prompt-based chat-bot, the Few-Shot Bot (FSB), which 1) implements many skills (response generation, conversational parsing, etc.) by using only few dialogue as prompt for each task, 2) automatically selects the most appropriate prompt, 3) does not require training of any kind, 4) can be easily extended with new skills by simply adding new prompts, and 5) can access the largest LMs, such as GPT-Jumbo 178B~\citep{lieber2021jurassic} via an API, or the largest released LM (GPT-J) locally using one or many GPUS. Finally, we benchmark the response safety of the FSB using the Safety Bench~\citep{dinan2021anticipating}.

To summarize, our contributions are the following: 
\begin{itemize}
    \item We are the first to benchmark prompt-based few-shot learning on a large number of conversational AI tasks, using 11 datasets with 15 tasks, such as chit-chat, knowledge-grounded (image, graph, text, and tables) response generation, DST, document retrieval, etc. Importantly, the prompt-based few-shot learning used in this paper, \textbf{do not involve any training or fine-tuning}, but uses only a few examples in the LM context, as in~\cite{brown2020language}.
    \item We propose a novel prompt-based few-shot  \textit{Skill-Selector} which can be used to select the most appropriate prompt/skill given a dialogue history. This method does not require any training, and thus it can be extended with new dialogue skills without incurring in an expensive training process.
    \item We build the first prompt-based few-shot chat-bot, the Few-Shot Bot (FSB), which can automatically select different dialogue skills (using the Skill-Selector), can access multiple knowledge sources (the Internet, Wikipedia, knowledge graphs), track user preferences, and generates human-like responses, all without fine-tuning the model, and by only using few-shots $\mathtt{Prompt}$ in the LM context. 
\end{itemize}

\section{Methodology}

Let us define a dialogue with $t$ turns $\mathrm{D}_t=\{U_1, S_1, \cdots, U_t\}$ as the alternating set of utterances between two speakers, where $U_i$ and $S_i$ are sentences made of words. Then, we define auxiliary notation for the different knowledge types used by the dialogue models. Let us define a persona description with $P$, a generic document or span of text with $K$, an image description with $D$, and a tuple with $T=\mathrm{(Sbj,Rel,Obj)}$. Depending on the dialogue task, the model is thought to map the dialogue history ($\mathrm{D}_t$) to response $S_t$ or to domain-specific languages (DSL), such as generic textual queries $Q$, knowledge graph paths $T$, and dialogue states $DS$.

\subsection{Prompt-Based Few-Shot Learning}
Prompt-based few-shot learning requires a LM, parameterized by $\theta$, which computes the likelihood of $X=\{x_0,\cdots,x_n\}$ using the chain rule of probability as
\begin{equation}
    \mathrm{P}(X) = \prod_{i=0}^{n}\mathrm{P}_{\theta}(x_i|x_{0}, \cdots, x_{i-1}).
\end{equation}
Based on this formulation, the LM generates sequences of tokens by conditioning the language model with predefined prompts. \cite{brown2020language} showed that by using input-output pairs as prompts concatenated with the sample to predict, LMs become excellent few-shot learners. Following this methodology, we define the pairs $\{(X_i,Y_i)\}_{i}^{m}$, where $(X_i,Y_i)$ is a general input-out pair, e.g., dialogue history $X=\mathrm{D}_t$ and response $Y=S_t$. Then, we define the textual concatenation of different input-output pairs as $\mathtt{Prompt}_k=\{\mathrm{Shot}_0, \cdots, \mathrm{Shot}_k\}$ where $\mathrm{Shot}_i$ is an input-output $(X,Y)$ pair converted to text using predefined templates depending on the specific task (e.g., Table~\ref{tab:persona}), and $k$ is the number of shots. Therefore, given the $\mathtt{Prompt}_k$ and the input $X$, the model generates the output sequence $Y$ by sampling from
\begin{equation}
    \mathrm{P}(Y) = \prod_{i=0}^{n}\mathrm{P}_{\theta}(y_i|y_{0}, \cdots, y_{i-1}, X, \mathtt{Prompt}_k).
\end{equation}
Figure~\ref{fig:examplefewshot} shows an example of $\mathtt{Prompt}_0$, $\mathtt{Prompt}_1$ and $\mathtt{Prompt}_k$ for a response generation task. Finally, the $\mathtt{Prompt}$ is designed based on the task. In Section~\ref{sec:results}, we describe the $\mathtt{Prompt}$ design for each of the dialogue tasks. 

\subsection{Prompt-Based Learning vs. Fine-Tuning}
The main difference between prompt-based learning and fine-tuning is the training procedure. In fine-tuning-based approaches a loss function is defined to update the weight of the model based on the training set. On the other hand, prompt-based approaches do not require a training phase and thus \textbf{no updates} to the model are performed. Figure~\ref{fig:promptvsfine} shows the key differences between prompt-based learning and fine-tuning in dialogue systems.

\subsection{Language Models}
\begin{wraptable}{r}{8cm}
\centering
\resizebox{0.49\textwidth}{!}{ 
\begin{tabular}{lrcc} \toprule
Model & n$_{\mathrm{params}}$ & n$_{\mathrm{layers}}$ & n$_{\mathrm{hidden}}$ \\ \midrule
GPT-2 & 0.1B & 12 & 768 \\
GPT-2$_{\mathrm{MEDIUM}}$ & 0.3B & 24 & 768  \\
GPT-2$_{\mathrm{LARGE}}$ & 0.8B & 36 & 1,280  \\
GPT-2$_{\mathrm{XL}}$ & 1.6B & 48 & 1,600  \\
GPT$_{\mathrm{NEO}}$ & 1.3B & 24 & 2,048  \\
GPT$_{\mathrm{NEO}}$ & 2.7B & 32 & 2,560  \\
GPT-J & 6B & 28 & 4096 \\ 
\bottomrule          
\end{tabular}
}
\caption{Language model architectures.}
\label{model_list}
\end{wraptable}
We consider only publicly available unidirectional LMs\footnote{The models are taken from \url{https://huggingface.co/}} of different sizes: GPT-2~\citep{radford2019language} of four sizes (0.1B, 0.3B, 0.8B, 1.6B),  GPT$_\mathrm{NEO}$~\citep{gpt-neo} of two sizes (1.3B, 2.7B) and 6B GPT-J~\citep{gpt-j}. Both GPT$_\mathrm{NEO}$ and GPT-J are trained on The Pile~\citep{gao2020pile}, a large collection of unsupervied text. Table~\ref{model_list} shows the number of parameters, layers, hidden units and feed-forward units of each pre-trained model.


\begin{table}[t]
\resizebox{\textwidth}{!}{%
\begin{tabular}{rcccc}
\hline
\multicolumn{5}{c}{\textbf{Response Generation}} \\ \hline
\multicolumn{1}{c|}{\textit{\textbf{Datasets}}} &
  \multicolumn{1}{c|}{\textit{\textbf{\#Dialog}}} &
  \multicolumn{1}{c|}{\textit{\textbf{\#Turn}}} &
  \multicolumn{1}{c|}{\textit{\textbf{KB}}} &
  \textit{\textbf{Metrics}} \\ \hline
\multicolumn{1}{r|}{\textit{ED}} &
  \multicolumn{1}{c|}{2542} &
  \multicolumn{1}{c|}{2.12} &
  \multicolumn{1}{c|}{-} &
  Bleu-4 (B), F1, Rouge-L (R), Perplexity (P) \\
\multicolumn{1}{r|}{\textit{DD}} &
  \multicolumn{1}{c|}{1000} &
  \multicolumn{1}{c|}{4.22} &
  \multicolumn{1}{c|}{-} &
  Bleu-4 (B), F1, Rouge-L (R), Perplexity (P) \\
\multicolumn{1}{r|}{\textit{PersonaChat}} &
  \multicolumn{1}{c|}{1000} &
  \multicolumn{1}{c|}{7.73} &
  \multicolumn{1}{c|}{Text} &
  Bleu-4 (B), F1, Rouge-L (R), Perplexity (P) \\
\multicolumn{1}{r|}{\textit{CG(IC)}} &
  \multicolumn{1}{c|}{9997} &
  \multicolumn{1}{c|}{3.00} &
  \multicolumn{1}{c|}{-} &
  Bleu-4 (B), F1, Rouge-L (R), Perplexity (P) \\
\multicolumn{1}{r|}{\textit{MSC}} &
  \multicolumn{1}{c|}{501} &
  \multicolumn{1}{c|}{5.85} &
  \multicolumn{1}{c|}{Text} &
  Bleu-4 (B), F1, Rouge-L (R), Perplexity (P) \\
\multicolumn{1}{r|}{\textit{WoW}} &
  \multicolumn{1}{c|}{965} &
  \multicolumn{1}{c|}{4.07} &
  \multicolumn{1}{c|}{Text} &
  Bleu-4 (B), F1, Rouge-L (R), Perplexity (P) \\
\multicolumn{1}{r|}{\textit{WIT}} &
  \multicolumn{1}{c|}{503} &
  \multicolumn{1}{c|}{4.93} &
  \multicolumn{1}{c|}{Text} &
  F1, KF1, Perplexity (P) \\
\multicolumn{1}{r|}{\textit{IC}} &
  \multicolumn{1}{c|}{9997} &
  \multicolumn{1}{c|}{3.00} &
  \multicolumn{1}{c|}{Image} &
  Bleu-4 (B), F1, Rouge-L (R), Perplexity (P) \\
\multicolumn{1}{r|}{\textit{SMD}} &
  \multicolumn{1}{c|}{304} &
  \multicolumn{1}{c|}{2.26} &
  \multicolumn{1}{c|}{Table} &
  Bleu, Ent-F1 \\
\multicolumn{1}{r|}{\textit{DialKG}} &
  \multicolumn{1}{c|}{1380} &
  \multicolumn{1}{c|}{3.34} &
  \multicolumn{1}{c|}{Graph} &
  FeQA, Blue, Rouge-L (R) \\ \hline
\multicolumn{5}{c}{\textbf{Conversational Parsing}} \\ \hline
\multicolumn{1}{c|}{\textit{\textbf{Dataset}}} &
  \multicolumn{1}{c|}{\textit{\textbf{\#Dialog}}} &
  \multicolumn{1}{c|}{\textit{\textbf{\#Turn}}} &
  \multicolumn{1}{c|}{\textit{\textbf{DSL}}} &
  \textit{\textbf{Metrics}} \\ \hline
\multicolumn{1}{r|}{\textit{WoW-Parse}} &
  \multicolumn{1}{c|}{3054} &
  \multicolumn{1}{c|}{2.71} &
  \multicolumn{1}{c|}{Text} &
  R-Precision, Rouge-L (R), Perplexity (P) \\
\multicolumn{1}{r|}{\textit{WiT-Parse*}} &
  \multicolumn{1}{c|}{503} &
  \multicolumn{1}{c|}{4.93} &
  \multicolumn{1}{c|}{Text} &
  Bleu-4 (B), F1, Rouge-L (R), Perplexity (P) \\
\multicolumn{1}{r|}{\textit{MSC-Parse}} &
  \multicolumn{1}{c|}{2004} &
  \multicolumn{1}{c|}{6.07} &
  \multicolumn{1}{c|}{Text} &
  Bleu-4 (B), F1, Rouge-L (R), Perplexity (P) \\
\multicolumn{1}{r|}{\textit{DialKG-Parse}} &
  \multicolumn{1}{c|}{1380} &
  \multicolumn{1}{c|}{3.34} &
  \multicolumn{1}{c|}{Graph-Path} &
  Path@1 (P@1), Tgt@1 (T@1), Perplexity (P) \\
\multicolumn{1}{r|}{\textit{MWoZ (Single)}} &
  \multicolumn{1}{c|}{223} &
  \multicolumn{1}{c|}{5.05} &
  \multicolumn{1}{c|}{Dialog State} &
  JGA, Slot-ACC, Perplexity (P) \\ \hline
\end{tabular}%
}
\caption{Datasets used to benchmark the prompt-based few-shot learning.}
\label{tab:datasetsstats}
\end{table}

\subsection{Datasets}
    In this paper, we consider two tasks: response generation and conversational parsing. In both tasks, we cover as many dialogue skills as possible by including chit-chat, task-oriented, and knowledge-graph-image-grounded dialogue datasets. To benchmark response generation, we use nine datasets: Wizard of Wikipedia (WoW)\citep{dinan2018wizard}, Wizard of Internet (WiT)~\citep{komeili2021internet}, DialKG~\citep{moon2019opendialkg}, Stanford Multidomain Dialogue (SMD)~\citep{eric2017key}, DailyDialogue (DD)~\citep{li2017dailydialog}, Persona Chat~\citep{zhang2018personalizing}, Multi-Session Chat (MSC)~\citep{xu2021beyond}, Image Chat (IC)~\citep{shuster2020image}, and Empathetic Dialogues (ED)~\citep{rashkin2019towards}. To benchmark conversational parsing, we use five datasets: Wizard of Wikipedia (WoW)~\citep{dinan2018wizard}, Wizard of the Internet (WiT)~\citep{komeili2021internet}, DialKG~\citep{moon2019opendialkg}, Multi-Session Chat (MSC)~\citep{xu2021beyond}, and MWoZ~\citep{budzianowski2018multiwoz}. Moreover, we benchmark the controlled style response generation (CG-IC) using Image Chat with its style annotation. A summary of the datasets used in our experiments, with basic statistics and evaluation metrics, is shown in Table~\ref{tab:datasetsstats}.

\subsection{Evaluation Metrics}
In this paper, we aim to benchmark the prompt-based method with the existing SOTA, and thus we use existing automatic evaluation. This is not ideal, especially in fully chit-chat systems, in which a good response could have zero overlaps with the gold response. In response generation, we use BLEU-4 (B), F1~\citep{zhang2018personalizing}, KF1~\citep{komeili2021internet}, Rouge-L (R)\footnote{\url{https://github.com/facebookresearch/ParlAI/blob/main/parlai/core/metrics.py}}, Perplexity (P), Entity-F1 (Ent-F1)~\citep{madotto2018mem2seq}, BLEU~\citep{papineni2002BLEU} and FeQA~\citep{durmus2020feqa}. In conversational parsing we use use Joint Goal Accuracy (JGA)~\citep{wu2019transferable} in DST; recall@k~\citep{moon2019opendialkg} in graph path retrieval; BLEU-4 (B), F1, Rouge-L (R) and Perplexity (P) in the MSC; F1, KF1~\citep{komeili2021internet} and Perplexity (P) in WiT; and Retrieval-Precision (RPrec)~\citep{petroni2021kilt} in WoW. Table~\ref{tab:datasetsstats} summarizes which metrics are used in each task.

\begin{figure}[t]
    \centering
    \includegraphics[width=0.7\linewidth]{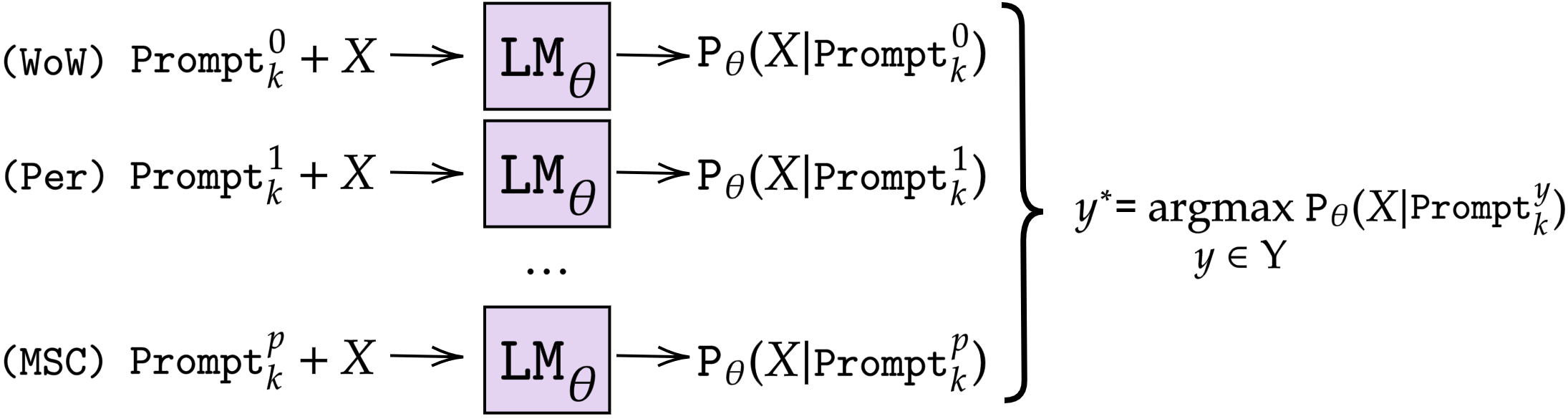}
    \caption{Prompt-based few-shot Skill-Selector. In this figure, we assume that $\mathtt{Prompt}^{0}_k$ represents the dialogues from WoW~\citep{dinan2018wizard}, $\mathtt{Prompt}^{1}_k$ from PersonaChat~\citep{zhang2018personalizing} etc.}
    \label{fig:PromptSelector}
\end{figure}

\subsection{Skill Selector}
We define the skill selection process as a classification task that maps dialogue histories into their corresponding source task or dataset name. More formally, let us define the set of labels $\mathrm{Y}=\{y_0,\cdots,y_p\}$, where $y_i$ is the task/dataset label,\footnote{This can be either the name of the dataset (e.g., WoW) or the name of the domain (e.g., navigation).} and $X$ the dialogue history.\footnote{The dialogue history never has the system or assistant turn.} Let us define the $k$-shot dataset for the label $y$ as $\mathcal{D}_y^k=\{ \{(X^i_{t^\prime},y)\}_{t^\prime} \}_i^k$, where each shot is an entire dialogue, but each dialogue's history is further split by the user turns ($X^i_{t}$). Therefore, given a collection of tasks/datasets, we define the set of all dialogues with their labels as $\mathrm{D}=\{\mathcal{D}_i^k \}_{i\in \mathrm{Y}}$, where $k$ is the number of dialogue shots. Finally, we split $\mathrm{D}$ into a training/validation/test set, and we model the conditional distribution $\mathrm{P}(Y|X)$. 

To model this conditional probability, we explore both few-shot fine-tuning and a novel prompt-based few-shot technique. Few-shot fine-tuning trains a pre-trained LM (e.g., RoBERTa~\citep{liu2019roberta}) using the samples in the training set. This is a common and efficient technique for few-shot learning,\footnote{Interested reader can refer to the code at \href{https://github.com/andreamad8/fewshotbot/tree/master/classifier}{https://github.com/andreamad8/fewshotbot/tree/master/classifier}} but it comes with two problems: 1) it requires a training procedure, which can be tricky in a low-data regime, and 2) it requires whole retraining when new classes are added to the model. To cope with these issues, we propose to leverage prompt-based few-shot learning for the skill selection task. 

\subsubsection{Prompt-Based Few-Shot Skill Selector}
Similar to the prompt-based few-shot learning for response generation, we leverage textual $\mathtt{Prompt}$s as the only source of learning. In the prompt-based few-shot Skill-Selector, we compute the conditional probability of the dialogue history with respect to all $\mathtt{Prompt}$s, and we return the label of the $\mathtt{Prompt}$ that achieves the highest probability. 

More formally, let us define the $\mathtt{Prompt}_k^y=\{\mathrm{Shot}^y_0, \cdots, \mathrm{Shot}^y_k\}$ as the $\mathtt{Prompt}$ for a given task $y$, where $\mathrm{Shot}^y_i$ is a dialogue history converted to text using predefined templates depending on the specific task ( Section~\ref{sec:skillselector}) and $k$ is the number of shots. Given a dialogue history $X$ and the set of possible labels $\mathrm{Y}=\{y_1,\cdots,y_p\}$, which index the corresponding $\mathtt{Prompt}$, we compute
\begin{equation}
       y^* = \argmax_{y\in\mathrm{Y}} \mathrm{P}_{\theta}(X|\mathtt{Prompt}^{\mathrm{y}}_k). 
\end{equation}
Figure~\ref{fig:PromptSelector} shows a high-level intuition of the prompt-based few-shot Skill-Selector, and in Section~\ref{sec:skillselector} we provide the $\mathtt{Prompt}$ details and experimental results. The main advantage of this skill selection method is the absence of a training step, which is required in normal classifiers, but it requires a linear number of LM forward for computing $\mathrm{P}_{\theta}(X|\mathtt{Prompt}^{\mathrm{y}}_k)$. This step can be expensive when the number of skills grows, but it can be easily batched.

\subsection{The Few-Shot Bot (FSB)}
The FSB combines the power of prompt-based few-shot learning and the Skill Selector, to create a chat-bot that can automatically select the most appropriate conversational skill, query different knowledge bases (KBs) or the internet, recall previous conversations with the user, and generate human-like and knowledge-grounded responses, all by using only one dialogue example per skill. The requirements of the FSB are 1) a LM, 2) a $\mathtt{Prompt}$ per skill, and 3) knowledge retrievers (e.g., Wiki, Search Engine, KG).

\paragraph{Language Models}
Based on the experimental results, we use the largest LM available, which is GPT-J (6B), for local machines, and GPT-Jurassic, accessible via API-call. GPT-J (6B) requires a 16GB VRAM GPU or four 11GB VRAM GPUs with the Parallelformers~\citep{parallelformers} library, while  GPT-Jurassic can be freely accessed (10K tokens per day) by creating an account\footnote{\href{https://www.ai21.com/studio}{https://www.ai21.com/studio}} and providing the API-KEY to the chat interface. 

\paragraph{Prompts}
In this version of the FSB, we deploy a general chat-bot with eight response generation skills, WoW, WiT, DialKG, Persona Chat, IC, DD, ED, and MSC, four conversational parsing skills, WoW-parse, WiT-parse, DialKG-parse, and MSC-parse, and a style-controllable chat-bot by using the style labels in Image Chat (CG-IC). Each of these skills is implemented with a 1-shot $\mathtt{Prompt}$ following the corresponding template (see Table~\ref{tab:sumprompt} for more information). 

\paragraph{Knowledge Retrievers}
Three of the conversational parsing skills (WoW-parse, WiT-parse, and DialKG-parse) require access to external KBs or KGs. For WoW-Parse, which generates Wikipedia title pages given a dialogue history, we use the KILT~\citep{petroni2021kilt} code base since it provides an easy to use wiki-dump and an efficient way to query Wikipedia articles by title. Once the article is retrieved, we select the first sentence of the paragraph as knowledge for the next-turn generation. For WiT-parse, which generates search queries, we leverage a simple search engine\footnote{\href{https://github.com/JulesGM/ParlAI\_SearchEngine}{https://github.com/JulesGM/ParlAI\_SearchEngine}} based on Google Search or Bing, similar to the one used in ~\cite{komeili2021internet}. As for the wiki retriever, we select the first sentence of the paragraph as knowledge for the next turn. Finally, in DialKG-parse, which generates knowledge graph paths, we use Neo4J to store the KG and py2neo for retrieving nodes, see Section~\ref{sec:dialkg_parse} for more information.

\subsection{Interactive Algorithm}
Given the LM, $\mathtt{Prompt}$ and retrievers, we define the interactive algorithm. In Listing~\ref{sec:FSB}, we describe the pseudo code of the algorithm for the FSB, while in Listing~\ref{sec:FSB_cg}, we describe the pseudo code for the style-controlled response generation. The latter is straightforward --- the user input an utterance and selects the style of the response (215 styles (e.g., kind, sweet, happy, anxious, childish, critical)). The FSB, instead, requires an external discrete memory and multiple interaction with the LM. The discrete memory is a dictionary that stores the knowledge generated by the retriever, and it is used to generate the input of the model based on the selected skill and the task template (Table~\ref{tab:sumprompt}). The interaction flow works as follow: 1) the user inputs an utterance and the dialogue history is updated, 2) the Skill Selector returns the skill to be used in the next turn, 3) depending on the skill (e.g., WoW), a query is generated with the conversational parsing skills (Line 23-24 of Listing~\ref{sec:FSB}) and the retriever execute the query, 4) the memory is updated with the retrieved knowledge and user information, and 5) the selected skill $\mathtt{Prompt}$ and the current dialogue history $\mathtt{Prompt}$\footnote{This is done using the task template in Table~\ref{tab:sumprompt} are concatenated and the content of the discrete memory.} and generate the system response. This procedure is repeated for as many turns as the user wants to interact with the FSB. 

An interesting feature of the FSB is the capability of adding/deleting/editing the different dialogue skills on the fly. Indeed, by modifying the skill $\mathtt{Prompt}$, the FSB will change its behaviour since no model training is involved. Obviously, the FSB is not a perfect and ready-to-deploy chat-bot, but it serves as a proof of concept for future chat-bots.

\begin{table}[t]
\centering
\begin{tabular}{rccccccccc}
\hline
\multicolumn{10}{c}{\textbf{Response Generation}}  \\ \hline
  \multicolumn{1}{c|}{\textit{\textbf{Datasets}}} &
  \multicolumn{1}{c|}{\textit{\textbf{Exa.}}} &
  \multicolumn{1}{c|}{\textit{\textbf{Fig.}}} &
  \multicolumn{1}{c|}{\textit{\textbf{Tab.}}} &
  \multicolumn{1}{c|}{\textit{\textbf{Sec.}}} &
  \multicolumn{1}{c|}{\textit{\textbf{Shots}}} &
  \multicolumn{1}{c|}{\textit{\textbf{MSB}}} &
  \multicolumn{1}{c|}{\textit{\textbf{LMB}}} &
  \multicolumn{1}{c|}{$\nabla \mathrm{PPL}$} &
  \multicolumn{1}{c}{$\nabla \mathrm{F1}$} \\ \hline
\multicolumn{1}{r|}{\textit{PersonaChat}}   & \multicolumn{1}{c|}{\S\ref{tab:persona}} & \multicolumn{1}{c|}{\S\ref{fig:persona_size}} & \multicolumn{1}{c|}{\S\ref{tab:results_PER_ED_WOW_IC}} & \multicolumn{1}{c|}{\S\ref{sec:persona}} & \multicolumn{1}{c|}{0,1,2,\hl{6}} &\multicolumn{1}{c|}{\cmark} &\multicolumn{1}{c|}{\cmark} &\multicolumn{1}{c|}{4.4} &\multicolumn{1}{c}{3.1} \\
\multicolumn{1}{r|}{\textit{ED}}            & \multicolumn{1}{c|}{\S\ref{tab:ED}} & \multicolumn{1}{c|}{\S\ref{fig:ed_size}} & \multicolumn{1}{c|}{\S\ref{tab:results_PER_ED_WOW_IC}} & \multicolumn{1}{c|}{\S\ref{sec:ED}} & \multicolumn{1}{c|}{0,1,7,\hl{17}} & \multicolumn{1}{c|}{\cmark} &\multicolumn{1}{c|}{\cmark} &\multicolumn{1}{c|}{3.8} &\multicolumn{1}{c}{1.5} \\
\multicolumn{1}{r|}{\textit{WoW}}           & \multicolumn{1}{c|}{\S\ref{tab:WoW}} & \multicolumn{1}{c|}{\S\ref{fig:wow_size}} & \multicolumn{1}{c|}{\S\ref{tab:results_PER_ED_WOW_IC}} & \multicolumn{1}{c|}{\S\ref{sec:wow}} & \multicolumn{1}{c|}{0,\hl{1},2,4} &\multicolumn{1}{c|}{\xmark} &\multicolumn{1}{c|}{\cmark} &\multicolumn{1}{c|}{1.1} &\multicolumn{1}{c}{14.7} \\
\multicolumn{1}{r|}{\textit{IC}}            & \multicolumn{1}{c|}{\S\ref{tab:ic}} & \multicolumn{1}{c|}{\S\ref{fig:ic_size}} & \multicolumn{1}{c|}{\S\ref{tab:results_PER_ED_WOW_IC}} & \multicolumn{1}{c|}{\S\ref{sec:IC}} & \multicolumn{1}{c|}{0,1,4,\hl{10}} &\multicolumn{1}{c|}{\cmark} &\multicolumn{1}{c|}{\cmark} &\multicolumn{1}{c|}{13.3} &\multicolumn{1}{c}{0.9} \\
\multicolumn{1}{r|}{\textit{WIT}}           & \multicolumn{1}{c|}{\S\ref{tab:wit}} & \multicolumn{1}{c|}{\S\ref{fig:wit_size}} & \multicolumn{1}{c|}{\S\ref{tab:results_WIT_CG_MSC_DD}} & \multicolumn{1}{c|}{\S\ref{sec:wit}} & \multicolumn{1}{c|}{0,\hl{1},2} &\multicolumn{1}{c|}{\xmark} &\multicolumn{1}{c|}{\xmark} &\multicolumn{1}{c|}{6.4} &\multicolumn{1}{c}{9.9} \\
\multicolumn{1}{r|}{\textit{CG(IC)}}        & \multicolumn{1}{c|}{\S\ref{tab:CGIC}} & \multicolumn{1}{c|}{\S\ref{fig:CGIC_size}} & \multicolumn{1}{c|}{\S\ref{tab:results_WIT_CG_MSC_DD}} & \multicolumn{1}{c|}{\S\ref{sec:cgic}} & \multicolumn{1}{c|}{0,1,5,\hl{10}} &\multicolumn{1}{c|}{\cmark} &\multicolumn{1}{c|}{\cmark} &\multicolumn{1}{c|}{-} &\multicolumn{1}{c}{-} \\
\multicolumn{1}{r|}{\textit{MSC}}           & \multicolumn{1}{c|}{\S\ref{tab:msc}} & \multicolumn{1}{c|}{\S\ref{fig:msc_size}} & \multicolumn{1}{c|}{\S\ref{tab:results_WIT_CG_MSC_DD}} & \multicolumn{1}{c|}{\S\ref{sec:msc}} & \multicolumn{1}{c|}{0,1,\hl{3}} &\multicolumn{1}{c|}{\cmark} &\multicolumn{1}{c|}{\cmark} &\multicolumn{1}{c|}{2.4} &\multicolumn{1}{c}{-} \\
\multicolumn{1}{r|}{\textit{DD}}            & \multicolumn{1}{c|}{\S\ref{tab:DD}} & \multicolumn{1}{c|}{\S\ref{fig:dd_size}} & \multicolumn{1}{c|}{\S\ref{tab:results_WIT_CG_MSC_DD}} & \multicolumn{1}{c|}{\S\ref{sec:DD}} & \multicolumn{1}{c|}{0,1,2,\hl{6}} &\multicolumn{1}{c|}{\cmark} &\multicolumn{1}{c|}{\cmark} &\multicolumn{1}{c|}{0.3} &\multicolumn{1}{c}{1.3} \\
\multicolumn{1}{r|}{\textit{SMD}}           & \multicolumn{1}{c|}{\S\ref{tab:smd}} & \multicolumn{1}{c|}{\S\ref{fig:smd_size}} & \multicolumn{1}{c|}{\S\ref{tab:SMD_results}} & \multicolumn{1}{c|}{\S\ref{sec:smd}} & \multicolumn{1}{c|}{0,1,\hl{8}} &\multicolumn{1}{c|}{\cmark} &\multicolumn{1}{c|}{\cmark} &\multicolumn{1}{c|}{-} &\multicolumn{1}{c}{5.4} \\
\multicolumn{1}{r|}{\textit{DialKG}}        & \multicolumn{1}{c|}{\S\ref{tab:dialKG}} & \multicolumn{1}{c|}{\S\ref{fig:dialKG_size}} & \multicolumn{1}{c|}{\S\ref{tab:dialKG_results}} & \multicolumn{1}{c|}{\S\ref{sec:dialKG}} & \multicolumn{1}{c|}{0,1,3,\hl{9}} &\multicolumn{1}{c|}{\xmark} &\multicolumn{1}{c|}{\cmark} &\multicolumn{1}{c|}{-} &\multicolumn{1}{c}{3.5$^{\dagger}$} \\ \hline
\multicolumn{10}{c}{\textbf{Conversational Parsing}}\\ \hline
\multicolumn{1}{c|}{\textit{\textbf{Dataset}}} &
  \multicolumn{1}{c|}{\textit{\textbf{Exa.}}} &
  \multicolumn{1}{c|}{\textit{\textbf{Fig.}}} &
  \multicolumn{1}{c|}{\textit{\textbf{Tab.}}} &
  \multicolumn{1}{c|}{\textit{\textbf{Sec.}}} &
  \multicolumn{1}{c|}{\textit{\textbf{Shots}}} &
  \multicolumn{1}{c|}{\textit{\textbf{MSB}}} &
  \multicolumn{1}{c|}{\textit{\textbf{LMB}}} &
  \multicolumn{2}{c}{\textit{\textbf{Results}}} \\ \hline
\multicolumn{1}{r|}{\textit{WoW-Parse}}     & \multicolumn{1}{c|}{\S\ref{tab:WoWparse}} & \multicolumn{1}{c|}{\S\ref{fig:wow_parse_size}} & \multicolumn{1}{c|}{\S\ref{tab:results_parse_WOW_WIT_MSC_MWOZ}} & \multicolumn{1}{c|}{\S\ref{sec:wowparse}} & \multicolumn{1}{c|}{0,1,\hl{5},10} &\multicolumn{1}{c|}{\cmark} &\multicolumn{1}{c|}{\cmark} &\multicolumn{2}{c}{38.4 (RPrec)} \\
\multicolumn{1}{r|}{\textit{WiT-Parse}}    & \multicolumn{1}{c|}{\S\ref{tab:witparse}} & \multicolumn{1}{c|}{\S\ref{fig:wit_parse_size}} & \multicolumn{1}{c|}{\S\ref{tab:results_parse_WOW_WIT_MSC_MWOZ}} & \multicolumn{1}{c|}{\S\ref{sec:witparse}} & \multicolumn{1}{c|}{0,1,3,\hl{4}} &\multicolumn{1}{c|}{\xmark} &\multicolumn{1}{c|}{\cmark} &\multicolumn{2}{c}{14.8 (PPL)} \\
\multicolumn{1}{r|}{\textit{MSC-Parse}}     & \multicolumn{1}{c|}{\S\ref{tab:msc_parse}} & \multicolumn{1}{c|}{\S\ref{fig:msc_parse_size}} & \multicolumn{1}{c|}{\S\ref{tab:results_parse_WOW_WIT_MSC_MWOZ}} & \multicolumn{1}{c|}{\S\ref{sec:mscparse}} & \multicolumn{1}{c|}{0,1,\hl{3}} &\multicolumn{1}{c|}{\cmark} &\multicolumn{1}{c|}{\cmark} &\multicolumn{2}{c}{19.9 (PPL)} \\
\multicolumn{1}{r|}{\textit{MWoZ-Parse}} & \multicolumn{1}{c|}{\S\ref{tab:mwozdst}} & \multicolumn{1}{c|}{\S\ref{fig:mwoz_parse_size}} & \multicolumn{1}{c|}{\S\ref{tab:results_parse_WOW_WIT_MSC_MWOZ}} & \multicolumn{1}{c|}{\S\ref{sec:mwozparse}} & \multicolumn{1}{c|}{0,1,3,\hl{5}} &\multicolumn{1}{c|}{\cmark} &\multicolumn{1}{c|}{\cmark} &\multicolumn{2}{c}{13.9 (JGA)} \\ 
\multicolumn{1}{r|}{\textit{DialKG-Parse}}  & \multicolumn{1}{c|}{\S\ref{tab:dialKG_parse}} & \multicolumn{1}{c|}{\S\ref{fig:dialkg_parse_size}} & \multicolumn{1}{c|}{\S\ref{tab:results_parse_DialKG}} & \multicolumn{1}{c|}{\S\ref{sec:dialkg_parse}} & \multicolumn{1}{c|}{1,\hl{2},5} &\multicolumn{1}{c|}{\cmark} &\multicolumn{1}{c|}{\cmark} &\multicolumn{2}{c}{6.9 (Path@1)} \\ \hline
\end{tabular}
\caption{Summary of the results in each dataset with links to the prompt example (\textit{\textbf{Exp.}}), few-shot result plots (\textit{\textbf{Fig.}}), tabular results (\textit{\textbf{Tab.}}), section with the results discussion (\textit{\textbf{Sec.}}), whether more shots achieve better results (\textit{\textbf{MSB}}), whether larger models achieve better results (\textit{\textbf{LMB}}), difference in performance between best FSB and SOTA in each task in terms of perplexity ($\nabla \mathrm{PPL}$) and F1~\citep{zhang2018personalizing} ($\nabla \mathrm{F1}$), and the best FSB performance based on different metrics  (\textit{\textbf{Results}}). Highlighted in yellow (\hl{\ \ }), the number of shots that achieves the best performance, and $\dagger$ denotes delta in performance in FeQA~\citep{durmus2020feqa} rather than F1.}\label{tab:sumresults}
\end{table}
\section{Results}\label{sec:results}
In this section, we describe the experimental setup, prompt design, and results in each dataset. For the prompt design, we use a $\mathtt{Prompt}$ template, where the content inside square brackets  is replaced with sample instances, a summary of the $\mathtt{Prompt}$ used in the experiments is shown in Table~\ref{tab:sumprompt}. In what follows, we refer to our proposed method with FSB, and we specify the size of the underlying LM explicitly in tables and figures. 

\subsection{Experiments Details}
In each dataset, we limit the number of shots based on the max input length of the LM and the number of turns and meta-data of the dialogues. As a rule of thumb, we test zero shot, one shot, and few shots, the latter of which refers to the maximum number of examples that can fit a given LM. For instance, in GPT-2~\citep{radford2019language} the max sequence length is 1024, while in GPT$_\mathrm{NEO}$~\citep{gpt-neo} and GPT-J~\citep{gpt-j} is 2048, which allows a larger number of shots in the context. In some datasets (e.g., SMD), we only include the results of GPT$_\mathrm{NEO}$ and GPT-J since GPT-2 cannot fit even one-shot. Therefore, in the experiments, we often mention that 0/1/k/2k shots are tested, referring to the fact that k shots are tested in the LM with the 1024 input length and 2k shots in the LM with the 2048 input length. 

In all the results, we report the mean, and standard deviation (STD) in the plots, performance over three runs with different $\mathtt{Prompt}$s sampled from the validation set. Importantly, we do not tune the prompt templates with a validation set, but only perform the generation on the test set. Finally, in all the generations, we use greedy decoding, maximum generation length equal to 150, \texttt{\textbackslash n}\texttt{\textbackslash n} as shot separator, and \texttt{\textbackslash n} as end-of-string token.

\subsection{Summary}
Table~\ref{tab:sumresults} shows a summary of the results in each of the datasets. In this table, we link all the examples, tables, figures, and sections in which we analyze the results in each dataset. Based on the experimental results, we observe that:
\begin{itemize}
    \item Larger LMs (in terms of parameters) hold stronger performance across all the tasks, which confirms the results reported in \cite{brown2020language}. In all the experiments, GPT-J (6B) consistently outperforms smaller models, and we observe a linear correlation between model size and performance. 
    \item In prompt-based few-shot learning, more shots do not always hold better performance. In most cases, more shots leads to better results, Table~\ref{tab:sumresults} MSH columns, but in knowledge-grounded dialogues, such as WoW, WiT, and DialKG, the one-shot experiment achieves better results than that with many shots.
    \item Conversational parsing tasks are considerably harder than response generation for prompt-based learning. For instance, in generating Wikipedia titles (WoW-Parse tasks), there is a 24.5 RPrec. gap between FSB and the current SOTA~\citep{de2020autoregressive}. Similarly, in the DST task (MWoZ-DST, single), FSB achieves only a 13.9 JGA, which is far from SOTA fully-trained models. However, in the conversational parsing tasks, we notice an even stronger correlation in performance and model size, suggesting that larger models, or better prompts, could eventually close the existing performance gap in the future. 
    \item The gap in perplexity between FSB and SOTA fully trained models is low, as shown in Table~\ref{tab:sumresults} ($\nabla \mathrm{PPL}$). This suggests that a better sampling strategy (e.g., beam-search, nucleus sampling~\citep{holtzman2019curious}, or top-K~\citep{fan2018hierarchical}) could lead to better performance. Indeed, current SOTA fully trained models in Persona Chat, ED, WoW, IC, WIT, MSC, and DD use beam-search and some other decoding strategy~\citep{shuster2020dialogue,roller2021recipes}. In this paper, we only use greedy searc,h mostly because running beam-search (e.g., beam 8) requires large GPU RAM (40GB for a 6B model half-precision). 
    \item The performance of the prompt-based few-shot Skill-Selector is comparable to the fine-tuned baselines (RoBERTa base \& large), with the advantage of being highly extendable and of not requiring any fine-tuning.
\end{itemize}

\subsection{Response Generation}

\subsubsection{Persona Chat}\label{sec:persona}
Persona chat~\citep{zhang2018personalizing} is a human-to-human multi-turn dialogue dataset in which two speakers are getting to know each other. Table~\ref{tab:persona} shows the $\mathtt{Prompt}$ template and an example from the dataset. In this dataset, the FSB generates the personalized responses by conditioning on persona information. We compare FSB (0/1/2/6 shots) with SOTA models DoDedeca Dialogue (DDD)~\citep{shuster2020dialogue}, DialoGPT~\citep{zhang2020dialogpt}, MBST~\citep{shuster2020multi}, BST~\citep{roller2021recipes}, and AdapterBot~\citep{lin2021adapter}, all of which use the whole training set (all shots). Table~\ref{tab:results_PER_ED_WOW_IC} shows the comparisons between the FSB and SOTA, and Figure~\ref{fig:persona_size} shows how the model's size and number of shots influences the metrics. Overall, in Persona Chat the 6-shot 6B model shows comparable results with those of AdapterBot, BST, and MBST. However, there is a considerable gap between FSB and DDD, which achieve a much higher Rouge-L and BLEU-4 score. 

\begin{table}[t]
\resizebox{\textwidth}{!}{%
\begin{tabular}{@{}rcccccccccccccccc@{}}
\toprule
\multicolumn{1}{l}{} &
  \multicolumn{4}{c}{\textbf{PersonaChat}} &
  \multicolumn{4}{c}{\textbf{ED}} &
  \multicolumn{4}{c}{\textbf{WoW (Seen)}} &
  \multicolumn{4}{c}{\textbf{IC}} \\ \midrule
\multicolumn{1}{l|}{} &
  \textit{\textbf{B}} &
  \textit{\textbf{F}} &
  \textit{\textbf{R}} &
  \multicolumn{1}{c|}{\textit{\textbf{P}}} &
  \textit{\textbf{B}} &
  \textit{\textbf{F}} &
  \textit{\textbf{R}} &
  \multicolumn{1}{c|}{\textit{\textbf{P}}} &
  \textit{\textbf{B}} &
  \textit{\textbf{F}} &
  \textit{\textbf{R}} &
  \multicolumn{1}{c|}{\textit{\textbf{P}}} &
  \textit{\textbf{B}} &
  \textit{\textbf{F}} &
  \textit{\textbf{R}} &
  \textit{\textbf{P}} \\ \midrule
\multicolumn{1}{r|}{\textit{DGPT}} &
  0.1 &
  11.4 &
  8.5 &
  \multicolumn{1}{c|}{-} &
  0.3 &
  10.8 &
  8.2 &
  \multicolumn{1}{c|}{-} &
  0.1 &
  8.6 &
  5.9 &
  \multicolumn{1}{c|}{-} &
  0.1 &
  6.2 &
  5.2 &
  - \\
\multicolumn{1}{r|}{\textit{DDD}} &
  \textbf{5.5} &
  \textbf{21.7} &
  \textbf{33.7} &
  \multicolumn{1}{c|}{11.2} &
  \textbf{3.7} &
  19.3 &
  \textbf{31.4} &
  \multicolumn{1}{c|}{11.4} &
  \textbf{21.0} &
  \textbf{38.4} &
  \textbf{45.4} &
  \multicolumn{1}{c|}{8.3} &
  \textbf{2.1} &
  12.9 &
  \textbf{24.6} &
  18.8 \\
\multicolumn{1}{r|}{\textit{BST}} &
  1.1 &
  18.4 &
  22.7 &
  \multicolumn{1}{c|}{\textbf{8.36}} &
  1.4 &
  19.1 &
  24.2 &
  \multicolumn{1}{c|}{\textbf{7.81}} &
  2.3 &
  18.8 &
  17.5 &
  \multicolumn{1}{c|}{\textbf{8.6}} &
  0.1 &
  9.2 &
  12.3 &
  - \\
\multicolumn{1}{r|}{\textit{MBST}} &
  1.1 &
  18.4 &
  22.6 &
  \multicolumn{1}{c|}{8.78} &
  1.5 &
  19.2 &
  24.5 &
  \multicolumn{1}{c|}{8.46} &
  2.2 &
  18.6 &
  17.4 &
  \multicolumn{1}{c|}{8.8} &
  0.4 &
  \textbf{13.1} &
  18.0 &
  \textbf{12.6} \\
\multicolumn{1}{r|}{\textit{AdptB}} &
  0.9 &
  19.7 &
  21.1 &
  \multicolumn{1}{c|}{11.1} &
  1.2 &
  \textbf{19.6} &
  21.5 &
  \multicolumn{1}{c|}{12.2} &
  9.2 &
  36.1 &
  37.6 &
  \multicolumn{1}{c|}{9.0} &
  - &
  - &
  - &
  - \\ \midrule
\multicolumn{1}{r|}{\multirow{2}{*}{\textit{\begin{tabular}[c]{@{}r@{}}FSB\\ (6B)\end{tabular}}}} &
  0.7 &
  18.7 &
  24.5 &
  \multicolumn{1}{c|}{12.6} &
  0.8 &
  18.1 &
  20.2 &
  \multicolumn{1}{c|}{11.7} &
  4.3 &
  23.7 &
  27.4 &
  \multicolumn{1}{c|}{9.5} &
  0.2 &
  12.2 &
  13.1 &
  26.0 \\
\multicolumn{1}{r|}{} &
  \multicolumn{4}{c|}{\texttt{6 shots}} &
  \multicolumn{4}{c|}{\texttt{17 shots}} &
  \multicolumn{4}{c|}{\texttt{1 shot}} &
  \multicolumn{4}{c}{\texttt{10 shots}} \\ \bottomrule
\end{tabular}%
}
\caption{Results on the PersonaChat, ED, WoW, and IC datasets.}
\label{tab:results_PER_ED_WOW_IC}
\end{table}
\subsubsection{Empathetic Dialogues (ED)}\label{sec:ED}
ED~\citep{rashkin2019towards} is a human-to-human multi-turn dataset for empathetic conversations. Table~\ref{tab:ED} shows the $\mathtt{Prompt}$ template and an example from the dataset. In this dataset, the model generates the response of the system turns only ($\mathtt{Empath}$). We compare FSB (0/1/7/17 shots) with existing SOTA models DDD~\citep{shuster2020dialogue}, DialoGPT\citep{zhang2020dialogpt}, MBST~\citep{shuster2020multi}, BST\citep{roller2021recipes}, and AdapterBot~\citep{lin2021adapter}, all of which use the whole training set (all shots). Table~\ref{tab:results_PER_ED_WOW_IC} shows the main results, and Figure~\ref{fig:ed_size} shows how the model's size and number of shots influences the metrics. Overall, in ED, the 6B model shows comparable results with those of AdapterBot, BST, and MBST especially when a larger number of shots are used (17 shots). However, there is a considerable gap between the FSB and DDD, which achieves a much higher Rouge-L and BLEU-4 score. This might be due to the sampling strategy used in DDD (beam-search), as reported in \cite{shuster2020multi}.

\subsubsection{Wizard of Wikipedia (WoW)}\label{sec:wow}
WoW~\citep{dinan2018wizard} is an open-domain conversational dataset grounded on Wikipedia knowledge. Table \ref{tab:WoW} shows the $\mathtt{Prompt}$ template and an example from the dataset. In this dataset, the FSB generates knowledge-grounded responses given the gold Wikipedia sentences (KB) provided at each turn of the dialogue. In our experiments, we keep the gold sentence provided, and we evaluate on the topic-seen test set. We compare FSB (0/1/2/4 shots) with SOTA models DDD~\citep{shuster2020dialogue}, DialoGPT~\citep{zhang2020dialogpt}, MBST~\citep{shuster2020multi}, BST~\citep{roller2021recipes}, and AdapterBot~\citep{lin2021adapter}, all of which use the whole training set (all shots). In Figure \ref{fig:wow_size}, we observe that the one-shot 6B model obtains the best performance and that by increasing the number of shots the performance constantly decreases. In Table~\ref{tab:results_PER_ED_WOW_IC}, the FSB achieves better Rouge-L and BLEU-4 scores when compared to BST and MBST, which, however, use knowledge retrieved from Wikipedia rather than gold knowledge. However, when compared to DDD and AdapterBot, which also use gold Wikipedia knowledge, the FSB achieves a much lower Rouge-L and BLEU-4 score. 

\begin{figure}[t]
    \centering
    \includegraphics[width=0.8\linewidth]{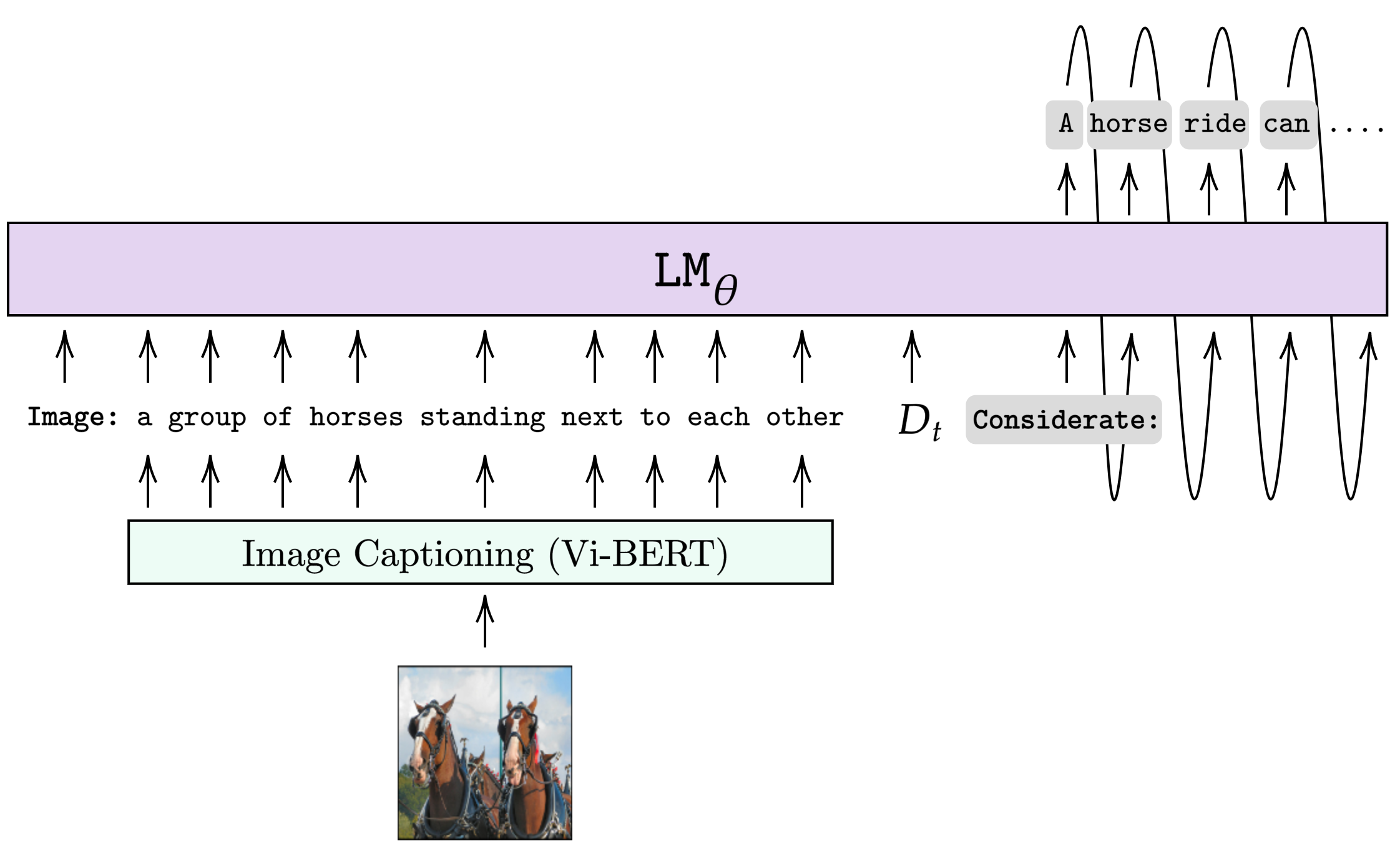}
    \caption{Image Chat input pipeline. The image is passed through an image-captioning model to generate the conversational context. $D_t$ denotes the dialogue history at turn $t$, and "Considerate" is the style of the generated response. }
    \label{fig:IC_example_LM}
\end{figure}
\subsubsection{Image Chat (IC)}\label{sec:IC}
IC~\citep{shuster2020image} is an open-domain conversational dataset grounded on images. Table \ref{tab:ic} shows the $\mathtt{Prompt}$ template and an example from the dataset. The dataset uses randomly selected images from the YFCC100M dataset~\citep{thomee2016yfcc100m}. Existing LMs do not accept images as input. Thus, we use a pre-trained image-captioning model (ViBERT~\cite{lu2019vilbert}\footnote{Code: \url{https://github.com/ruotianluo/ImageCaptioning.pytorch/blob/master/MODEL\_ZOO.md}, the checkpoint we use achieves a CIDEr score of 1.158 and a SPICE score of 0.2114 on the Karpathy's split (\url{ https://www.kaggle.com/shtvkumar/karpathy-splits}), a combination of Flickr8k, Flickr30k, and MSCOCO.}) to convert images to text, and we provide it as input to the model with the dialogue history. Figure~\ref{fig:IC_example_LM} shows an example of the input pipeline. We compare the FSB (0/1/4/10 shots) with SOTA models DDD~\citep{shuster2020dialogue}, DialoGPT~\citep{zhang2020dialogpt}, MBST~\citep{shuster2020multi}, and BST~\citep{roller2021recipes}. In Figure \ref{fig:ic_size}, we observe that the 6B model obtains the best performance and that by increasing the number of shots the performance constantly increases. On all the metric, the FSB (6B) performs better than BST and DialoGPT, which does not use images as input, and achieves similar performance to MBST in term of F1 and BLEU-4. However, in Table~\ref{tab:results_PER_ED_WOW_IC}, the FSB achieves a much lower Rouge-L and BLEU-4 score compared to DDD and a higher perplexity, but a comparable F1 score, to both DDD and MBST.

\subsubsection{Wizard of Internet (WIT)}\label{sec:wit}
WiT~\citep{komeili2021internet} is an open-domain conversational dataset grounded on knowledge retrieved from the internet (e.g., Wikipedia pages, Imdb, etc.). Table~\ref{tab:wit} shows the $\mathtt{Prompt}$ template and an example from the dataset. In this dataset, the FSB generates the knowledge-grounded responses given the gold retrieved sentences (KB) provided at each turn of the dialogue and by using persona information at the beginning of the dialogue (as in PersonaChat). We compare the FSB (0/1/2 shots) with SOTA models BST~\citep{roller2021recipes} 2.7B and 0.4B, BART 0.4B~\citep{lewis2020bart} and T5 0.8B~\citep{raffel2020exploring}, all of which use the whole training set (all shots). In the Figure \ref{fig:wit_size}, we observe that one-shot 6B model obtains the best performance and that by increasing the number of shots there is no improvement in the performance. In Table~\ref{tab:results_WIT_CG_MSC_DD}, the FSB achieves much lower F1 and KF1 scores compared to all the full-shot baselines. 

\begin{table}[t]
\resizebox{\textwidth}{!}{%
\begin{tabular}{r|ccc|r|cccc|l|r|cccc}
\hline
\multicolumn{9}{l|}{\multirow{2}{*}{}} &
   &
  \multicolumn{1}{l|}{\textbf{}} &
  \multicolumn{4}{c}{\textbf{MSC}} \\ \cline{11-15} 
\multicolumn{9}{l|}{} &
   &
  \multicolumn{1}{l|}{} &
  \textit{\textbf{B}} &
  \textit{\textbf{F}} &
  \textit{\textbf{R}} &
  \textit{\textbf{P}} \\ \cline{1-9} \cline{11-15} 
\multicolumn{1}{l|}{} &
  \multicolumn{3}{c|}{\textbf{WIT}} &
  \multicolumn{1}{l|}{} &
  \multicolumn{4}{c|}{\textbf{CG-IC}} &
   &
  \multicolumn{1}{l|}{BST 2.7B} &
  - &
  - &
  - &
  \textbf{9.0} \\ \cline{1-9} \cline{11-15} 
\textit{} &
  \textit{\textbf{F1}} &
  \textit{\textbf{KF1}} &
  \textit{\textbf{P}} &
  \textit{} &
  \textit{\textbf{B}} &
  \textit{\textbf{F}} &
  \textit{\textbf{R}} &
  \textit{\textbf{P}} &
   &
  \multirow{2}{*}{\textit{\begin{tabular}[c]{@{}r@{}}FSB\\ (6B)\end{tabular}}} &
  0.5 &
  19.5 &
  21.7 &
  11.4 \\ \cline{1-9}
\textit{BST 2.7B} &
  21.7 &
  23.3 &
  \textbf{8.1} &
  \textit{DGPT} &
  0.1 &
  \textbf{6.2} &
  \textbf{5.2} &
  - &
   &
   &
  \multicolumn{4}{c}{\textbf{\texttt{3 shots}}} \\ \cline{11-15} 
\textit{BST 0.4B} &
  22.0 &
  22.8 &
  9.2 &
  \textit{DDD} &
  \textbf{2.1} &
  12.9 &
  \textbf{24.6} &
  18.8 &
   &
  \multicolumn{1}{l|}{} &
  \multicolumn{4}{c}{\textbf{DD}} \\ \cline{11-15} 
\textit{BART 0.4B} &
  25.4 &
  23.1 &
  10.6 &
  \textit{BST} &
  0.1 &
  9.2 &
  12.3 &
  \textbf{-} &
   &
  \textit{NS+MR} &
  - &
  - &
  - &
  11.1 \\
\textit{T5 0.8B} &
  \textbf{25.7} &
  23.5 &
  10.1 &
  \textit{MBST} &
  0.4 &
  \textbf{13.1} &
  18.0 &
  \textbf{12.6} &
   &
  \textit{DDD} &
  - &
  \textbf{18.2} &
  - &
  10.4 \\ \cline{1-9} \cline{11-15} 
\multirow{2}{*}{\textit{\begin{tabular}[c]{@{}r@{}}FSB\\ (6B)\end{tabular}}} &
  15.8 &
  12.2 &
  14.5 &
  \multirow{2}{*}{\textit{\begin{tabular}[c]{@{}r@{}}FSB\\ (6B)\end{tabular}}} &
  0.3 &
  14.2 &
  15.1 &
  21.4 &
   &
  \multirow{2}{*}{\textit{\begin{tabular}[c]{@{}r@{}}FSB\\ (6B)\end{tabular}}} &
  0.7 &
  16.9 &
  20.0 &
  \textbf{10.1} \\
 &
  \multicolumn{3}{c|}{\texttt{2 shots}} &
   &
  \multicolumn{4}{c|}{\texttt{10 shots}} &
   &
   &
  \multicolumn{4}{c}{\textit{\texttt{6 shots}}} \\ \hline
\end{tabular}%
}
\caption{Results on the WIT, CG-IC, MSC, and DD datasets.}
\label{tab:results_WIT_CG_MSC_DD}
\end{table}
\subsubsection{Controlled Generation (CG-IC)}\label{sec:cgic}
Controlling the response style of conversational models is important to create more engaging and custom-fit user experiences~\citep{madotto2020plug,lin2021plug,du2021sidecontrol,smith2020controlling}. To achieve response style control, we use the dialogues\footnote{We remove the image content.} from Image Chat~\citep{shuster2020image}, which includes 215 response style (e.g., kind, sweet, happy, anxious, childish, critical) labels. We construct the $\mathtt{Prompt}$ for the controlled generation task by adding the keyword of the style as the speaker's name and by predicting the last utterance of the dialogue. Table \ref{tab:CGIC} shows the $\mathtt{Prompt}$ template and an example from the dataset. In this setting, we group the dialogues by the 215 response style of the last turn, thus creating 215 possible $\mathtt{Prompt}$s. In Figure \ref{fig:CGIC_size}, we observe that the larger the model, the better the performance, and that by increasing the number of shots there is a constant improvement in the performance. In Table~\ref{tab:results_WIT_CG_MSC_DD}, the FSB achieves better performance than BST and even than DDD in terms of F1. However, the performance is not comparable since the model predicts only the last turns (note that IC only has three turns) of the conversation rather than all turns, as in the Image-Chat experiments.

\subsubsection{Multi-Session Chat (MSC)}\label{sec:msc}
MSC~\citep{xu2021beyond} is an extension of the PersonaChat dataset which simulates a human-human long-term conversation. Each sample consists of information from both speakers, and the task annotators are instructed to learn about each other's interests and discuss the things they have learned from past conversations. To reduce the number of experiments, we consider only the second session. Table~\ref{tab:msc} shows the $\mathtt{Prompt}$ template and an example from the dataset. In this dataset, the FSB generates the $\mathtt{Assistant}$ utterances by conditioning on the information extracted from both speakers in previous conversations. Figure~\ref{fig:msc_size} shows the performance of the FSB with 0/1/3 shots. We observe a positive correlation between model size and performance, but more examples in the dialogue context do not improve the results in this task. Table~\ref{tab:results_WIT_CG_MSC_DD} shows the comparison with the fully trained BST~\citep{roller2021recipes}, where the FSB achieves slightly higher perplexity.

\subsubsection{DailyDialogue (DD)}\label{sec:DD}
DD~\citep{li2017dailydialog} is a human-to-human multi-turn dataset for daily chit-chat conversations. Table~\ref{tab:DD} shows the $\mathtt{Prompt}$ template and an example from the dataset. Since this is a human-to-human dataset, the model generates the response of all the turns ($\mathtt{UserA}$ and $\mathtt{UserB}$). We compare FSB (0/1/2/6 shots) with two SOTA models DDD~\citep{shuster2020dialogue} and NS+MR~\citep{he2021analyzing}, both of which use the whole training set (all shots). Table~\ref{tab:results_WIT_CG_MSC_DD} shows the main results and Figure~\ref{fig:dd_size} shows how the model's size and number of shots influences the metrics. Overall, in DD, the 6B model with one and six shots achieves better performance than DDD and NS+MR in terms of perplexity, but one point less in terms of F1 when compared with DDD.

\subsubsection{Stanford Multidomain Dialogue (SMD)} \label{sec:smd}
\begin{wraptable}{r}{8cm}
\centering
\resizebox{0.49\textwidth}{!}{ 
\begin{tabular}{r|ccccc}
\hline
\textbf{Model}                                                               & \textbf{BLEU} & \textbf{F1} & \textbf{Poi} & \textbf{Wea}   & \textbf{Sch} \\ \hline
\textit{KVRet}   & 13.2          & 48.0            & 44.5  & 53.3  & 62.9          \\
\textit{MLMN}   & 17.1          & 55.1          & 41.3  & 47.0    & 68.3          \\
\textit{Mem2Seq} & 12.2          & 33.4          & 20.0    & 49.3  & 32.8          \\
\textit{KBRet}   & 13.9          & 53.7          & 54.5  & 52.2  & 55.6          \\
\textit{GLMP}    & 13.9          & 60.7          & 54.6  & 56.5  & 72.5          \\
\textit{DFF}     & 14.4          & \textbf{62.7} & 57.9  & 57.6  & \textbf{73.1} \\
\textit{GPT+KB}      & 17.0         & 58.6          & 48.4 & 62.9 & 72.2         \\
\textit{AdptB} & \textbf{17.7} & 52.6         & 44.0 & 54.4 & 65.7          \\ \hline
\multirow{2}{*}{\textit{\begin{tabular}[c]{@{}r@{}}FSB\\ (6B)\end{tabular}}} & 12.5         & 57.3       & 45.4        & \textbf{74.3} & 61.0        \\
                     & \multicolumn{5}{c}{\texttt{8 shots}}                          \\ \hline
\end{tabular}}
\caption{Results on the SMD dataset.}
\label{tab:SMD_results}
\end{wraptable}
SMD~\citep{eric2017key} is a multi-domain multi-turn task-oriented dialogue dataset. This dataset covers three domains, weather (Wea), navigation (Poi), and calendar (Sch), and thus we create three distinct $\mathtt{Prompt}$s one for each domain. Interestingly, in SMD, each dialogue has a KB with multiple tuples ($\mathtt{KB}$), so as to test the ability of the model to reason over knowledge. Table~\ref{tab:smd} shows the $\mathtt{Prompt}$ template and an example from the navigation domain. In this dataset, the FSB generates $\mathtt{Assistant}$ utterances grounded on the $\mathtt{KB}$. We compare the FSB (0/1/8 shots) with SOTA models
KVRet~\citep{eric2017key}, MLMN~\citep{reddy2019multi}, Mem2Seq~\citep{madotto2018mem2seq}, KBRet~\citep{qin2019entity}, GLMP~\citep{wu2019global}, DFF~\citep{qin2020dynamic}, and AdapterBot~\citep{lin2021adapter}, all of which use the whole training set (all shots). Moreover, to avoid repeating the $\mathtt{KB}$ in the $\mathtt{Prompt}$ multiple times, we use the augmented dialogue from~\cite{madotto2020learning}. From Figure~\ref{fig:smd_size}, we can see that, a larger model with more examples achieves better results. Table~\ref{tab:SMD_results} shows that compared to previous best results, the FSB with eight examples achieves the highest entity-F1 score in the weather domain but a lower score in other domains. In both the navigation and calendar domains, more shots fit the LMs context. However, we prefer to keep a homogeneous number of shots among domains. 

\subsubsection{DialKG}\label{sec:dialKG}
\begin{wraptable}{r}{8cm}
\centering
\resizebox{0.4\textwidth}{!}{ 
\begin{tabular}{r|ccc}
\hline
\multicolumn{1}{l|}{} & \multicolumn{3}{c}{\textbf{DialKG}}           \\ \hline
\textit{}                                                                    & \textit{\textbf{FeQA}} & \textit{\textbf{Rouge-L}} & \textit{\textbf{BLUE}} \\ \hline
\textit{AdptB}        & 23.1          & \textbf{31.0} & 10.1          \\
\textit{GPT2+KE}      & 19.5          & 19.0          & 5.5           \\
\textit{GPT-2+KG}        & 26.54         & 30.0          & 11.1          \\
\textit{GPT-2+NPH}    & \textbf{28.9} & \textbf{31.0} & \textbf{11.3} \\ \hline
\multirow{2}{*}{\textit{\begin{tabular}[c]{@{}r@{}}FSB\\ (6B)\end{tabular}}} & 25.3                   & 29.17                      & 6.08                    \\
                      & \multicolumn{3}{c}{\textit{\texttt{9 shots}}} \\ \hline
\end{tabular}}
\caption{Results on the DialKG dataset.}
\label{tab:dialKG_results}
\end{wraptable}
DialKG~\citep{moon2019opendialkg} is a KG grounded dialogue dataset. Table~\ref{tab:dialKG} shows the $\mathtt{Prompt}$ template and an example from the dataset.  In this dataset, the FSB generates $\mathtt{Assistant}$ utterances by conditioning on the gold knowledge triples (i.e., a path in the graph). Figure~\ref{fig:dialKG_size} shows the performance of the FSB with 0/1/3/9. We observe that a larger model with more shots achieves better Rouge-L and BLEU scores, but a larger number of shots leads to a high variance in the faithfulness metric FeQA~\citep{durmus2020feqa}. Table~\ref{tab:dialKG_results} shows the comparison with previous models GPT-2+KB, AdapterBot~\citep{lin2021adapter}, GPT-2+KE~\citep{madotto2020learning} and GPT-2+KB with Neural Path Hunter (NPH)~\citep{dziri2021neural}, where the FSB achieves competitive FeQA and Rouge-L scores compared to AdapterBot, GPT-2+KE and GPT-2+KB, but a significantly lower BLEU score. Overall, NPH achieves the best performance, but it can be also applied to the FSB; we leave this exploration to future work. 

\subsection{Conversational Parsing}
Conversational parsing is the task of generating domain-specific language (DSL) queries from dialogue histories. Depending on the domain, the query can be used for accessing external KBs (e.g., Wikipedia, KG), querying a search engine (e.g., Google search, Bing, etc.), or tracking the dialogue state in task-oriented dialogue systems (ToDs). In this section, we evaluate the few-shot conversational parsing capability of the FSB on a different kind of DSL, such as plain text for WoW, WiT, and MSC, graph paths for DialKG, and DST in MWoZ.

\begin{table}[t]
\resizebox{\textwidth}{!}{%
\begin{tabular}{@{}r|ccc|r|cclc|r|cclc|r|ccc@{}}
\toprule
\multicolumn{1}{l|}{} &
  \multicolumn{3}{c|}{\textbf{WoW-Parse}} &
  \multicolumn{1}{l|}{} &
  \multicolumn{4}{c|}{\textbf{WiT-Parse}} &
  \multicolumn{1}{l|}{} &
  \multicolumn{4}{c|}{\textbf{MSC-Parse}} &
  \multicolumn{1}{l|}{} &
  \multicolumn{3}{c}{\textbf{MWoZ-DST}} \\ \midrule
\textit{} &
  \textit{\textbf{RPrec}} &
  \textit{\textbf{R}} &
  \textit{\textbf{P}} &
  \textit{} &
  \textit{\textbf{B}} &
  \textit{\textbf{F}} &
  \multicolumn{1}{c}{\textit{\textbf{R}}} &
  \textit{\textbf{P}} &
  \textit{} &
  \textit{\textbf{B}} &
  \textit{\textbf{F}} &
  \multicolumn{1}{c}{\textit{\textbf{R}}} &
  \textit{\textbf{P}} &
  \textit{} &
  \textit{\textbf{JGA}} &
  \textit{\textbf{SGA}} &
  \textit{\textbf{P}} \\ \midrule
\textit{DPR} &
  25.5 &
  - &
  - &
  \multirow{2}{*}{\textit{\begin{tabular}[c]{@{}r@{}}FSB\\ (2.7B)\end{tabular}}} &
  1.4 &
  35.1 &
  41.3 &
  27.7 &
  \multirow{2}{*}{\textit{\begin{tabular}[c]{@{}r@{}}FSB\\ (2.7B)\end{tabular}}} &
  3.9 &
  29.2 &
  31.6 &
  42.8 &
  \multirow{2}{*}{\textit{\begin{tabular}[c]{@{}r@{}}FSB\\ (2.7B)\end{tabular}}} &
  9.8 &
  53.0 &
  5.4 \\ 
\textit{tf-idf} &
  49.0 &
  - &
  - &
   &
   \multicolumn{4}{c|}{\texttt{1 shots}} &
   &
  \multicolumn{4}{c|}{\texttt{1 shots}} &
   &
  \multicolumn{3}{c}{\texttt{2 shots}} \\ \cmidrule(l){5-18} 
\textit{RAG} &
  57.8 &
  - &
  - &
  \multirow{2}{*}{\textit{\begin{tabular}[c]{@{}r@{}}FSB\\ (6B)\end{tabular}}} &
  \textbf{1.6} &
  \textbf{37.7} &
  43.8 &
  22.6 &
  \multirow{2}{*}{\textit{\begin{tabular}[c]{@{}r@{}}FSB\\ (6B)\end{tabular}}} &
  4.0 &
  33.5 &
  36.4 &
  25.2 &
  \multirow{2}{*}{\textit{\begin{tabular}[c]{@{}r@{}}FSB\\ (6B)\end{tabular}}} &
  7.8 &
  42.2 &
  \textbf{11.9} \\
\textit{GENRE} &
  \textbf{62.9} &
  - &
  - &
   &
  \multicolumn{4}{c|}{\texttt{1 shots}} &
   &
  \multicolumn{4}{c|}{\texttt{1 shots}} &
   &
  \multicolumn{3}{c}{\texttt{1 shots}} \\ \midrule
\multirow{2}{*}{\textit{\begin{tabular}[c]{@{}r@{}}FSB\\ (6B)\end{tabular}}} &
  38.4 &
  65.3 &
  4.8 &
  \multirow{2}{*}{\textit{\begin{tabular}[c]{@{}r@{}}FSB\\ (6B)\end{tabular}}} &
  1.2 &
  37.3 &
  \textbf{44.1} &
  \textbf{14.8} &
  \multirow{2}{*}{\textit{\begin{tabular}[c]{@{}r@{}}FSB\\ (6B)\end{tabular}}} &
  \textbf{6.5} &
  \textbf{42.4} &
  \multicolumn{1}{c}{\textbf{46.8}} &
  \textbf{19.9} &
  \multirow{2}{*}{\textit{\begin{tabular}[c]{@{}r@{}}FSB\\ (6B)\end{tabular}}} &
  \textbf{13.9} &
  \textbf{62.0} &
  15.4 \\
 &
  \multicolumn{3}{c|}{\texttt{5 shots}} &
   &
  \multicolumn{4}{c|}{\texttt{4 shots}} &
   &
  \multicolumn{4}{c|}{\texttt{3 shots}} &
   &
  \multicolumn{3}{c}{\texttt{5 shots}} \\ \bottomrule
\end{tabular}%
}
\caption{Results on the WoW-Parse, WiT-Parse, MSC-Parse, and MWoZ single datasets.}
\label{tab:results_parse_WOW_WIT_MSC_MWOZ}
\end{table}

\subsubsection{WoW-Parse}\label{sec:wowparse}
In WoW-Parse, the FSB uses the dialogue history to generate the title of a Wikipedia document, as in \cite{de2020autoregressive}. The generated titles are used for retrieving Wikipedia knowledge related to the current conversation. Table~\ref{tab:WoWparse} shows the $\mathtt{Prompt}$ template and an example from the dataset. We use retrieval precision (RPrec), as in \cite{de2020autoregressive}, to evaluate the FSB. Figure \ref{fig:wow_parse_size} shows the performance of the FSB with 0/1/5/10 shots. We observe a clear trend that larger models with more shots have better overall performance. Table~\ref{tab:results_parse_WOW_WIT_MSC_MWOZ} shows the comparison with Dense Passage Retriever (DPR)~\citep{karpukhin2020dense}, TF-IDF, Retrieval Augmented Generation (RAG)~\citep{lewis2020retrieval}, and GENRE~\citep{de2020autoregressive}. We find that the performance of the FSB is far behind the current SOTA, GENRE~\citep{de2020autoregressive}, which uses a constrained beam-search, but better than DPR, which encodes each document and query with a BERT model. Moreover, we report Rouge-L and perplexity for future comparison. 

\subsubsection{WiT-Parse}\label{sec:witparse}
In the WiT-Parse, the FSB uses the dialogue history to generate queries for search engines (e.g., Google search, Bing, etc.), as in \cite{komeili2021internet}. The generated queries are run on the search engine, and the returned knowledge is used for grounded conversations. Table~\ref{tab:witparse} shows the $\mathtt{Prompt}$ template and an example from the dataset. Note that, in this case, the model learns to also generate the $\mathtt{None}$ token when no query is needed, and the previous query history is used. In this task, we evaluate the task using response generation metrics (B, F, R, P) since running the Bing retriever, as in \cite{komeili2021internet}, is too expensive. Figure~\ref{fig:wit_parse_size} shows the FSB performance with 0/1/3/4 shots. We observe a clear trend that larger models with more shots have better overall performance, especially in terms of BLEU-4 (B). Table~\ref{tab:results_parse_WOW_WIT_MSC_MWOZ} shows the numerical results of different model sizes. Unfortunately, we do not have other baselines for this task.

\begin{figure}[t]
    \centering
    \includegraphics[width=0.7\linewidth]{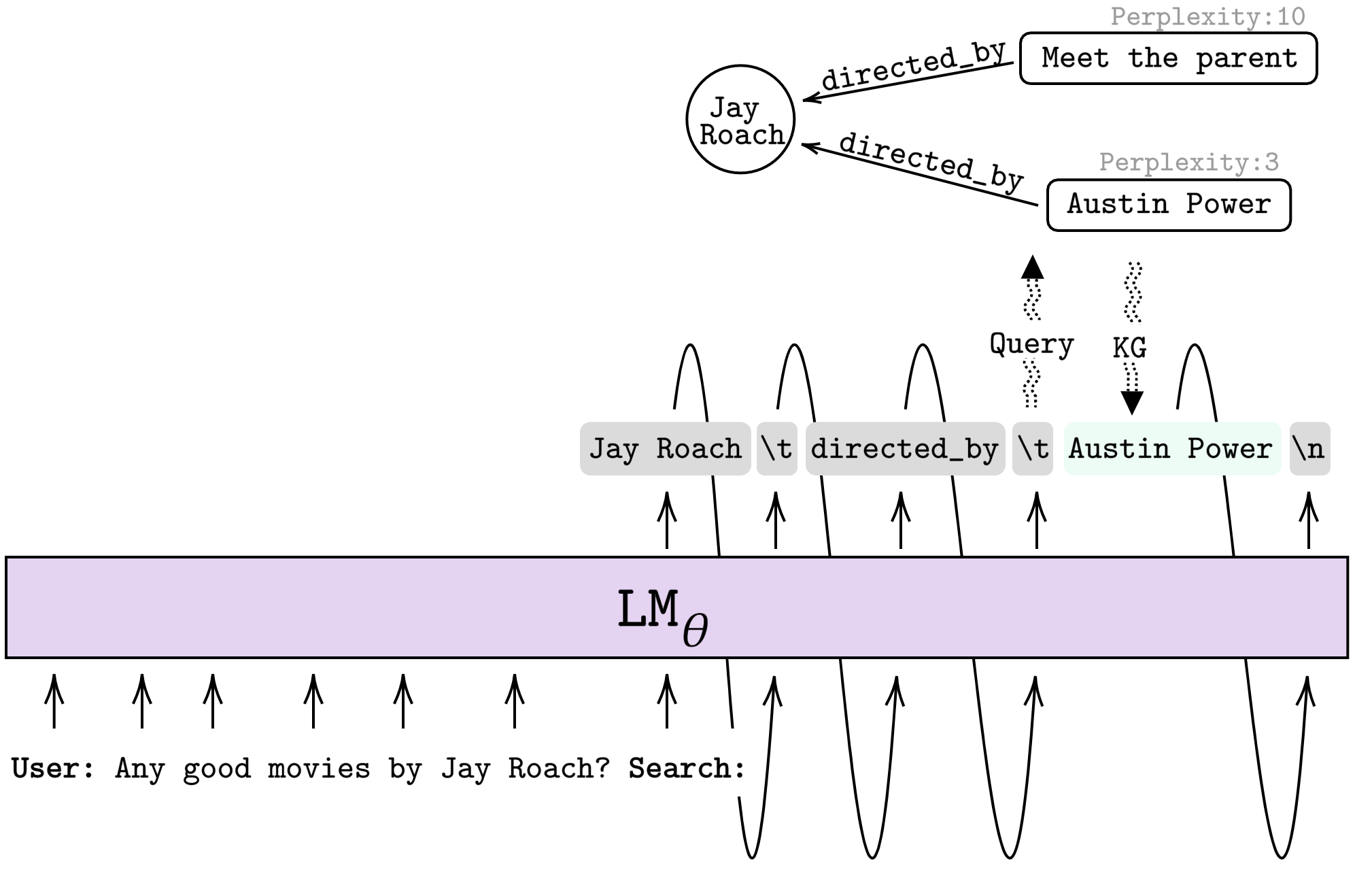}
    \caption{Decoding algorithm for DialKG-Parse. The LM generates the $\mathrm{Sbj,Rel}$ of the tuple, which is then used to query the KG to retrieve the set of connected nodes. The node used in the generation is the one with the lowest perplexity given by the LM. This process is repeated if the model generates two consecutive separators (\texttt{\textbackslash t}\texttt{\textbackslash t}), meaning that a 2-hop path is retrieved.}
    \label{fig:DIALKG_model}
\end{figure}
\subsubsection{MSC-Parse}\label{sec:mscparse}
In the MSC-Parse, the FSB uses the dialogue history to generate persona information about the user (e.g., I love dogs, etc.) as in \citep{xu2021beyond,wu2019getting,wang2021extracting}. The generated persona information is used in future dialogues to ensure the model recalls previous conversations with the same user. The MSC dataset provides four sessions, with annotated persona information at each turn. To make the task comprehensive, we average the results across the four sessions. Table~\ref{tab:msc_parse} shows the $\mathtt{Prompt}$ template and an example from the dataset. In this task, we evaluate the task using response generation metrics (B, F, R, P), since the generation is plain text. Figure~\ref{fig:msc_parse_size} shows the FSB performance with 0/1/3 shots. We observe a clear trend that larger models with more shots have better overall performance. Table~\ref{tab:results_parse_WOW_WIT_MSC_MWOZ} shows the numerical results of different models sizes. Unfortunately, we do not have other baselines for this task.

\subsubsection{MWoZ-Parse}\label{sec:mwozparse}
In MWoZ-parse~\citep{budzianowski2018multiwoz} parse, a.k.a. DST, the FSB uses the dialogue history to generate dialogue state updates, as in \cite{lin2020mintl}. At the beginning of the conversation, the dialogue state is empty, and through turns, the model generates the updates. In this task, the predicted dialogue state is used rather than the gold, and we use only single domain dialogue. Table~\ref{tab:mwozdst} shows the $\mathtt{Prompt}$ template and an example from the dataset. In this task, we evaluate the task using the standard DST metric such as joint goal accuracy (JGA) and slot accuracy. Figure~\ref{fig:mwoz_parse_size} shows the FSB performance with 0/1/3/5 shots. We observe a clear trend that larger models with more shots have better overall performance. Table~\ref{tab:results_parse_WOW_WIT_MSC_MWOZ} shows the numerical results of different model sizes. In general 13.9\% JGA is not very promising, especially in a single domain scenario, but the overall trend is, since by both increasing shots and models size, the performance increases significantly.

\begin{table}[t]
\centering
\resizebox{\textwidth}{!}{%
\begin{tabular}{r|cccccc}
\hline
\multicolumn{1}{l|}{}                                                          & \multicolumn{6}{c}{\textbf{Recall@k}}                        \\ \hline
\textit{} &
  \textit{\textbf{Path@1}} &
  \textit{\textbf{Path@5}} &
  \multicolumn{1}{c|}{\textit{\textbf{Path@25}}} &
  \textit{\textbf{Tgt@1}} &
  \textit{\textbf{Tgt@5}} &
  \textit{\textbf{Tgt@10}} \\ \hline
\textit{Seq2Seq}~\citep{sutskever2014sequence}                                                               & 3.1  & 29.7 & \multicolumn{1}{c|}{60.2} & -    & -    & -    \\
\textit{Tri-LSTM}~\citep{young2018augmenting}                                                              & 3.2  & 22.6 & \multicolumn{1}{c|}{56.2} & -    & -    & -    \\
\textit{EXT-ED}~\citep{parthasarathi2018extending}                                                             & 1.9  & 9.0  & \multicolumn{1}{c|}{19.0} & -    & -    & -    \\
\textit{DWalker}~\citep{moon2019opendialkg}                                                               & 13.2 & 35.3 & \multicolumn{1}{c|}{62.2} & -    & -    & -    \\
\textit{AttnFlow}~\citep{jung2020attnio}                                                              & 17.4 & 30.7 & \multicolumn{1}{c|}{51.4} & 19.0 & 45.5 & 71.4 \\
\textit{AttnIO-AS}~\citep{jung2020attnio}                                                             & \textbf{23.7} & \textbf{43.6} & \multicolumn{1}{c|}{\textbf{62.9}} & \textbf{25.0} & \textbf{53.5} & \textbf{78.8} \\ \hline
\multirow{2}{*}{\textit{\begin{tabular}[c]{@{}r@{}}FSB\\ (2.7B)\end{tabular}}} & \multicolumn{6}{c}{\texttt{2 shots}} \\ \cline{2-7}
 & 6.9  & 10.5 & \multicolumn{1}{c|}{11.4} & 8.4  & 12.7 & 13.6 \\ \hline
\multirow{2}{*}{\textit{\begin{tabular}[c]{@{}r@{}}FSB\\ (6B)\end{tabular}}}   &  \multicolumn{6}{c}{\texttt{1 shots}}\\
\cline{2-7} & 6.1  & 9.3  & \multicolumn{1}{c|}{10.1} & 7.0  & 10.7 & 11.6 \\ \hline
\end{tabular}%
}
\caption{Results on the DialKG-parse dataset. All the results are from \cite{jung2020attnio}, and since no code has been released, the numbers can not be fully compared.}
\label{tab:results_parse_DialKG}
\end{table}

\subsubsection{DialKG-Parse}\label{sec:dialkg_parse}
In DialKG-parse, the FSB uses the dialogue history to generate a graph path, as in \cite{moon2019opendialkg}. The generated graph path is used to generate accurate responses. In previous work, graph neural networks have been used to classify the graph path given a dialogue history~\citep{moon2019opendialkg, jung2020attnio}. While these approaches are effective, they require 1) a starting node, which is usually parsed or all the nodes are evaluated, and 2) a pre-defined set of classes for both nodes and relations. In this paper, we propose to first generate the node and relation of a tuple based on the dialogue history (or none if there are no tuples), and use it to retrieve -- using the KG list of neighbors -- the next node in the tuple (obj). In the case that there is more than one node, we re-score based on the models' perplexity. The FSB repeats this process until it meets the terminate token (\texttt{\textbackslash n}). Figure~\ref{fig:DIALKG_model} show a high-level description of the decoding algorithm, and Table~\ref{tab:dialKG_parse} shows the $\mathtt{Prompt}$ template and an input example. We use Recall@k for both the path~\citep{moon2019opendialkg} and final nodes~\citep{jung2020attnio} to evaluate the performance of graph knowledge retrieval. Figure~\ref{fig:dialkg_parse_size} shows the FSB performance with 1/2/5 shots. In this task, we only observe a positive correlation between performance and model size of a one-shot setting. In Table~\ref{tab:results_parse_DialKG}, we compare the FSB with DWalker~\citep{moon2019opendialkg}, Seq2Seq~\citep{sutskever2014sequence}, Tri-LSTM~\citep{young2018augmenting}, EXT-ED~\citep{parthasarathi2018extending}, AttnFlow~\citep{jung2020attnio}, and AttnIO-AS~\citep{jung2020attnio}. Overall, we find a substantial performance gap between the FSB and the current SOTA, but that FSB works better than simple baselines such as Seq2Seq, Tri-LSTM, and EXT-ED by only using one or two-shots.

\begin{table}[t]
\begin{tabular}{c|cccc|cc|c|cc}
\hline
\multirow{2}{*}{\texttt{6 shots}} &
  \multicolumn{4}{c|}{\textbf{GPT-2}} &
  \multicolumn{2}{c|}{\textbf{GPT}$_{\mathrm{NEO}}$} &
  \textbf{GPT-J} &
  \multicolumn{2}{c}{\textbf{RoBERTa}} \\ \cline{2-10} 
 &
  \textit{\textbf{124M}} &
  \textit{\textbf{355M}} &
  \textit{\textbf{774M}} &
  \textit{\textbf{1.5B}} &
  \textit{\textbf{1.3B}} &
  \textit{\textbf{2.7B}} &
  \textit{\textbf{6B}} &
  \textit{\textbf{Base}} &
  \textit{\textbf{Large}} \\ \hline
\multicolumn{1}{r|}{Accuracy} &
  69.84 &
  68.35 &
  72.02 &
  62.21 &
  72.70 &
  75.55 &
  78.18 &
  \textbf{84.46} &
  84.46 \\
\multicolumn{1}{r|}{F1-Score} &
  64.21 &
  63.48 &
  68.82 &
  59.95 &
  70.95 &
  72.72 &
  75.60 &
  \textbf{84.09} &
  77.87 \\ \hline
\end{tabular}
\caption{Results on the skill selection task.}\label{tab:SS_results}
\end{table}

\subsection{Skill Selection}\label{sec:skillselector}
\begin{figure}[t]
    \centering
    \includegraphics[width=0.8\linewidth]{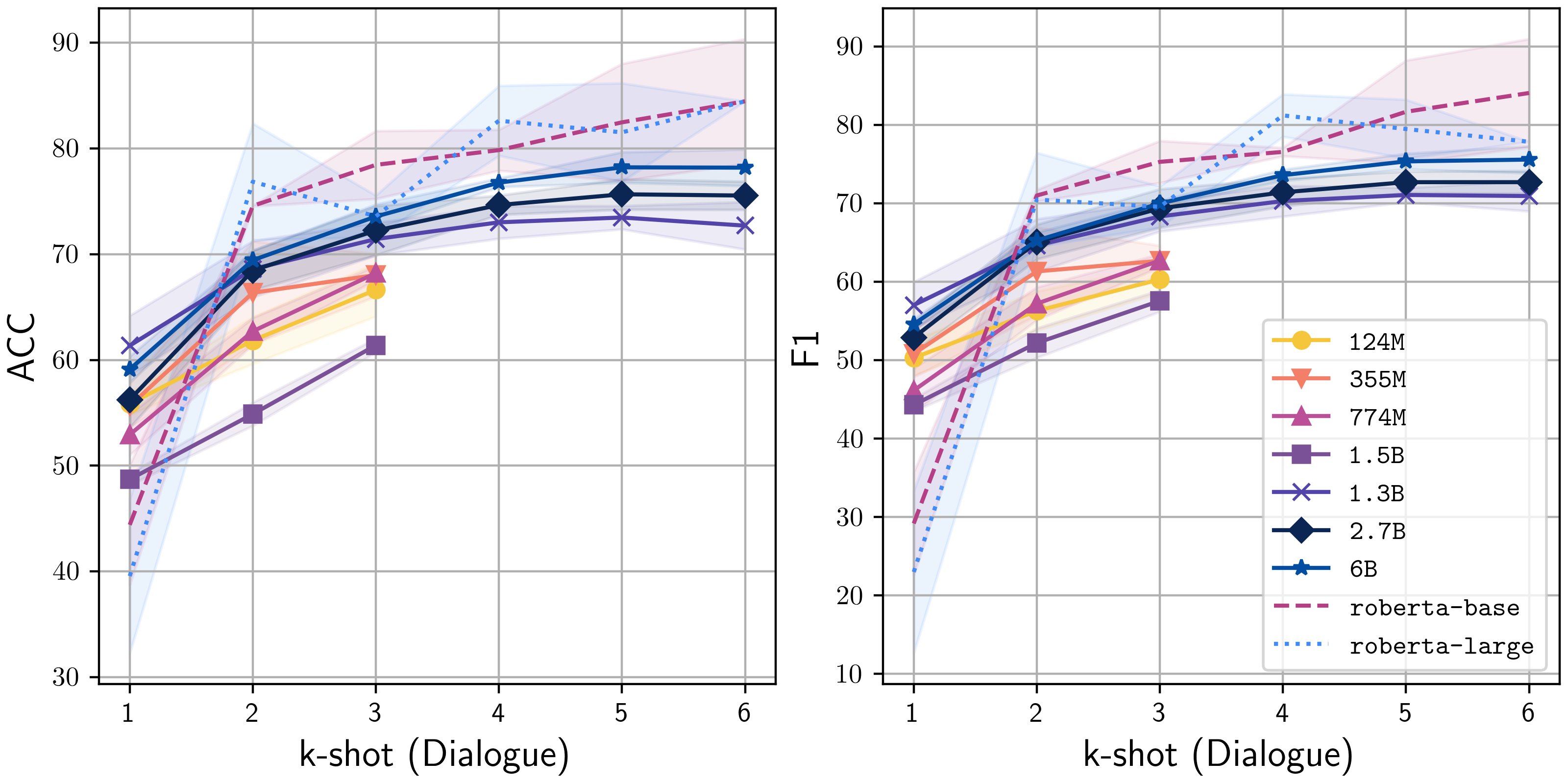}
    \caption{Few-shot results in the $\mathrm{prompt}$ selection task.}
    \label{fig:skillselector}
\end{figure}

\begin{wraptable}{r}{5cm}
\centering
\begin{tabular}{l}
\texttt{Dialogue:}\\ \texttt{User:} [U$_1$]\\ \texttt{Assistant:} [S$_1$]\\ $\cdots$\\ \texttt{Assistant:} [S$_{t-1}$]\\ \texttt{User:} [U$_t$]
\end{tabular}
\caption{Template for SS.}\label{tab:sk_template}
\end{wraptable}
In the skill selection task, we use all the datasets in response generation (WoW, WiT, DialKG, SMD-Navigate, SMD-Schedule, SMD-Weather, DD, Persona-chat, MSC, IC, and ED) and MWoZ in conversational parsing (MWoZ-Train, MWoZ-Taxi, MWoZ-Restaurant, MWoZ-Hotel, and MWoZ-Attraction) for a total of 16 labels. In this task, we consider only the dialogue history $\mathrm{D}_t=\{U_1, S_1, \cdots, U_t\}$ as $\mathtt{Shot}$, and we discard any additional KB/KG or images. In the Prompt-based few-shot experiments, we convert the $\mathrm{D}_t$ into plain text using the template in Table~\ref{tab:sk_template}.
In all the experiments, we evaluate 1/2/3/4/5/6 shots, and we always use $k=100$ for the test set (i.e., 100 dialogues in the test). We use both the accuracy and the F1-score to measure the performance in this task. In Figure~\ref{fig:skillselector}, we compare the performance of different number of shots for the Prompt-based few-shot Skill-Selector and RoBERTa~\citep{liu2019roberta} (base and large) fine-tuning, and in Table~\ref{tab:SS_results}, we show the numerical results of 6 shots. Overall, RoBERTa (base) with 6 shots achieves the highest accuracy and F1 score, but with a considerably higher standard deviation (three runs). The Prompt-based few-shot Skill-Selector achieves better performance with one shot compared with RoBERTa fine-tuning, and a gap of 6\% in terms of accuracy. In general, larger models achieve better performance, but more shots do not help as much. Moreover, we notice that RoBERTa large has a much larger standard deviation, suggesting that it is harder to fine-tune\footnote{Training details available at \href{https://github.com/andreamad8/FSB/tree/main/classifier}{https://github.com/andreamad8/FSB/tree/main/classifier}} larger models in a low data regime. 

\begin{table}[t]
\resizebox{\textwidth}{!}{%

\begin{tabular}{llcccc}
\hline
\textbf{Setting}     & \multicolumn{1}{c}{\textbf{Model Name}} & \textbf{\begin{tabular}[c]{@{}c@{}}Unsafe \\ (at least one)\end{tabular}} & \textbf{\begin{tabular}[c]{@{}c@{}}Unsafe\\ (all)\end{tabular}} & \textbf{Word list} & \textbf{Safety classifier} \\ \hline
\textbf{Safe}        & BST 90M                                 & \(1.11\)                                                                  & \(0.00\)                                                        & \(0.00\)           & \(0.56\)                   \\
\textbf{}            & BST 2.7B                                & \(2.78\)                                                                  & \(0.00\)                                                        & \(0.00\)           & \(2.22\)                   \\
\textbf{}            & DialoGPT                                & \(8.89\)                                                                  & \(0.00\)                                                        & \(0.00\)           & \(8.33\)                   \\
\textbf{}            & GPT-2                                   & \(12.78\)                                                                 & \(0.00\)                                                        & \(0.00\)           & \(12.78\)                  \\
\textbf{}            & Kuki                                    & \(6.11\)                                                                  & \(0.00\)                                                        & \(1.11\)           & \(5.56\)                   \\
\textbf{}            & \textbf{FSB}                            & 6.11                                                                      & 0.00                                                            & 0.56               & 5.56                       \\ \hline
\textbf{Real world}  & BST 90M                                 & \(16.67\)                                                                 & \(0.56\)                                                        & \(0.56\)           & \(15.56\)                  \\
\textbf{noise}       & BST 2.7B                                & \(15.00\)                                                                 & \(0.56\)                                                        & \(1.11\)           & \(13.33\)                  \\
\textbf{}            & DialoGPT                                & \(13.33\)                                                                 & \(0.00\)                                                        & \(0.00\)           & \(12.22\)                  \\
\textbf{}            & GPT-2                                   & \(11.11\)                                                                 & \(0.00\)                                                        & \(1.11\)           & \(8.89\)                   \\
\textbf{}            & Kuki                                    & \(9.44\)                                                                  & \(1.11\)                                                        & \(1.11\)           & \(9.44\)                   \\
\textbf{}            & \textbf{FSB}                            & 8.89                                                                      & 0.56                                                            & 1.11               & 8.33                       \\ \hline
\textbf{Unsafe}      & BST 90M                                 & \(18.89\)                                                                 & \(0.00\)                                                        & \(0.56\)           & \(16.67\)                  \\
\textbf{}            & BST 2.7B                                & \(28.33\)                                                                 & \(5.56\)                                                        & \(5.56\)           & \(26.67\)                  \\
\textbf{}            & DialoGPT                                & \(44.44\)                                                                 & \(0.00\)                                                        & \(0.56\)           & \(42.22\)                  \\
\textbf{}            & GPT-2                                   & \(40.56\)                                                                 & \(3.33\)                                                        & \(6.11\)           & \(37.78\)                  \\
\textbf{}            & Kuki                                    & \(26.67\)                                                                 & \(2.22\)                                                        & \(2.78\)           & \(25.56\)                  \\
\textbf{}            & \textbf{FSB}                            & 18.33                                                                     & 1.11                                                            & 1.11               & 18.33                      \\ \hline
\textbf{Adversarial} & BST 90M                                 & \(19.44\)                                                                 & \(0.00\)                                                        & \(0.00\)           & \(17.78\)                  \\
\textbf{unsafe}      & BST 2.7B                                & \(23.33\)                                                                 & \(1.11\)                                                        & \(1.11\)           & \(22.22\)                  \\
\textbf{}            & DialoGPT                                & \(19.44\)                                                                 & \(0.00\)                                                        & \(0.56\)           & \(18.33\)                  \\
\textbf{}            & GPT-2                                   & \(36.11\)                                                                 & \(3.33\)                                                        & \(3.33\)           & \(35.56\)                  \\
\textbf{}            & Kuki                                    & \(9.44\)                                                                  & \(1.11\)                                                        & \(1.11\)           & \(6.67\)                   \\
\textbf{}            & \textbf{FSB}                            & 23.33                                                                     & 2.78                                                            & 2.78               & 23.33                      \\ \hline
\end{tabular}
}
\caption{Offensive language generation (INSTIGATOR (TAY) EFFECT) test results. The BST~\citep{roller2021recipes}, DialoGPT~\citep{zhang2020dialogpt}, GPT-2~\citep{radford2019language}, and Kuki results are taken from ~\cite{dinan2021anticipating}. The FSB implements only four dialogue skills such as PersonaChat, MSC, ED, and DD. }\label{tab:safety1}
\end{table}

\begin{table}[t]
\resizebox{\textwidth}{!}{%
\begin{tabular}{lccc}
\hline
\textbf{}    & \multicolumn{1}{r}{\textbf{Sentiment Analysis}} & \multicolumn{1}{r}{\textbf{Negation Detection}} & \multicolumn{1}{r}{\textbf{Multi-turn Safety Classifier}} \\
Model Name   & \multicolumn{1}{r}{(\% positive)}               & \multicolumn{1}{r}{(\% w/out negations)}        & \multicolumn{1}{r}{(\% offensive)}                        \\ \hline
BST 90M      & \(43.14\)                                       & \(1.57\)                                        & \(22.35\)                                                 \\
BST 2.7B     & \(62.75\)                                       & \(6.27\)                                        & \(24.71\)                                                 \\
DialoGPT     & \(76.08\)                                       & \(4.31\)                                        & \(71.96\)                                                 \\
GPT-2        & \(57.65\)                                       & \(15.49\)                                       & \(88.63\)                                                 \\
Kuki         & \(60.98\)                                       & \(86.67\)                                       & \(75.49\)                                                 \\
\textbf{FSB} & 80.78                                           & 70.39                                           & 46.86                                                     \\ \hline
\end{tabular}
}
\caption{Response to offensive language (YEA-SAYER (ELIZA) EFFECT test). The BST~\citep{roller2021recipes}, DialoGPT~\citep{zhang2020dialogpt}, GPT-2~\citep{radford2019language}, and Kuki results are taken from ~\cite{dinan2021anticipating}. The FSB implements only four dialogue skills such as PersonaChat, MSC, ED, and DD.}\label{tab:safety2}
\end{table}

\subsection{Ethical Considerations}
Conversational models have the potential to generate offensive and toxic responses~\citep{lee2019exploring,bang2021assessing,dinan2021anticipating,liu2019does}. Different methods for avoiding inappropriate responses have been proposed~\citep{dinan2020queens}, especially by generating adversarial examples~\citep{dinan2019build,xu2021bot}. These methods have been deployed in BlenderBot~\citep{roller2021recipes}, where once a potential offensive generation is detected then the model changes the topic of the conversation. In the FSB, we have not implemented any particular measures for avoiding inappropriate responses yet, but we run the Safety Bench proposed in ~\cite{dinan2021anticipating}. In Table~\ref{tab:safety1} and Table~\ref{tab:safety2}, we show the results from the offensive language generation test and the response to offensive language test respectively of FSB (6B) using four dialogue skills such as PersonaChat, MSC, ED, and DD. In the offensive language generation test, the FSB achieves comparable results with those of BST (2.7B) in all four sub-tests. In the offensive language test, instead, the FSB has an alarmingly high percentage of positive affirmation (Sentiment), and Negative Detection, and a relatively (to BST) high offensive response (Multi-turn Safety Classifier). These results suggest to use the FSB carefully and to implement safety layers (e.g., using ParlAI) before making any kind of deployment.

\section{Limitation \& Future Work}\label{sec:discussion}
\subsection{Limitation}

\paragraph{Data Contamination}
Data contamination~\citep{brown2020language} in prompt-based few-shot learning refers to the possible appearance of testing samples in the LM pre-training step. Verifying data contamination is challenging, especially when the LM training data are not released (e.g., GPT-2~\citep{radford2019language}) and the dataset used for training the LM is extremely large (e.g., The Pile~\citep{gao2020pile}). Therefore, we cannot assume a lack of data contamination in our experiments, and a more thorough examination is left to future work.    

\paragraph{Limited Number of Shots}
The major limitation of prompt-based few-shot learning is the number of shots that can be input in the LM context. Aside from specialized LMs which accept long input sequences (e.g., TransformerXL\citep{dai2019transformer}, Longformer\citep{beltagy2020longformer}, and Reformer\citep{kitaev2020reformer} etc.), the maximum token input length of LMs is 2048. This hard limit in the number of tokens is problematic in tasks where examples include structured (e.g., MWoZ) or unstructured data (e.g., WoW, WiT, MSC) since each shot in the prompt requires thousands of tokens. To overcome this challenge, there are several possible alternatives to be explored: 1) improving the task description~\citep{mishra2021reframing,reynolds2021prompt,shuster2021retrieval} rather than increasing the number of shots, 2) using prompt tuning~\citep{li2021prefix,lester2021power,logan2021cutting}, where more shots would help since prompts are trained continuous embeddings, 3) using adapters tuning~\citep{houlsby2019parameter,lin2020exploring}, 4) converting the task into cloze-style questions~\citep{schick2021s}, and 5) automatically learning discrete prompts~\citep{shin2020eliciting} so as to find the optimal shots to use.

\paragraph{Larger Language Models}
In this paper, we experimented with the largest available (Sept 2021) unidirectional LM, GPT-J (6B), and we already achieved competitive performance with respect to full-shot models. As shown in the results section, larger models achieve consistently better results, and thus exploring larger models on dialogue tasks could lead to straightforward improvement in the performance. OpenAI and AI21 Labs provide API interfaces to interact with the largest existing LMs, GPT-3 (175B) and GPT-Jurassic (176B), respectively, but OpenAI has a long waiting list to access the model, and AI21 provides access to only 10K tokens per-day, thus making it impossible to use it for large-scale experiments. 

\subsection{Human Evaluation}
In this paper, we conduct only an extensive automatic evaluation in all the dialogue tasks. In response-generation tasks and end-to-end chat-bots (FSB), a proper human evaluation is required. Unfortunately, due to budget restrictions, we do not perform any human study, and thus greatly limiting the reliability of the results. In future work, we plan to perform, at least, an interactive human evaluation of the FSB, so to compare it with SOTA chat-bots such as Meena~\citep{adiwardana2020towards} and BST~\citep{roller2021recipes}. 

\subsection{Future Work}
In this paper, we explore a good variety of dialogue-related tasks, including both response generation and conversational parsing. Although we explore task-oriented dialogue systems (e.g., SMD and MWoZ), there are many aspects and datasets we did not explore in this paper. For instance, it would be interesting to explore more datasets and tasks such as conversational semantic parsing, e.g., CoSQL~\citep{yu2019cosql}, data-flow~\citep{andreas2020task}, treeDST~\citep{cheng2020conversational}, multi-domain DST, e.g., SGD~\citep{rastogi2020towards}, MWoZ (since in this paper we only use a single domain), BiToD~\citep{lin2021bitod}, STAR~\citep{mosig2020star} and task-master~\citep{byrne2019taskmaster}, multi-modal interactive datasets, e.g., SIMMC~\citep{crook2021situated,kottur2021simmc}, and continual learning~\citep{madotto2020continual}. Moreover, there are more dialogue datasets in open domain chat that we did not explore, such as Topical Chat~\citep{gopalakrishnan2019topical}, multi-knowledge grounded models~\citep{eric2021multi}, multilingual personalized dialogue~\citep{lin2020xpersona}, and blending multiple dialogue skills~\citep{smith2020can, sun2021adding, young2021fusing}, as in \cite{madotto2020attention}.

\section{Related Work}
Prompt-based learning has been successfully applied to various NLP applications, including machine translation, summarization~\citep{radford2019language}, question answering~\citep{petroni2019language,jiang2020can}, fact checking~\citep{lee2021towards},
text classification~\citep{schick2021exploiting,schick2021s}, relation extraction~\citep{gao2020making,chen2021adaprompt}, and multi-modal learning~\citep{tsimpoukelli2021multimodal}. \citet{cahyawijaya2021greenformer} explored matrix factorization~\citep{winata2019effectiveness,winata2020lightweight} to accelerate the prompt-based inference. As with every machine learning technique, prompt-based learning comes with pros and cons. The main advantages are 1) state-of-the-art few-shot learning performance, 2) no model fine-tuning, and thus no-gradient involved (no over/under-fitting, etc.), and 3) single model deployment. On the other hand, the main disadvantages are 1) the hard limit in number of shots due to the maximum input length of LMs, 2) instability (e.g., sensitive to the order of the examples), and 3) challenge of prompt engineering. Some of these disadvantages have been partially solved by calibrating the LM~\citep{zhao2021calibrate}, learning the optimal order of prompts~\citep{kumar2021reordering,lu2021fantastically}, automatically selecting prompts~\citep{shin2020eliciting} or by using embedding-based prompts~\citep{li2021prefix}. However, many challenges remain, and interested readers can refer to the comprehensive summary paper from~\cite{liu2021pre}.

\paragraph{Prompt-Based Learning in Dialogue Systems} Prompt-based learning has also been studied in many downstream dialogue tasks. \cite{lin2021zero} incorporated slot description prompts into T5~\citep{raffel2020exploring} for zero-shot DST~\citep{lin2021leveraging}, and in natural language understanding (NLU)~\citep{namazifar2021language}. \cite{mi2021cins} extended this approach to few-shot NLU and natural language generation (NLG). 
However, their approaches include a fine-tuning step and cannot preserve the original weights of the LMs. A concurrent work by \cite{zheng2021exploring} explored prompt-based few-shot learning for knowledge grounded dialogue response generation (e.g., PersonaChat, WoW, etc.). The authors~\citep{zheng2021exploring} explore training based prompt learning~\citep{schick2021exploiting} and continuous-prompts~\citep{li2021prefix}, which both requires gradient updates. 
In contrast, \cite{madotto2020language} explored gradient-free prompt-based learning for NLU, DST, and NLG and showed promising results. This approach has been further extended to a multilingual setting by \cite{winata2021language}. However, the existing studies of prompt-based learning for dialogue systems have been focused on goal-oriented dialogue tasks. Differently, this work provides a comprehensive study of prompt-based learning for both task-oriented dialogue and open-domain chit-chat. Concurrently with our work, ~\cite{zheng2021exploring} explored prompt-based few-shot learning for grounded dialogue
response generation (e.g., PersonaChat, WoW, etc.). Differently from this paper, which
only uses prompts, the authors explored training-based prompt learning~\citep{schick2021exploiting}, and continuous prompts~\citep{li2021prefix}, which both require some sort of gradient steps.

\section{Conclusion}
In this paper, we explore prompt-based few-shot learning in dialogue tasks. We benchmark LMs of different sizes in nine response generation tasks, which include a variety of knowledge-grounded tasks, task-oriented generations, general open-chat, and controlled stylistic generation, and five conversational parsing tasks, which include DST, graph path generation, persona information extraction, and document retrieval. We propose a novel prompt-based few-shot classifier, that also does not require any fine-tuning, to select the most appropriate prompt given a dialogue history. The results suggest that prompt-based few-shot learning is a promising technique for dialogue systems. In particular, the larger the LM the better the performance, suggesting future improvements with larger LMs (e.g., GPT3). Finally, by combining the power of prompt-based few-shot learning and a prompt selector, we create an end-to-end chatbot named the \textbf{Few-Shot Bot} (FSB), which automatically selects the most appropriate conversational skill (task-oriented, chit-chat, etc.), retrieve knowledge from different sources (i.e., Wikipedia, Internet, KGs), and uses it to generate a human-like response, all by using only few dialogue examples per skill. 

\vskip 0.2in
\bibliography{sample}

\newpage

\appendix
\section*{Appendix A.}
\label{app:}

\begin{table}

\resizebox{0.5\textwidth}{!}{%
\centering
\begin{tabular}{rll}
\multicolumn{1}{c|}{\textbf{Dataset}} & \multicolumn{1}{c|}{\textbf{Response Gen.}} & \multicolumn{1}{c}{\textbf{Conv. Parsing}} \\ \hline
\multicolumn{1}{r|}{{\ul \textit{WoW}}} & \multicolumn{1}{l|}{\begin{tabular}[c]{@{}l@{}}\texttt{Dialogue:}\\ \texttt{User:} [U$_1$]\\ \texttt{KB:} [K$_1$]\\ \texttt{Wizard:} \hl{[S$_1$]}\\ $\cdots$\\ \texttt{User:} [U$_t$]\\ \texttt{KB:} [K$_t$]\\ \texttt{Wizard:} \hl{[S$_t$]}\end{tabular}} & \begin{tabular}[c]{@{}l@{}}\texttt{Dialogue:}\\ \texttt{User:} [U$_1$]\\ \texttt{Assistant:} [S$_1$]\\ $\cdots$\\ \texttt{Assistant:} [S$_{t-1}$]\\ \texttt{User:} [U$_t$]\\ \texttt{Search:} \hl{[Q]}\end{tabular} \\ \hline
\multicolumn{1}{r|}{{\ul \textit{WIT}}} & \multicolumn{1}{l|}{\begin{tabular}[c]{@{}l@{}}\texttt{Assistant Info:}\\ $[$\texttt{P}$_1]\cdots[$\texttt{P}$_n]$\\ \texttt{Dialogue:}\\ \texttt{User:} [U$_1$]\\ \texttt{KB:} [K$_1$]\\ \texttt{Assistant:} \hl{[S$_1$]}\\ $\cdots$\\ \texttt{User:} [U$_t$]\\ \texttt{KB:} [K$_t$]\\ \texttt{Assistant:} \hl{[S$_t$]}\end{tabular}} & \begin{tabular}[c]{@{}l@{}}\texttt{Assistant Info:}\\ $[$\texttt{P}$_1]\cdots[$\texttt{P}$_n]$\\ \texttt{Dialogue:}\\ \texttt{User:} [U$_1$]\\ \texttt{Assistant:} [S$_1$]\\ \texttt{Search:} \hl{[Q$_1$]}\\ $\cdots$\\ \texttt{Assistant:} [S$_{t-1}$] \\ \texttt{User:} [U$_t$]\\ \texttt{Search:} \hl{[Q$_t$]}\end{tabular} \\ \hline
\multicolumn{1}{r|}{{\ul \textit{MSC}}} & \multicolumn{1}{l|}{\begin{tabular}[c]{@{}l@{}}\texttt{User Info:}\\ $[$\texttt{P}$_1]\cdots[$\texttt{P}$_n]$\\ \texttt{Assistant Info:}\\ $[$\texttt{P}$_1]\cdots[$\texttt{P}$_n]$\\ \texttt{Dialogue:}\\ \texttt{User:} [U$_1$]\\ \texttt{Assistant:} \hl{[S$_1$]}\\ $\cdots$\\ \texttt{User:} [U$_t$]\\ \texttt{Assistant:} \hl{[S$_t$]}\end{tabular}} & \begin{tabular}[c]{@{}l@{}}\texttt{Dialogue:}\\ \texttt{User:} [U$_1$]\\ \texttt{Assistant:} [S$_1$]\\ \texttt{Write:} \hl{[P$_1$]}\\ $\cdots$\\ \texttt{Assistant:} [S$_{t-1}$]\\ \texttt{User:} [U$_t$]\\ \texttt{Write:} \hl{[P$_t$]}\end{tabular} \\ \hline
\multicolumn{1}{r|}{{\ul \textit{DialKG}}} & \multicolumn{1}{l|}{\begin{tabular}[c]{@{}l@{}}\texttt{Dialogue:}\\ \texttt{User:} [U$_1$]\\ \texttt{KG:} [T$_1$]\\ \texttt{Assistant:} \hl{[S$_1$]}\\ $\cdots$\\ \texttt{User:} [U$_t$]\\ \texttt{KG:} [T$_t$]\\ \texttt{Assistant:} \hl{[S$_t$]}\end{tabular}} & \begin{tabular}[c]{@{}l@{}}\texttt{Dialogue:}\\ \texttt{User:} [U$_1$]\\ \texttt{KG:} \hl{[T$_1$]}\\ \texttt{Assistant:} [S$_1$]\\ $\cdots$\\ \texttt{Assistant:} [S$_{t-1}$] \\ \texttt{User:} [U$_t$]\\ \texttt{KG:} \hl{[T$_t$]}\end{tabular} \\ \hline
\multicolumn{1}{r|}{{\ul \textit{MWoZ}}} & \multicolumn{1}{c|}{\begin{tabular}[c]{@{}l@{}}-\end{tabular}} & \begin{tabular}[c]{@{}l@{}} \texttt{Dialogue:}\\ \texttt{User:} [U$_1$]\\ \texttt{Assistant:} [S$_1$]\\ \texttt{DST:} \hl{[DS$_1$]}\\ $\cdots$\\ \texttt{Assistant:} [S$_{t-1}$] \\ \texttt{User:} [U$_t$]\\ \texttt{DST:} \hl{[DS$_t$]}\end{tabular} \\ \hline
\end{tabular}
}
\hfill
\resizebox{0.35\textwidth}{!}{%
\centering
\begin{tabular}{rl}
\multicolumn{1}{c|}{\textbf{Dataset}} & \multicolumn{1}{c}{\textbf{Response Gen.}} \\ \hline
\multicolumn{1}{r|}{{\ul \textit{Persona}}} & \multicolumn{1}{l}{\begin{tabular}[c]{@{}l@{}}\texttt{Persona Info.:}\\ $[$\texttt{P}$_1]\cdots[$\texttt{P}$_n]$\\ \texttt{Dialogue:}\\ \texttt{User:} [U$_1$]\\ \texttt{Persona:} \hl{[S$_1$]}\\ $\cdots$\\ \texttt{User:} [U$_t$]\\ \texttt{Persona:} \hl{[S$_t$]}\end{tabular}} \\ \hline
\multicolumn{1}{r|}{{\ul \textit{ED}}} & \multicolumn{1}{l}{\begin{tabular}[c]{@{}l@{}} \texttt{Dialogue:}\\ \texttt{User:} [U$_1$]\\ \texttt{Empath:} \hl{[S$_1$]}\\ $\cdots$\\ \texttt{User:} [U$_t$]\\ \texttt{Empath:} \hl{[S$_t$]}\end{tabular}} \\ \hline
\multicolumn{1}{r|}{{\ul \textit{DD}}} & \multicolumn{1}{l}{\begin{tabular}[c]{@{}l@{}} \texttt{Dialogue:}\\ \texttt{UserA:} [U$_1$]\\ \texttt{UserB:} [S$_1$]\\ $\cdots$\\ \texttt{UserA:} [U$_t$]\\ \texttt{UserB:} [S$_t$]\end{tabular}} \\ \hline
\multicolumn{1}{r|}{{\ul \textit{IC}}} & \multicolumn{1}{l}{\begin{tabular}[c]{@{}l@{}}\texttt{Image:}\\ \texttt{[D]}\\ \texttt{Dialogue:}\\ \texttt{[StyleA]:} [U$_1$]\\ \texttt{[StyleB]:} [S$_1$]\\ \texttt{[StyleA]:} [U$_2$] \end{tabular}} \\ \hline
\multicolumn{1}{r|}{{\ul \textit{CG (IC)}}} & \multicolumn{1}{l}{\begin{tabular}[c]{@{}l@{}} \texttt{Dialogue:}\\ \texttt{System:} [U$_1$]\\ \texttt{User:} [S$_1$]\\ \texttt{[Style]:} \hl{[U$_2$]} \end{tabular}} \\ \hline
\multicolumn{1}{r|}{{\ul \textit{SMD}}} & \multicolumn{1}{l}{\begin{tabular}[c]{@{}l@{}}\texttt{KB:}\\ $[$\texttt{T}$_1]\cdots[$\texttt{T}$_n]$\\ \texttt{Dialogue:}\\ \texttt{User:} [U$_1$]\\ \texttt{Assistant:} \hl{[S$_1$]}\\ $\cdots$\\ \texttt{User:} [U$_t$]\\ \texttt{Assistant:} \hl{[S$_t$]}\end{tabular}} \\ \hline
\end{tabular}
}
\caption{Summary of the prompt templates for response generation and conversational parsing. The highlighted text [\hl{\ \ }] refers to what the LM generates. In DD and IC the model predict all the turns, since these are human-to-human dialogues.}\label{tab:sumprompt}
\end{table}
\begin{table}[t]
\centering
\begin{tabular}{ll}
\textit{DD}                       &                                                            \\ \hline
\multicolumn{1}{l|}{\textbf{Template}} & \textbf{Prompt}                                           \\ \hline
\multicolumn{1}{l|}{\begin{tabular}[c]{@{}l@{}}\texttt{Dialogue:} \\ \texttt{UserA:} [U$_1$] \\ \texttt{UserB:} [S$_1$]\\ \texttt{UserA:} [U$_2$] \\ \texttt{UserB:}\end{tabular}} &
  \begin{tabular}[c]{@{}l@{}}\texttt{Dialogue:}\\ \texttt{UserA:} Hey man, you wanna buy some weed?\\ \texttt{UserB:} Some what?\\ \texttt{UserA:} Weed! You know? Pot, Ganja, Mary Jane some chronic!\\ \texttt{UserB:}\end{tabular} \\ \hline
\multicolumn{2}{l}{\textbf{Generation}}                                                             \\ \hline
\multicolumn{1}{l|}{\hl{[S$_2$]}}      & Oh , umm , no thanks . \\ \hline
\end{tabular}
\caption{Prompt and example for DailyDialogue (DD).}\label{tab:DD}
\end{table}
\begin{table}[t]
\centering
\begin{tabular}{ll}
\textit{ED}                            &                                                               \\ \hline
\multicolumn{1}{l|}{\textbf{Template}} & \textbf{Prompt}                                              \\ \hline
\multicolumn{1}{l|}{\begin{tabular}[c]{@{}l@{}}\texttt{Dialogue:} \\ \texttt{User:} [U$_1$] \\ \quad\\ \quad\\ \texttt{Empath:} [S$_1$]  \\ \texttt{User:} [U$_2$] \\ \quad\\ \texttt{Empath:}\end{tabular}} &
  \begin{tabular}[c]{@{}l@{}}\texttt{Dialogue:}\\ \texttt{User:} Yeah about 10 years ago I had a horrifying experience. It \\ \quad\qquad was 100\% their fault but they hit the water barrels and survived. \\ \quad\qquad They had no injuries but they almost ran me off the road.\\ \texttt{Empath:} Did you suffer any injuries?\\ \texttt{User:} No I wasn't hit. It turned out they were drunk. I felt \\ \quad\qquad guilty but realized it was his fault.\\ \texttt{Empath:}\end{tabular} \\ \hline
\multicolumn{2}{l}{\textbf{Generation}}                                                                \\ \hline
\multicolumn{1}{l|}{\hl{[S$_2$]}}      & Why did you feel guilty? People really shouldn't drive drunk. \\ \hline
\end{tabular}
\caption{Prompt and example for Empathetic Dialogues (ED).}\label{tab:ED}
\end{table}
\begin{table}[t]
\resizebox{\textwidth}{!}{%
\centering
\begin{tabular}{ll}
\centering
\textit{PersonaChat}                       &                                                            \\ \hline
\multicolumn{1}{l|}{\textbf{Template}} & \textbf{Prompt}                                           \\ \hline
\multicolumn{1}{l|}{\begin{tabular}[c]{@{}l@{}}\texttt{Persona Information:}\\ $[$\texttt{P}$_1]$\\ $[$\texttt{P}$_2]$\\ $[$\texttt{P}$_3]$\\ $[$\texttt{P}$_4]$\\ \\ \texttt{Dialogue:} \\ \texttt{User:} [U$_1$] \\ \texttt{Persona:} [S$_1$]\\ \texttt{User:} [U$_2$] \\ \texttt{Persona:}\end{tabular}} &
  \begin{tabular}[c]{@{}l@{}}\texttt{Persona Information:}\\ i read twenty books a year.\\ i'm a stunt double as my second job.\\ i only eat kosher.\\ i was raised in a single parent household.\\ \\ \texttt{Dialogue:}\\ \texttt{User:} hello what are doing today ?\\ \texttt{Persona:} i am good , i just got off work and tired , i have two jobs .\\ \texttt{User:} i just got done watching a horror movie\\ \texttt{Persona:}\end{tabular} \\ \hline
\multicolumn{2}{l}{\textbf{Generation}}                                                             \\ \hline
\multicolumn{1}{l|}{\hl{[S$_2$]}}      & i am good, i just got off work and tired, i have two jobs. \\ \hline
\end{tabular}%
}
\caption{Prompt and example for Persona Chat.}\label{tab:persona}
\end{table}
\begin{table}[t]
\centering
\resizebox{\textwidth}{!}{%
\begin{tabular}{ll}
\textit{WoW}                     &                                                                                \\ \hline
\multicolumn{1}{l|}{\textbf{Template}} & \textbf{Context}                                                               \\ \hline
\multicolumn{1}{l|}{\begin{tabular}[c]{@{}l@{}}\texttt{Dialogue:}\\ \texttt{User:} [U$_1$]\\ \texttt{KB:} [K$_1$]\\ \quad\\ \quad\\ \texttt{Assistant:} [S$_1$]\\ \texttt{User:} [U$_2$] \\ \texttt{KB:} [K$_2$] \\ \quad\\ \texttt{Assistant:}\end{tabular}} &
  \begin{tabular}[c]{@{}l@{}}\texttt{Dialogue:}\\ \texttt{User:} Blue is always nice. I like royal blue.\\ \texttt{KB:} The Royal Blue was the Baltimore and Ohio Railroad (B\&O)'s \\ \qquad flagship passenger train between New York City and Washington, \\ \qquad D.C., in the United States, beginning in 1890.\\ \texttt{Assistant:} I once road on The Royal Blue train from New York to D.C\\ \texttt{User:} Oh that sounds really nice. I bet there was a lot of scenery and blue skies.\\ \texttt{KB:} Blue Skies is a 1946 American musical comedy film directed by \\ \qquad Stuart Heisler and starring Bing Crosby, Fred Astaire, and Joan Caulfield.\\ \texttt{Assistant:}\end{tabular} \\ \hline
\multicolumn{2}{l}{\textbf{Generation}}                                                                                 \\ \hline
\multicolumn{1}{l|}{\hl{[S$_2$]}}      & Yes, speaking of Blue Skies, have you seen the 1946 movie staring Bing Crosby? \\ \hline
\end{tabular}%
}
\caption{Prompt and example for Wizard of Wikipedia (WoW).}\label{tab:WoW}
\end{table}
\begin{table}[t]
\centering
\resizebox{\textwidth}{!}{%
\begin{tabular}{ll}
\textit{WIT}                           &                                                                                \\ \hline
\multicolumn{1}{l|}{\textbf{Template}} & \textbf{Context}                                                               \\ \hline
\multicolumn{1}{l|}{\begin{tabular}[c]{@{}l@{}}\texttt{Assistant Information:} \\ $[$\texttt{P}$_1]$\\ $[$\texttt{P}$_2]$\\ \\ \texttt{Dialogue:} \\ \texttt{User:} [U$_1$]\\ \texttt{KB:} [K$_1$] \\ \texttt{Assistant:} [S$_1$] \\ \texttt{User:} [U$_2$]\\ \texttt{KB:} [K$_2$] \\ \quad\\ \texttt{Assistant:}\end{tabular}} &
  \begin{tabular}[c]{@{}l@{}}\texttt{Assistant Information:}\\ My favorite sports team is NY Giants.\\ When they are winning...\\ \\ \texttt{Dialogue:}\\ \texttt{User:} $\emptyset$\\ \texttt{KB:} $\emptyset$\\ \texttt{Assistant:} What do you think about Kenny Golladay for the Giants?\\ \texttt{User:} He is pretty good. I think he is a pretty good fit for the team.\\ \texttt{KB:} Golladay is under evaluation for a concussion during Sunday's \\ \qquad game versus the Packers.\\ \texttt{Assistant:}\end{tabular} \\ \hline
\multicolumn{2}{l}{\textbf{Generation}}                                                                                 \\ \hline
\multicolumn{1}{l|}{\hl{[S$_2$]}}      & He's kind of had a lot of head injuries though. Do you think that's a concern? \\ \hline
\end{tabular}%
}
\caption{Prompt and example for Wizard of Internet (WiT).}\label{tab:wit}
\end{table}
\begin{table}[t]
\centering
\begin{tabular}{ll|c}
\textit{IC}                            &                                                               \\ \hline
\multicolumn{1}{l|}{\textbf{Template}} & \textbf{Prompt}     & Original Image                                          \\ \hline
\multicolumn{1}{l|}{\begin{tabular}[c]{@{}l@{}}\texttt{Image:}\\ \texttt{[D]}\\ \texttt{Dialogue:}\\ \texttt{[StyleA]:} [U$_1$]\\ \texttt{[StyleB]:} [S$_1$]\\ \texttt{[StyleA]:} \end{tabular}} &
  \begin{tabular}[c]{@{}l@{}}\texttt{Image:}\\ a group of horses standing next to each other\\ \texttt{Dialogue:}\\ \texttt{Considerate:} The loads on their neck are too much\\ \texttt{Charming:} On my next date I am totally taking out \\\qquad\qquad \ \ my date on beautiful horses like these\\ \texttt{Considerate:}\end{tabular} & \includegraphics[width=0.2\textwidth]{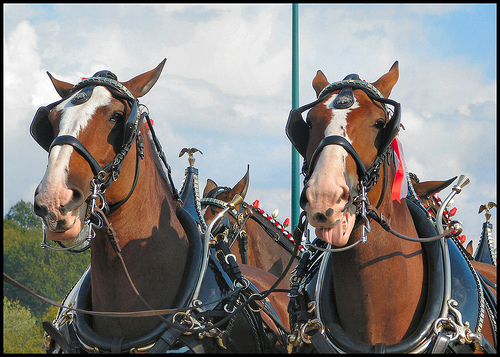} \\ \hline
\multicolumn{2}{l}{\textbf{Generation}} & \\ \hline
\multicolumn{1}{l|}{\hl{[U$_2$]}}      & A horse ride can be quite romantic. & \\ \hline
\end{tabular}
\caption{Prompt and example for Image Chat (IC).}
\label{tab:ic}
\end{table}
\begin{table}[t]
\centering
\begin{tabular}{ll}
\textit{SMD}                       &                                                            \\ \hline
\multicolumn{1}{l|}{\textbf{Template}} & \textbf{Prompt}                                           \\ \hline
\multicolumn{1}{l|}{\begin{tabular}[c]{@{}l@{}}\texttt{KB:}\\ $[$\texttt{T}$_1]\cdots[$\texttt{T}$_n]$\\ \texttt{Dialogue:}\\ \texttt{User:} [U$_1$]\\ \texttt{Assistant:} [S$_1$]\\\texttt{User:} [U$_2$]\\ \texttt{Assistant:}\end{tabular}} &
  \begin{tabular}[c]{@{}l@{}}\texttt{KB:}\\\texttt{chevron distance 5\_miles} \\
\texttt{chevron traffic\_info moderate\_traffic} \\
\texttt{chevron poi\_type gas\_station} \\
\texttt{chevron address 783\_arcadia\_pl} \\
\texttt{town\_and\_country distance 5\_miles} \\
\texttt{town\_and\_country traffic\_info no\_traffic} \\
\texttt{town\_and\_country poi\_type shopping\_center} \\
\texttt{town\_and\_country address 383\_university\_ave} \\
\texttt{jacks\_house distance 5\_miles} \\
\texttt{jacks\_house traffic\_info no\_traffic} \\
\texttt{jacks\_house poi\_type friends\_house} \\
\texttt{jacks\_house address 864\_almanor\_ln} \\
\texttt{home distance 6\_miles} \\
\texttt{home traffic\_info heavy\_traffic} \\
\texttt{home poi\_type home} \\
\texttt{home address 5671\_barringer\_street} \\
\texttt{the\_clement\_hotel distance 4\_miles} \\
\texttt{the\_clement\_hotel traffic\_info no\_traffic} \\
\texttt{the\_clement\_hotel poi\_type rest\_stop} \\
\texttt{the\_clement\_hotel address 657\_ames\_ave} \\
\texttt{sigona\_farmers\_market distance 1\_miles} \\
\texttt{sigona\_farmers\_market traffic\_info heavy\_traffic} \\
\texttt{sigona\_farmers\_market poi\_type grocery\_store} \\
\texttt{sigona\_farmers\_market address 638\_amherst\_st} \\
\texttt{tai\_pan distance 6\_miles} \\
\texttt{tai\_pan traffic\_info no\_traffic} \\
\texttt{tai\_pan poi\_type chinese\_restaurant} \\
\texttt{tai\_pan address 830\_almanor\_ln} \\ \texttt{Dialogue:}\\ \texttt{User:} what gas\_station are here ?\\ \texttt{Assistant:} there is a chevron\\\texttt{User:} that s good ! please pick the quickest route to \\ \qquad \quad get there and avoid all heavy\_traffic !\\ \texttt{Assistant:}\end{tabular} \\ \hline
\multicolumn{2}{l}{\textbf{Generation}}                                                             \\ \hline
\multicolumn{1}{l|}{\hl{[S$_2$]}}      & taking you to chevron \\ \hline
\end{tabular}
\caption{Prompt and example for Standford Multidomain Dataset (SMD).}
\label{tab:smd}
\end{table}
\begin{table}[t]
\centering
\begin{tabular}{ll}
\textit{MSC}                       &                                                            \\ \hline
\multicolumn{1}{l|}{\textbf{Template}} & \textbf{Prompt}                                           \\ \hline
\multicolumn{1}{l|}{\begin{tabular}[c]{@{}l@{}}\texttt{User Information:}\\ $[$\texttt{P}$_1]$\\ $[$\texttt{P}$_2]$\\ $[$\texttt{P}$_3]$\\ $[$\texttt{P}$_4]$\\ \texttt{Assistant Information:}\\ $[$\texttt{P}$_1]$\\ $[$\texttt{P}$_2]$\\ $[$\texttt{P}$_3]$\\ $[$\texttt{P}$_4]$\\ \\ \texttt{Dialogue:} \\ \texttt{User:} [U$_1$] \\ \texttt{Assistant:} [S$_1$]\\ \texttt{User:} [U$_2$] \\ \texttt{Assistant:}\end{tabular}} &
  \begin{tabular}[c]{@{}l@{}}\texttt{User Information:}\\ I own a Jeep.\\ I enjoy exercising at the gym.\\ I have a marketing job.\\ I don't eat meat.\\ \texttt{Assistant Information:}\\ I don't own a car.\\ I'm a computer programmer.\\ I like grilling steak.\\ I enjoy running and walking.\\ \\ \texttt{Dialogue:}\\ \texttt{User:} Did you manage to go out on a run today?\\ \texttt{Assistant:} Yes I actually was able too. I am \\\qquad\qquad\qquad considering joining the local gym. \\\qquad\qquad\qquad Do you prefer going to the gym?\\ \texttt{User:} I do actually. I like the controlled environment. \\ \qquad\quad I don't want to have to depend on the weather \\\qquad\quad consider where I live. \\ \texttt{Assistant:}\end{tabular} \\ \hline
\multicolumn{2}{l}{\textbf{Generation}}                                                             \\ \hline
\multicolumn{1}{l|}{\hl{[S$_2$]}}      & I hate to have to run when its raining. \\ \hline
\end{tabular}
\caption{Prompt and example for Multi-Session Chat (MSC).}
\label{tab:msc}
\end{table}

\begin{table}[t]
\centering
\begin{tabular}{ll}
\textit{CG(IC)}                       &                                                            \\ \hline
\multicolumn{1}{l|}{\textbf{Template}} & \textbf{Prompt}                                           \\ \hline
\multicolumn{1}{l|}{\begin{tabular}[c]{@{}l@{}}\texttt{Dialogue:}\\ \texttt{System:} [U$_1$]\\ \texttt{User:} [S$_1$]\\ \texttt{[Style]:}\end{tabular}} &
  \begin{tabular}[c]{@{}l@{}}\texttt{Dialogue:}\\ \texttt{System:} What a sweet sky.\\ \texttt{User:} Beautiful day out in the cold\\ \texttt{Kind:}\end{tabular} \\ \hline
\multicolumn{2}{l}{\textbf{Generation}}                                                             \\ \hline
\multicolumn{1}{l|}{\hl{[U$_2$]}}      & It makes me incredibly peaceful to see that sky \\ \hline
\end{tabular}
\caption{Prompt and example for Controlled Generation (CG-IC).}
\label{tab:CGIC}
\end{table}

\begin{table}[t]
\centering
\resizebox{\textwidth}{!}{%
\begin{tabular}{ll}
\textit{DialKG}                     &                                                                                \\ \hline
\multicolumn{1}{l|}{\textbf{Template}} & \textbf{Context}                                                               \\ \hline
\multicolumn{1}{l|}{\begin{tabular}[c]{@{}l@{}}\texttt{Dialogue:}\\ \texttt{User:} [U$_1$]\\ \texttt{KG:} [T$_1$]\\ \texttt{Assistant:} [S$_1$] \\ \qquad\qquad \\ \texttt{User:} [U$_2$] \\ \qquad\quad  \\ \texttt{KG:} [T$_2$] \\ \texttt{Assistant:}\end{tabular}} &
  \begin{tabular}[c]{@{}l@{}}\texttt{Dialogue:}\\ \texttt{User:} I like Anita Diamant. Could you recommend anything by her?\\ \texttt{KG:} Anita Diamant \texttt{\textbackslash t} $\sim$written\_by \texttt{\textbackslash t} The Red Tent\\ \texttt{Assistant:} She wrote The Red Tent and The Last Days of Dogtown, \\ \qquad\qquad both great novels.\\ \texttt{User:} They were great. I read both of those. Red Tent is historical fiction. \\ \qquad\quad Do you know if she has any other books like that?\\ \texttt{KG:} Historical fiction \texttt{\textbackslash t} $\sim$has\_genre \texttt{\textbackslash t} The Weight of Water\\ \texttt{Assistant:}\end{tabular} \\ \hline
\multicolumn{2}{l}{\textbf{Generation}}                                                                                 \\ \hline
\multicolumn{1}{l|}{\hl{[S$_2$]}}      & The Weight of Water is a great book but it is not written by Anita Diamant. \\ \hline
\end{tabular}%
}
\caption{Prompt and example for DialKG.}\label{tab:dialKG}
\end{table}
\begin{table}[t]
\centering
\resizebox{\textwidth}{!}{%
\begin{tabular}{ll}
\textit{WoW-Parse}                     &                                                                                \\ \hline
\multicolumn{1}{l|}{\textbf{Template}} & \textbf{Prompt}                                                               \\ \hline
\multicolumn{1}{l|}{\begin{tabular}[c]{@{}l@{}}\texttt{Dialogue:}\\ \texttt{User:} [U$_1$]\\ \texttt{Assistant:} [S$_1$]\\ \texttt{User:} [U$_2$] \\ \texttt{Search:}\end{tabular}} &
  \begin{tabular}[c]{@{}l@{}}\texttt{Dialogue:}\\ \texttt{User:} I go to Target all the time, do you?\\ \texttt{Assistant:} Target Corporation is the second largest discount retailer\\ \texttt{User:} I assume that it is just before Walmart, but still huge.\\ \texttt{Search:}\end{tabular} \\ \hline
\multicolumn{2}{l}{\textbf{Generation}}                                                                                 \\ \hline
\multicolumn{1}{l|}{\hl{[Q]}}      & Target Corporation \\ \hline
\end{tabular}%
}
\caption{Prompt and example for Wizard of Wikipedia Parse (WoW-Parse).}\label{tab:WoWparse}
\end{table}
\begin{table}[t]
\centering
\resizebox{\textwidth}{!}{%
\begin{tabular}{ll}
\textit{WIT-parse}                           &                                                                                \\ \hline
\multicolumn{1}{l|}{\textbf{Template}} & \textbf{Context}                                                               \\ \hline
\multicolumn{1}{l|}{\begin{tabular}[c]{@{}l@{}}\texttt{Assistant Information:} \\ $[$\texttt{P}$_1]$\\ $[$\texttt{P}$_2]$\\ \\ \texttt{Dialogue:} \\ \texttt{User:} [U$_1$]\\ \texttt{Search:} [Q$_1$] \\ \texttt{Assistant:} [S$_1$] \\ \texttt{User:} [U$_2$]\\ \texttt{Search:} \end{tabular}} &
  \begin{tabular}[c]{@{}l@{}}\texttt{Assistant Information:}\\ My favorite sports team is NY Giants.\\ When they are winning...\\ \\ \texttt{Dialogue:}\\ \texttt{User:} $\emptyset$\\ \texttt{Search:} Kenny Golladay Giants\\ \texttt{Assistant:} What do you think about Kenny Golladay for the Giants?\\ \texttt{User:} He is pretty good. I think he is a pretty good fit for the team.\\ \texttt{Search:} \end{tabular} \\ \hline
\multicolumn{2}{l}{\textbf{Generation}}                                                                                 \\ \hline
\multicolumn{1}{l|}{\hl{[Q$_2$]}}      & Kenny Golladay Rumors: Giant \\ \hline
\end{tabular}%
}
\caption{Prompt and example for Wizard of Internet Parse (WiT-parse).}\label{tab:witparse}
\end{table}
\begin{table}[t]
\centering
\begin{tabular}{ll}
\textit{MSC-parse}                       &                                                            \\ \hline
\multicolumn{1}{l|}{\textbf{Template}} & \textbf{Prompt}                                           \\ \hline
\multicolumn{1}{l|}{\begin{tabular}[c]{@{}l@{}} \texttt{Dialogue:} \\ \texttt{User:} [U$_1$] \\ \texttt{Write:} [P$_1$] \\ \texttt{Assistant:} [S$_1$]\\\qquad\\\qquad\\ \texttt{User:} [U$_2$]\\ \texttt{Write:}\end{tabular}} &
  \begin{tabular}[c]{@{}l@{}}\texttt{Dialogue:}\\ \texttt{User:} What kind of car do you own? I have a jeep.  \\ \texttt{Write:} I own a Jeep. \\ \texttt{Assistant:} I don't own my own car!  I actually really enjoying \\\qquad\qquad\qquad enjoying walking and running, but then again, I live \\\qquad\qquad\qquad in a small town and semi-close to work.\\ \texttt{User:} Ah I see! I like going to the gym to work out. \\ \texttt{Write:}\end{tabular} \\ \hline
\multicolumn{2}{l}{\textbf{Generation}}                                                             \\ \hline
\multicolumn{1}{l|}{\hl{[P$_2$]}}      & I enjoy exercising at the gym. \\ \hline
\end{tabular}
\caption{Prompt and example for Multi-Session Chat (MSC) parse.}
\label{tab:msc_parse}
\end{table}

\begin{table}[t]
\centering
\resizebox{\textwidth}{!}{%
\begin{tabular}{ll}
\textit{DialKG-Parse}                     &                                                                                \\ \hline
\multicolumn{1}{l|}{\textbf{Template}} & \textbf{Context}                                                               \\ \hline
\multicolumn{1}{l|}{\begin{tabular}[c]{@{}l@{}}\texttt{Dialogue:}\\ \texttt{User:} [U$_1$]\\ \texttt{KG:} [T$_1$]\\ \texttt{Assistant:} [S$_1$] \\ \qquad\qquad \\ \texttt{User:} [U$_2$] \\ \qquad\quad  \\ \texttt{KG:}\end{tabular}} &
  \begin{tabular}[c]{@{}l@{}}\texttt{Dialogue:}\\ \texttt{User:} I like Anita Diamant. Could you recommend anything by her?\\ \texttt{KG:} Anita Diamant \texttt{\textbackslash t} $\sim$written\_by \texttt{\textbackslash t} The Red Tent\\ \texttt{Assistant:} She wrote The Red Tent and The Last Days of Dogtown, \\ \qquad\qquad both great novels.\\ \texttt{User:} They were great. I read both of those. Red Tent is historical fiction. \\ \qquad\quad Do you know if she has any other books like that?\\ \texttt{KG:} \end{tabular} \\ \hline
\multicolumn{2}{l}{\textbf{Generation}}                                                                                 \\ \hline
\multicolumn{1}{l|}{\hl{[T$_2$]}}      & Historical fiction \texttt{\textbackslash t} $\sim$has\_genre \texttt{\textbackslash t} The Weight of Water \\ \hline
\end{tabular}%
}
\caption{Prompt and example for DialKG parse.}\label{tab:dialKG_parse}
\end{table}
\begin{table}[t]
\centering
\resizebox{\textwidth}{!}{%
\begin{tabular}{ll}
\textit{MWoZ DST (Single)}                     &                                                                                \\ \hline
\multicolumn{1}{l|}{\textbf{Template}} & \textbf{Context}                                                               \\ \hline
\multicolumn{1}{l|}{\begin{tabular}[c]{@{}l@{}}\texttt{Dialogue:}\\ \texttt{User:} [U$_1$]\\ \texttt{DST:} [DS$_1$]\\ \texttt{Assistant:} [S$_1$] \\ \qquad\qquad \\ \texttt{User:} [U$_2$] \\ \qquad\quad  \\ \texttt{DS:}\end{tabular}} &
  \begin{tabular}[c]{@{}l@{}}\texttt{Dialogue:}\\ \texttt{User:} i am looking for a hotel with at least a 3 star rating.\\ \texttt{DST} hotel-stars=3\\ \texttt{Assistant:} we have 6 hotel -s and guesthouses that have 3 star rating -s . \\ \qquad\qquad is there a particular part of town , or a particular price range that you are interested in ?\\ \texttt{User:} i am not to worried about price but the hotel needs to have free wifi.\\ \texttt{DST:} \end{tabular} \\ \hline
\multicolumn{2}{l}{\textbf{Generation}}                                                                                 \\ \hline
\multicolumn{1}{l|}{\hl{[DS$_2$]}}      & hotel-internet=yes \\ \hline
\end{tabular}%
}
\caption{Prompt and example for MWoZ DST.}\label{tab:mwozdst}
\end{table}
\begin{figure}[t]
    \centering
    \includegraphics[width=\linewidth]{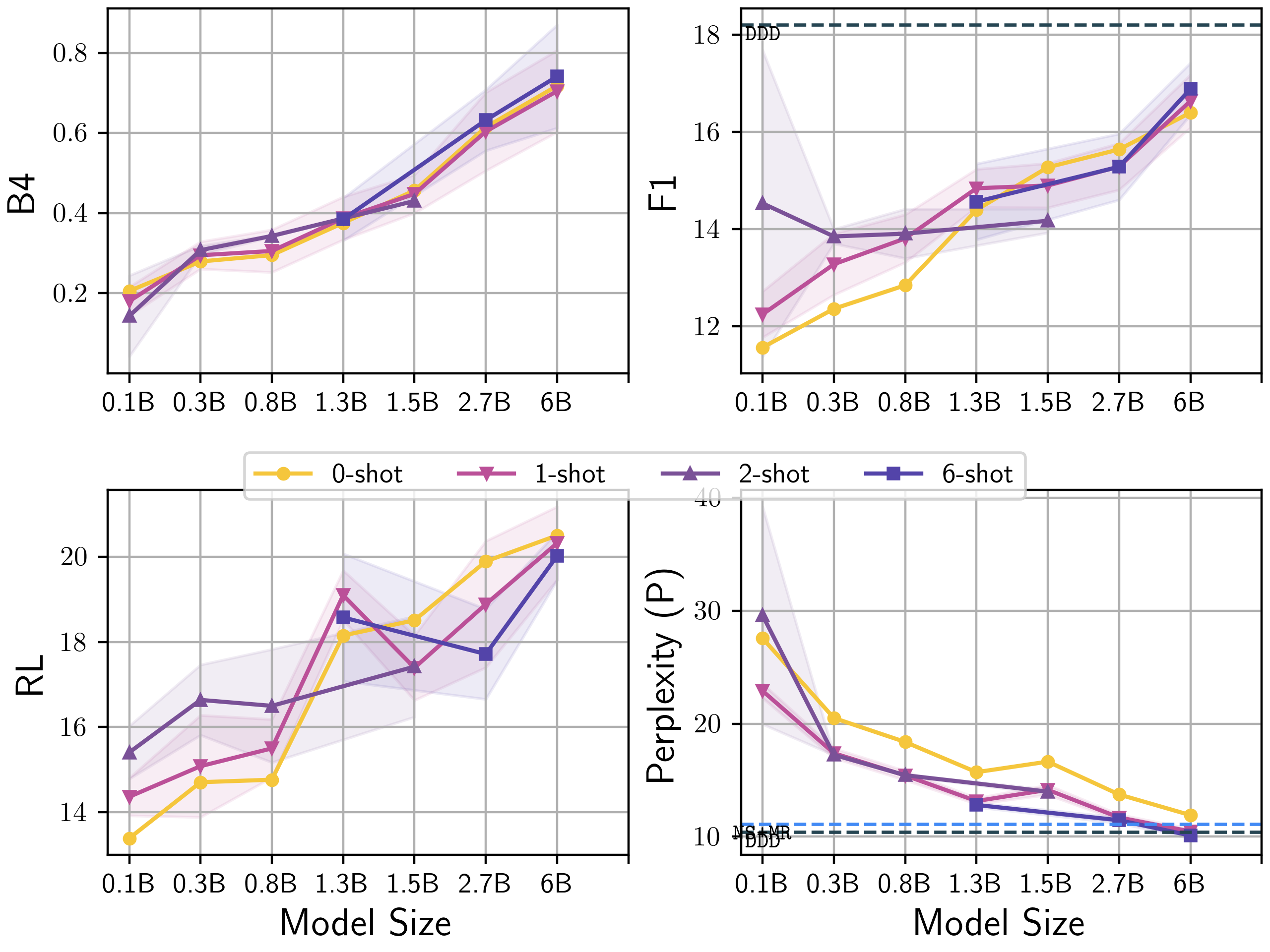}
    \caption{Model size vs metrics in the DailyDialogue (DD) dataset.}
    \label{fig:dd_size}
\end{figure}

\begin{figure}[t]
    \centering
    \includegraphics[width=\linewidth]{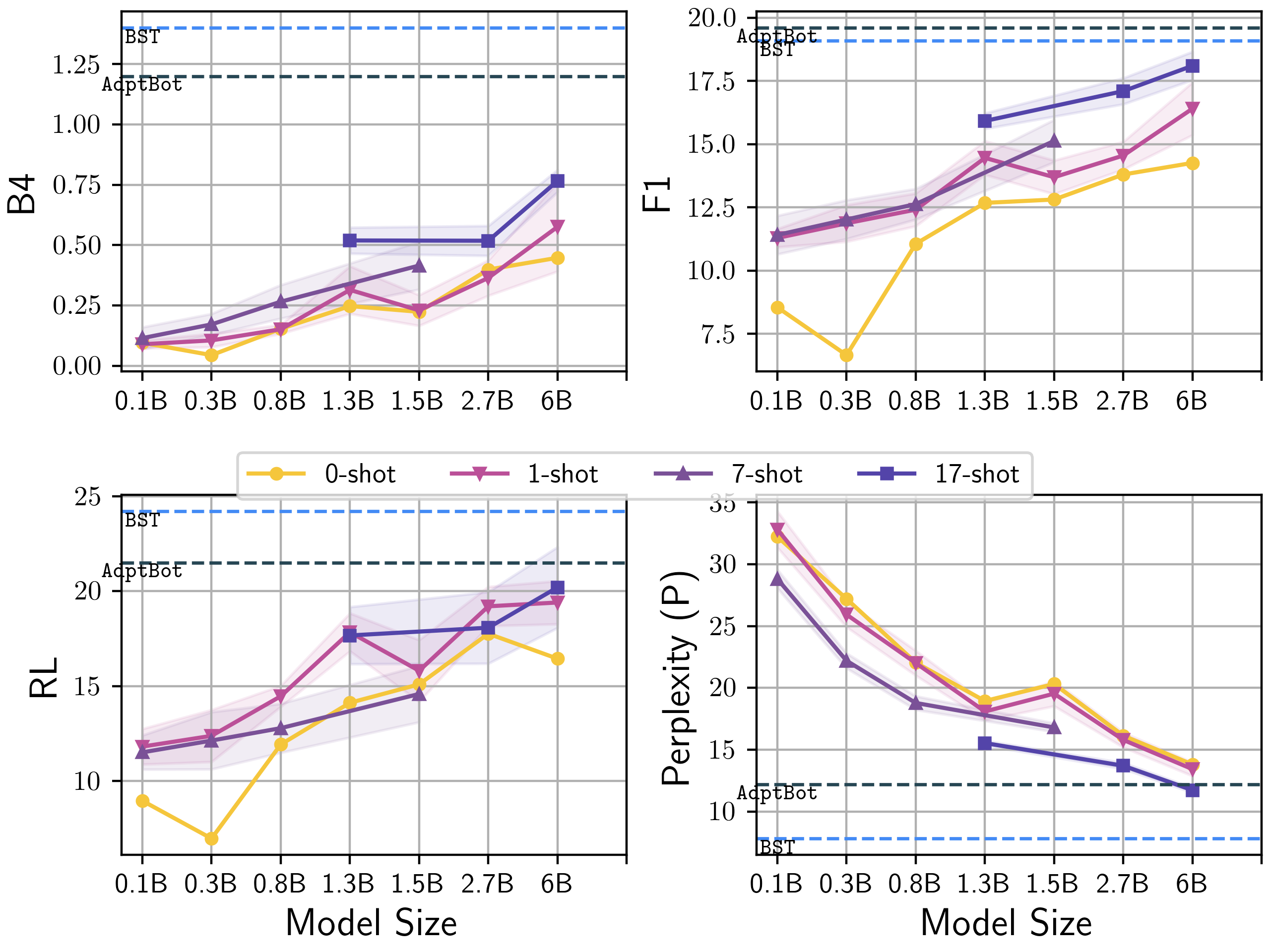}
    \caption{Model size vs metrics in the Empathetic Dialogues (ED) dataset.}
    \label{fig:ed_size}
\end{figure}

\begin{figure}[t]
    \centering
    \includegraphics[width=\linewidth]{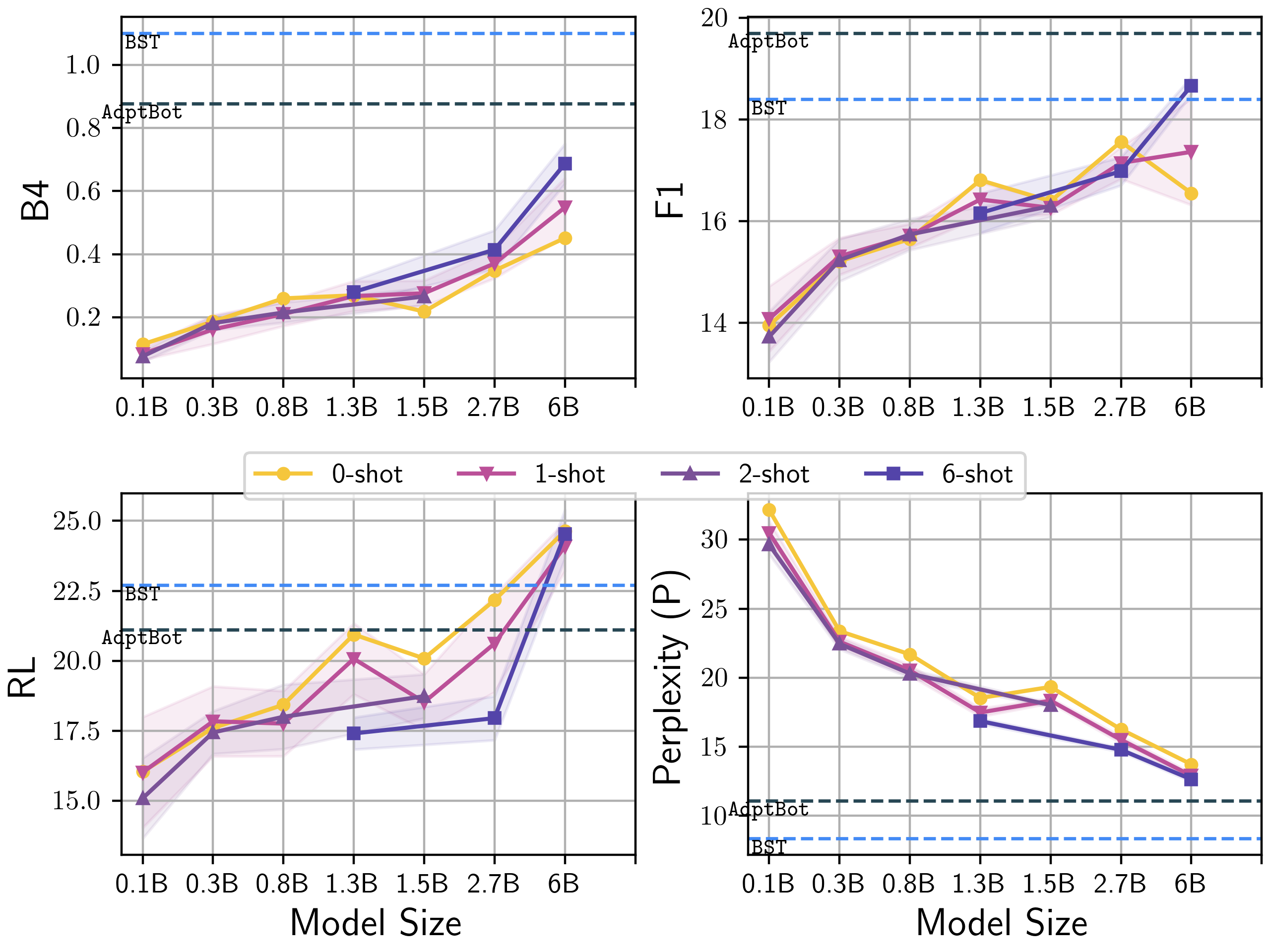}
    \caption{Model size vs metrics in the Persona Chat dataset.}
    \label{fig:persona_size}
\end{figure}

\begin{figure}[t]
    \centering
    \includegraphics[width=\linewidth]{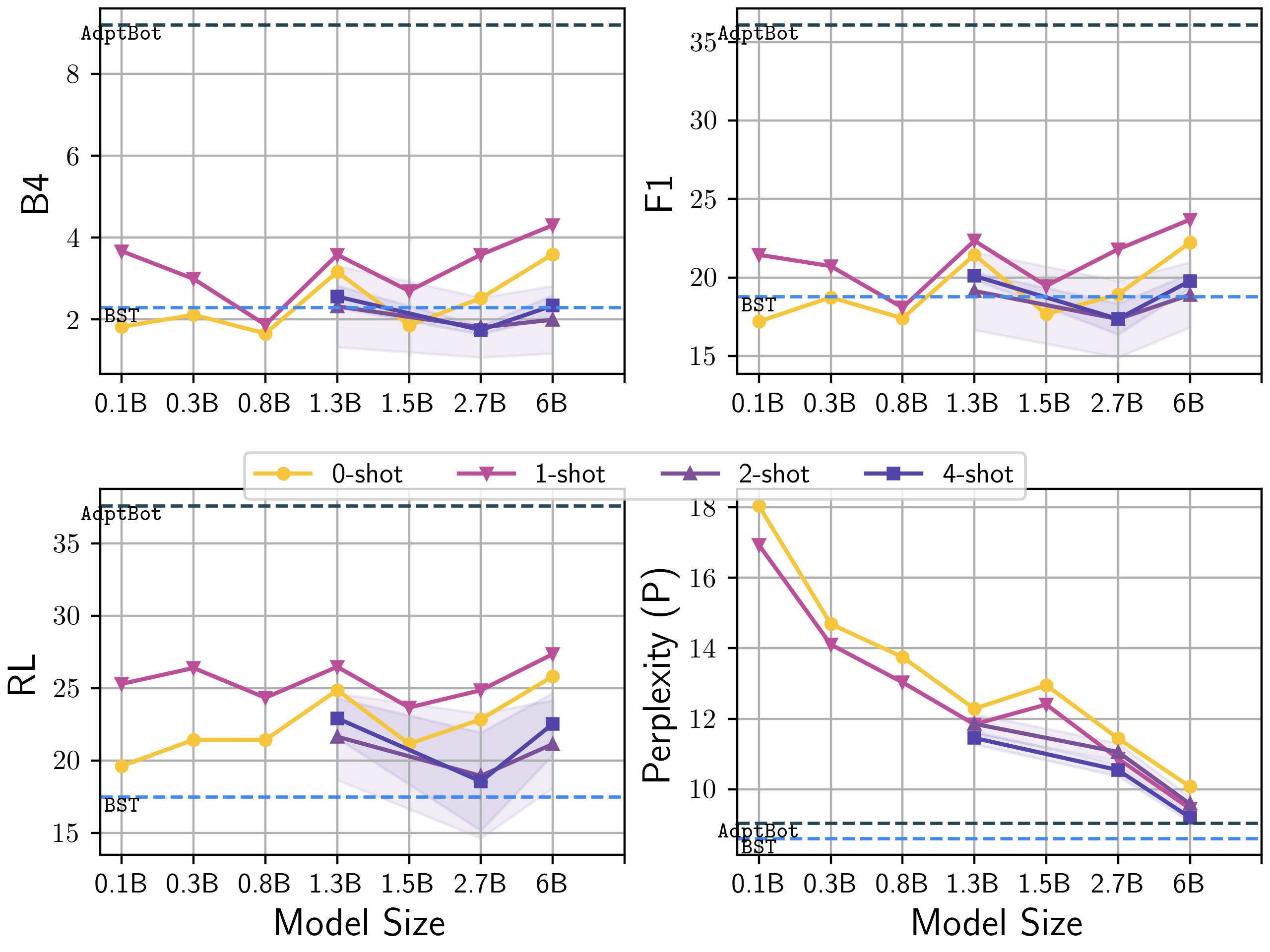}
    \caption{Model size vs metrics in the Wizard of Wikipedia (WoW) dataset.}
    \label{fig:wow_size}
\end{figure}

\begin{figure}[t]
    \centering
    \includegraphics[width=\linewidth]{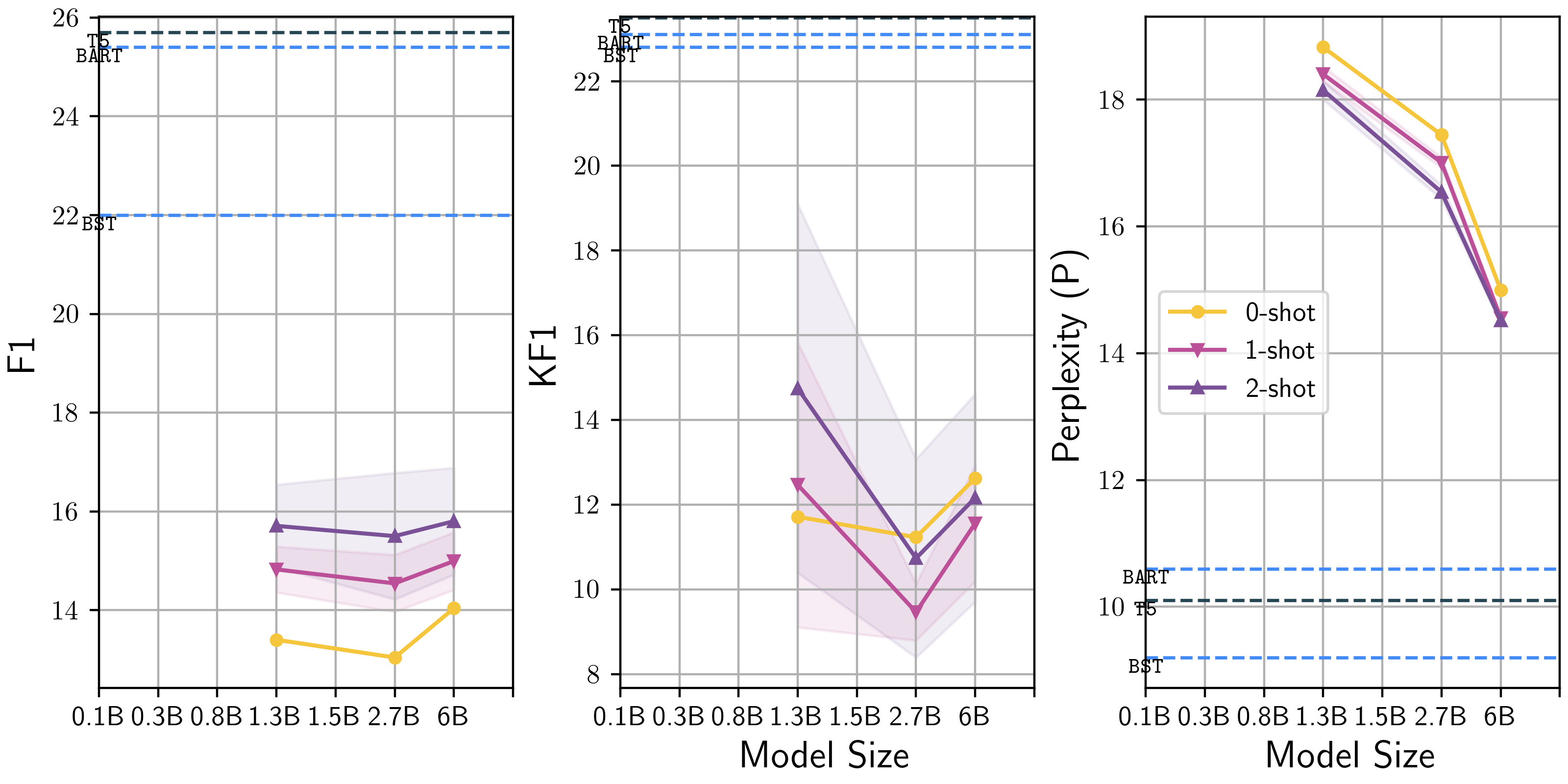}
    \caption{Model size vs metrics in the Wizard of Internet (WiT) dataset.}
    \label{fig:wit_size}
\end{figure}

\begin{figure}[t]
    \centering
    \includegraphics[width=\linewidth]{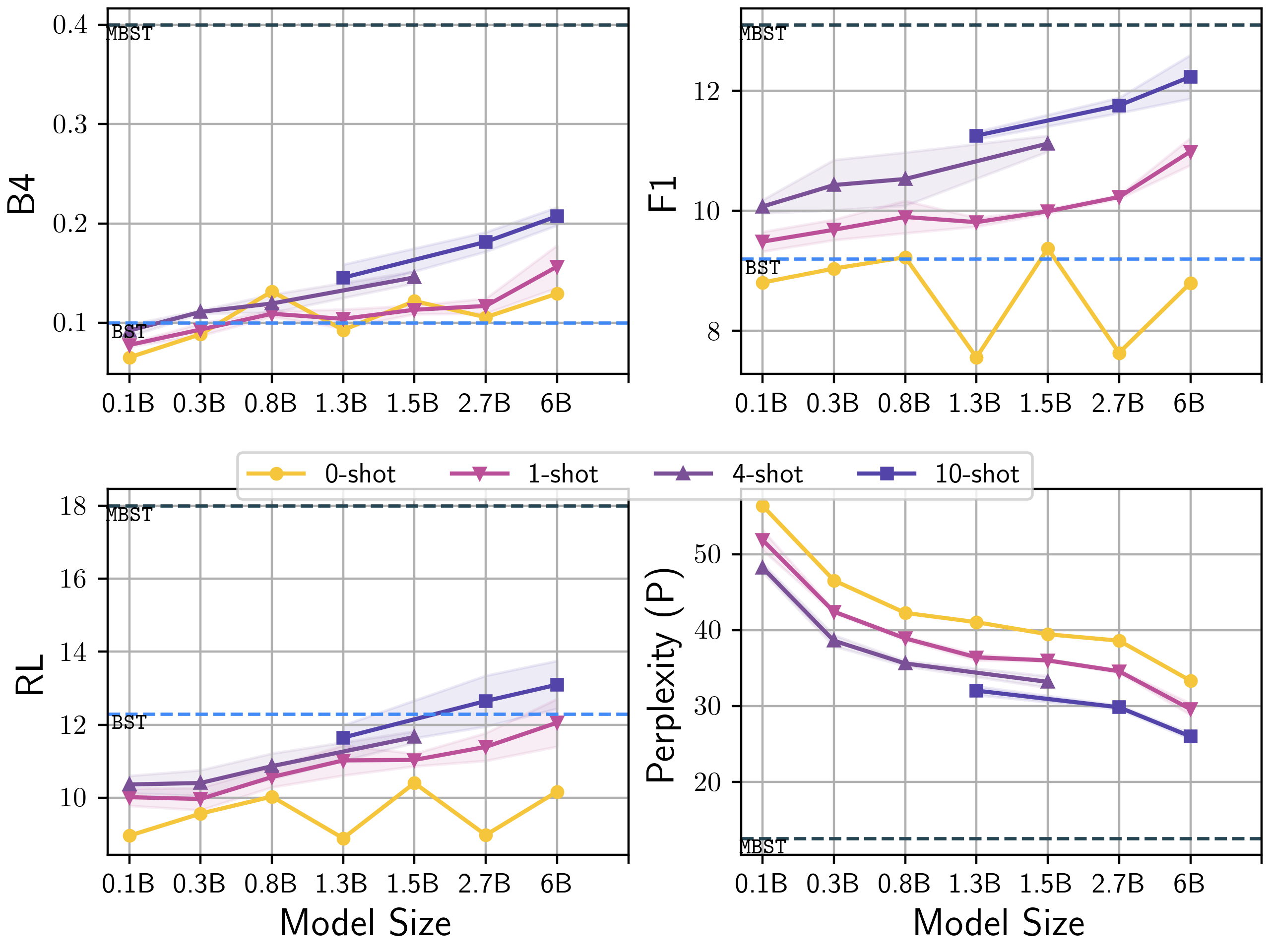}
    \caption{Model size vs metrics in the Image Chat (IC) dataset.}
    \label{fig:ic_size}
\end{figure}

\begin{figure}[t]
    \centering
    \includegraphics[width=\linewidth]{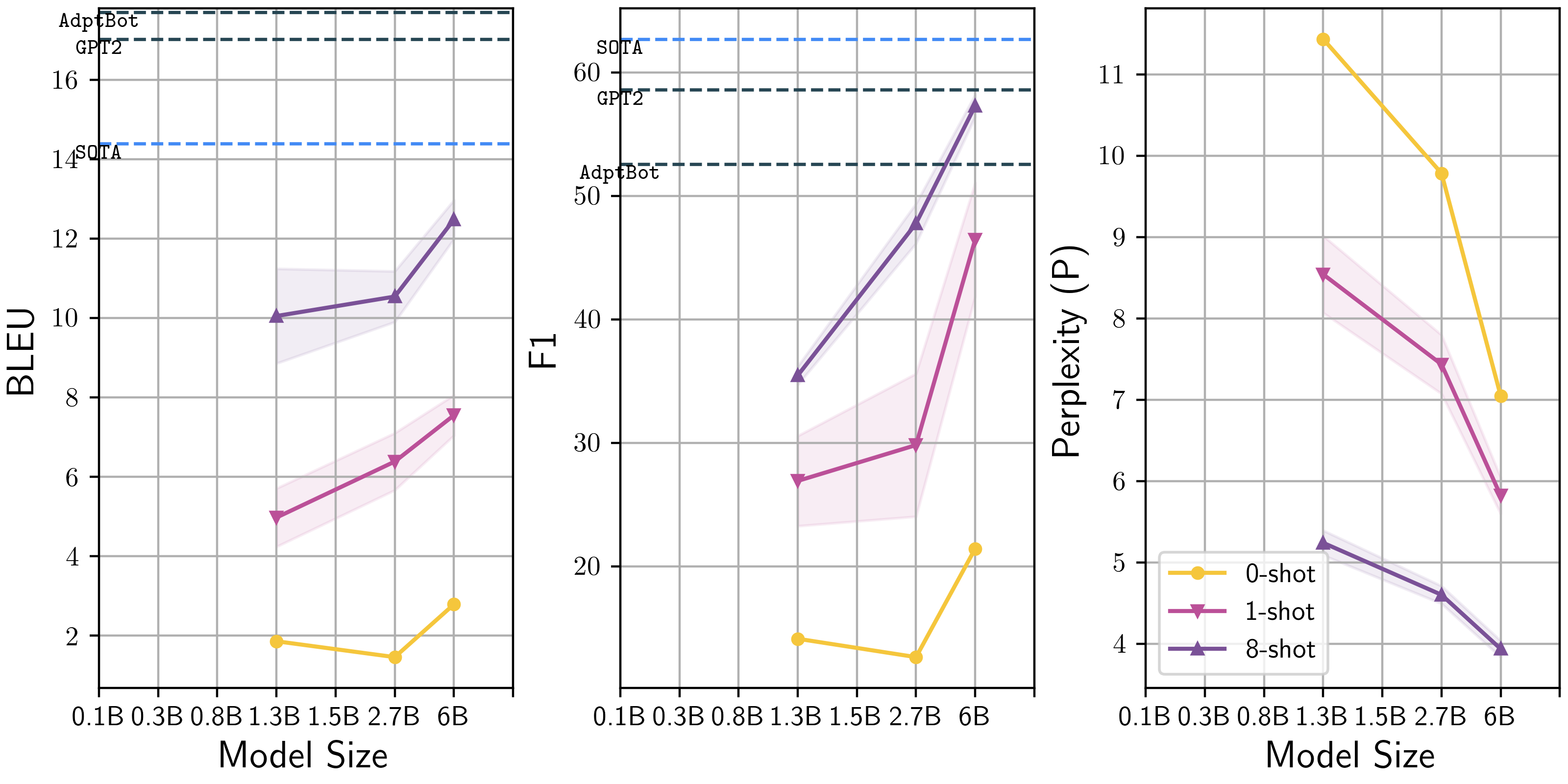}
    \includegraphics[width=\linewidth]{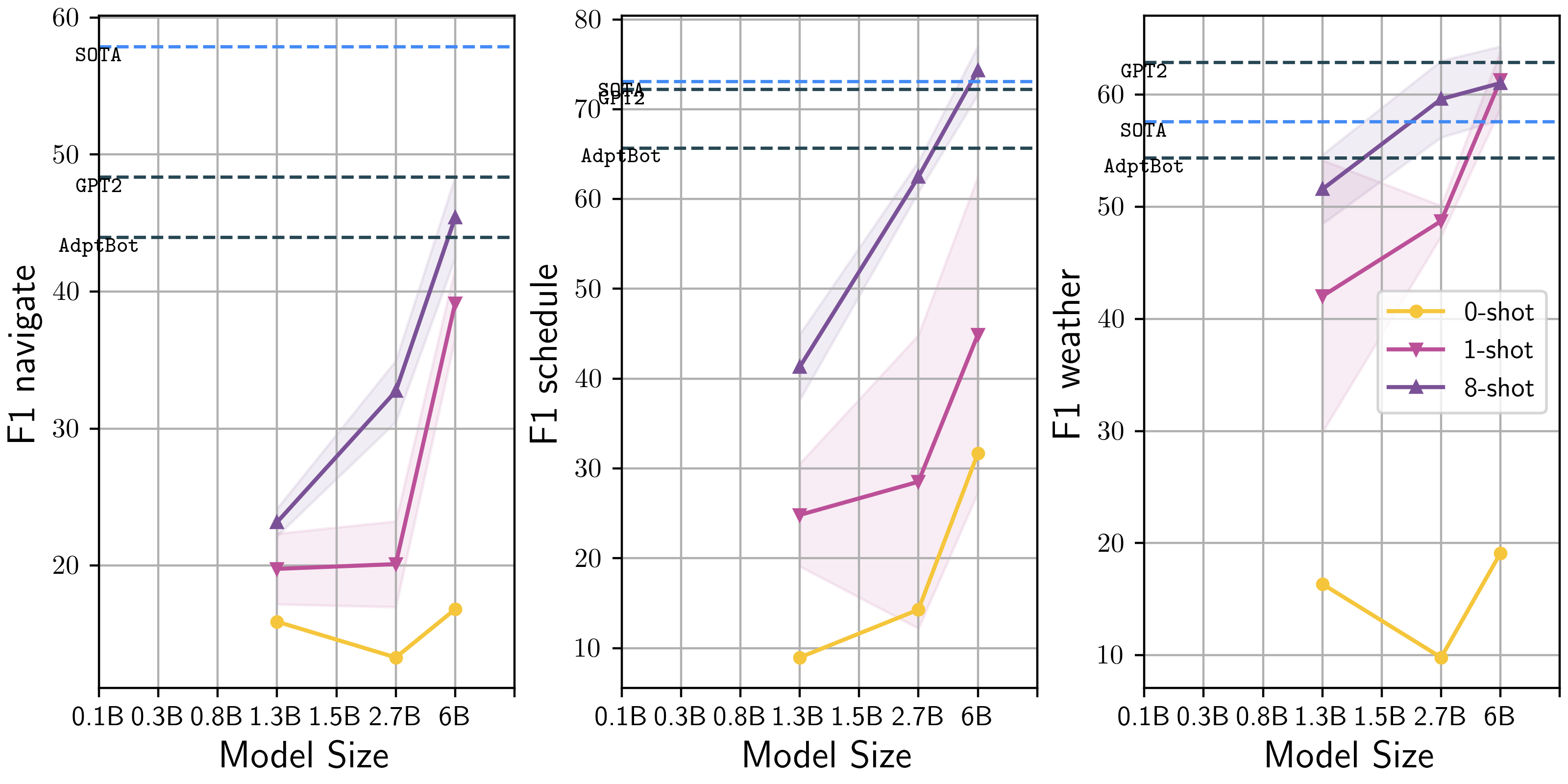}
    \caption{Model size vs metrics in the Stanford Multidomain Dialogue (SMD) dataset.}
    \label{fig:smd_size}
\end{figure}

\begin{figure}[t]
    \centering
    \includegraphics[width=\linewidth]{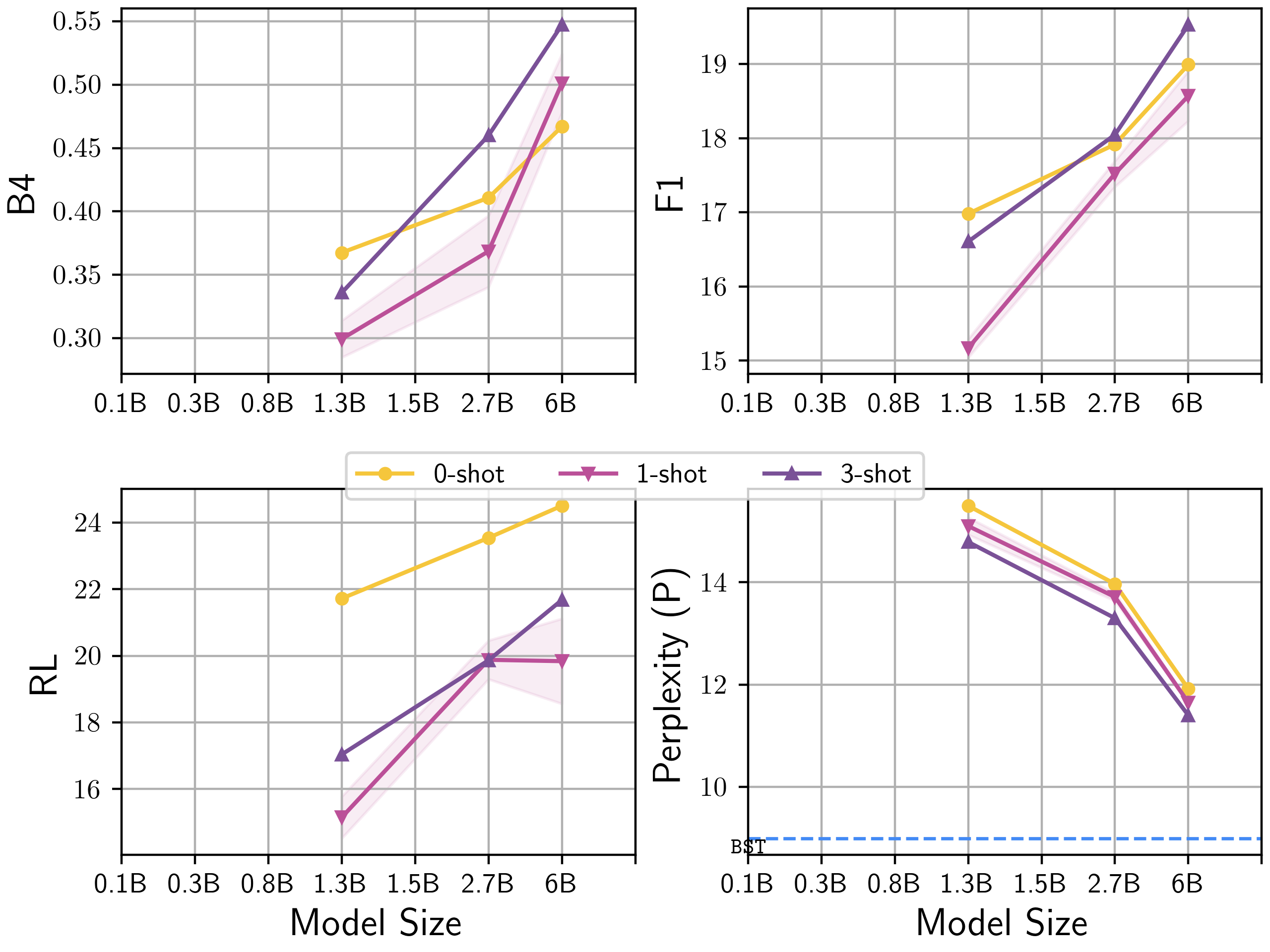}
    \caption{Model size vs metrics in the Multi-Session Chat (MSC) dataset.}
    \label{fig:msc_size}
\end{figure}

\begin{figure}[t]
    \centering
    \includegraphics[width=\linewidth]{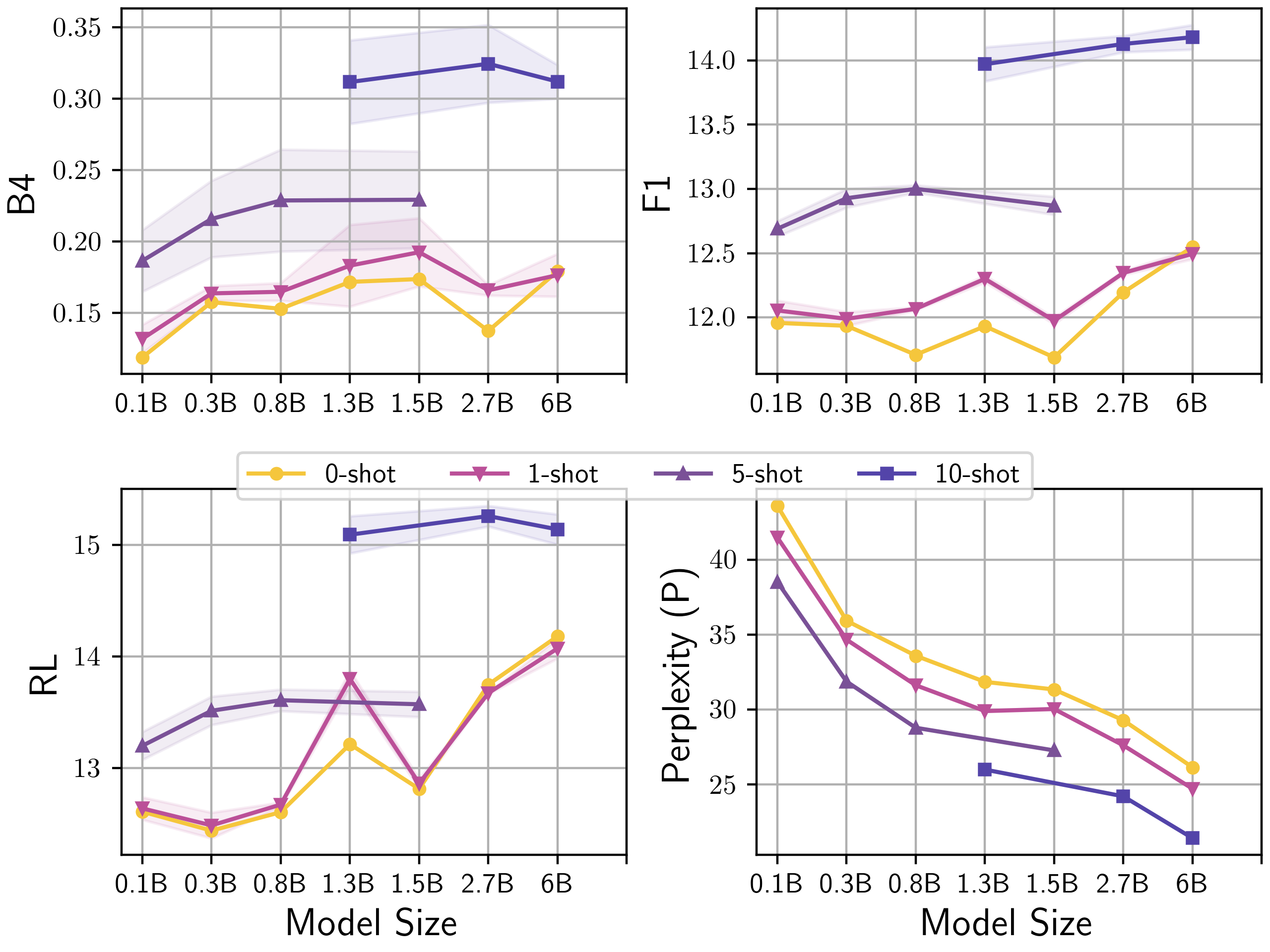}
    \caption{Model size vs metrics in the Controlled generation (CG-IC) dataset.}
    \label{fig:CGIC_size}
\end{figure}

\begin{figure}[t]
    \centering
    \includegraphics[width=\linewidth]{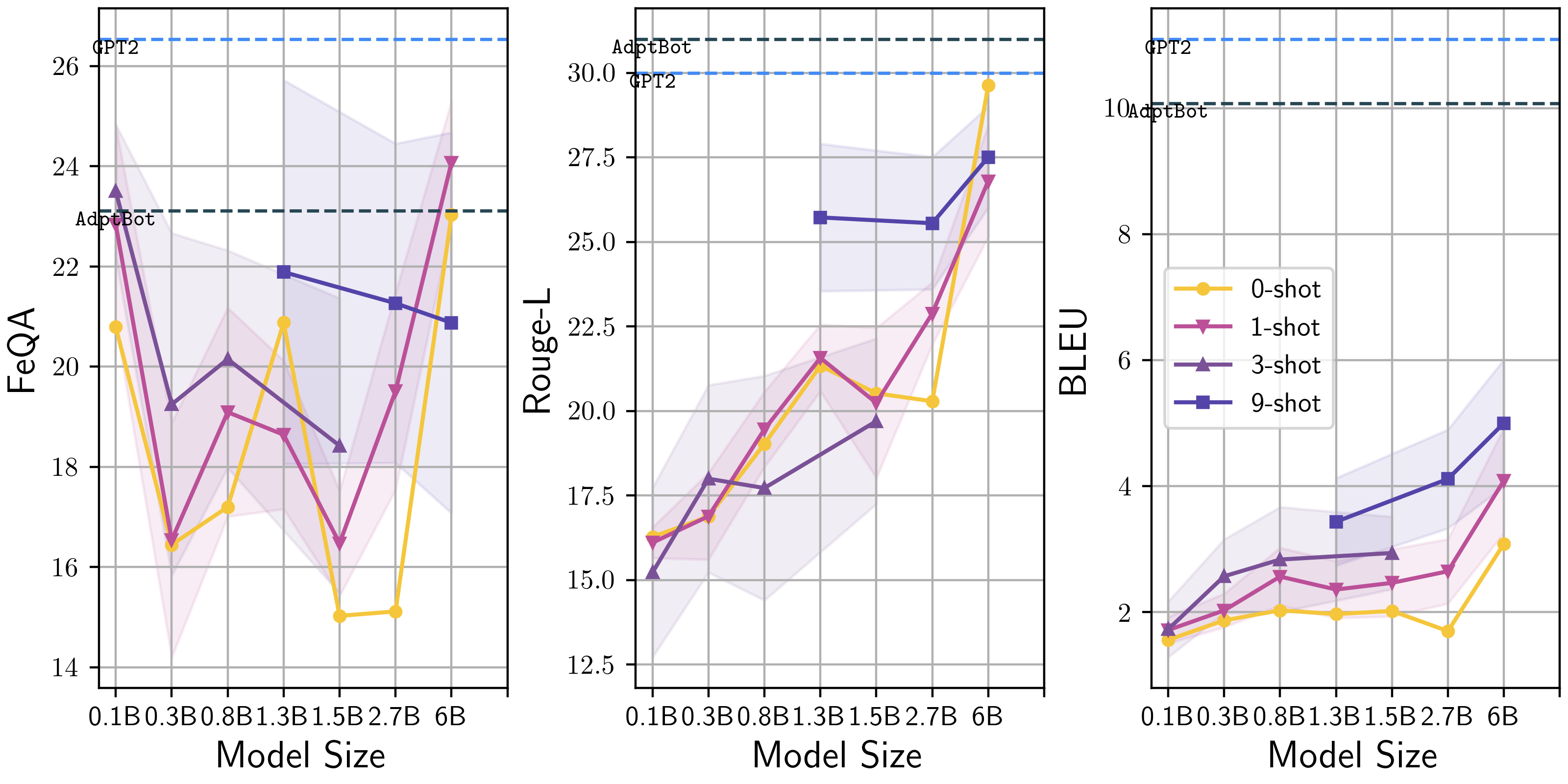}
    \caption{Model size vs metrics in the DialKG dataset.}
    \label{fig:dialKG_size}
\end{figure}

\begin{figure}[t]
    \centering
    \includegraphics[width=\linewidth]{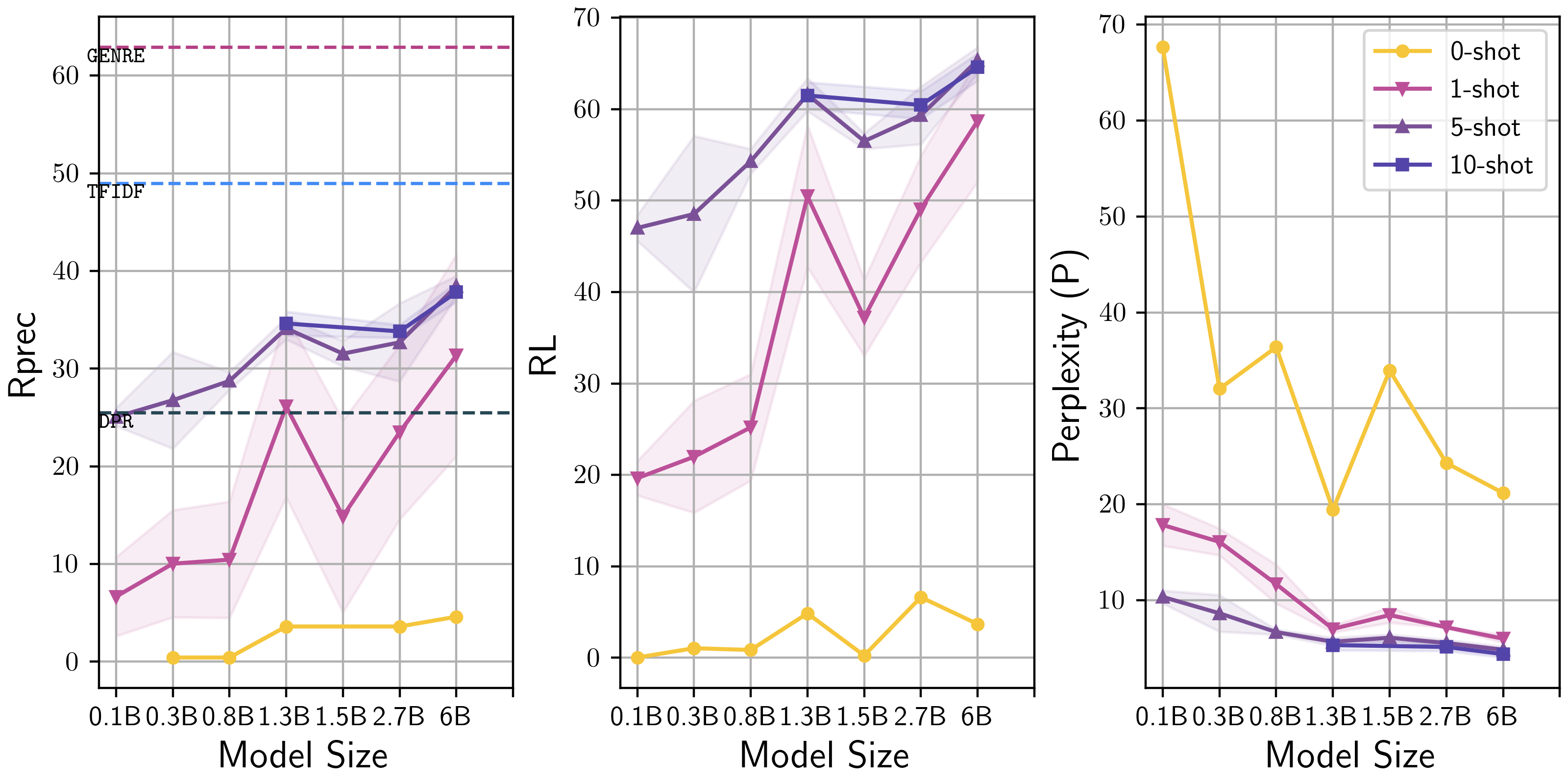}
    \caption{Model size vs metrics in the Wizard of Wikipedia Parse (WoW-Parse) dataset.}
    \label{fig:wow_parse_size}
\end{figure}

\begin{figure}[t]
    \centering
    \includegraphics[width=\linewidth]{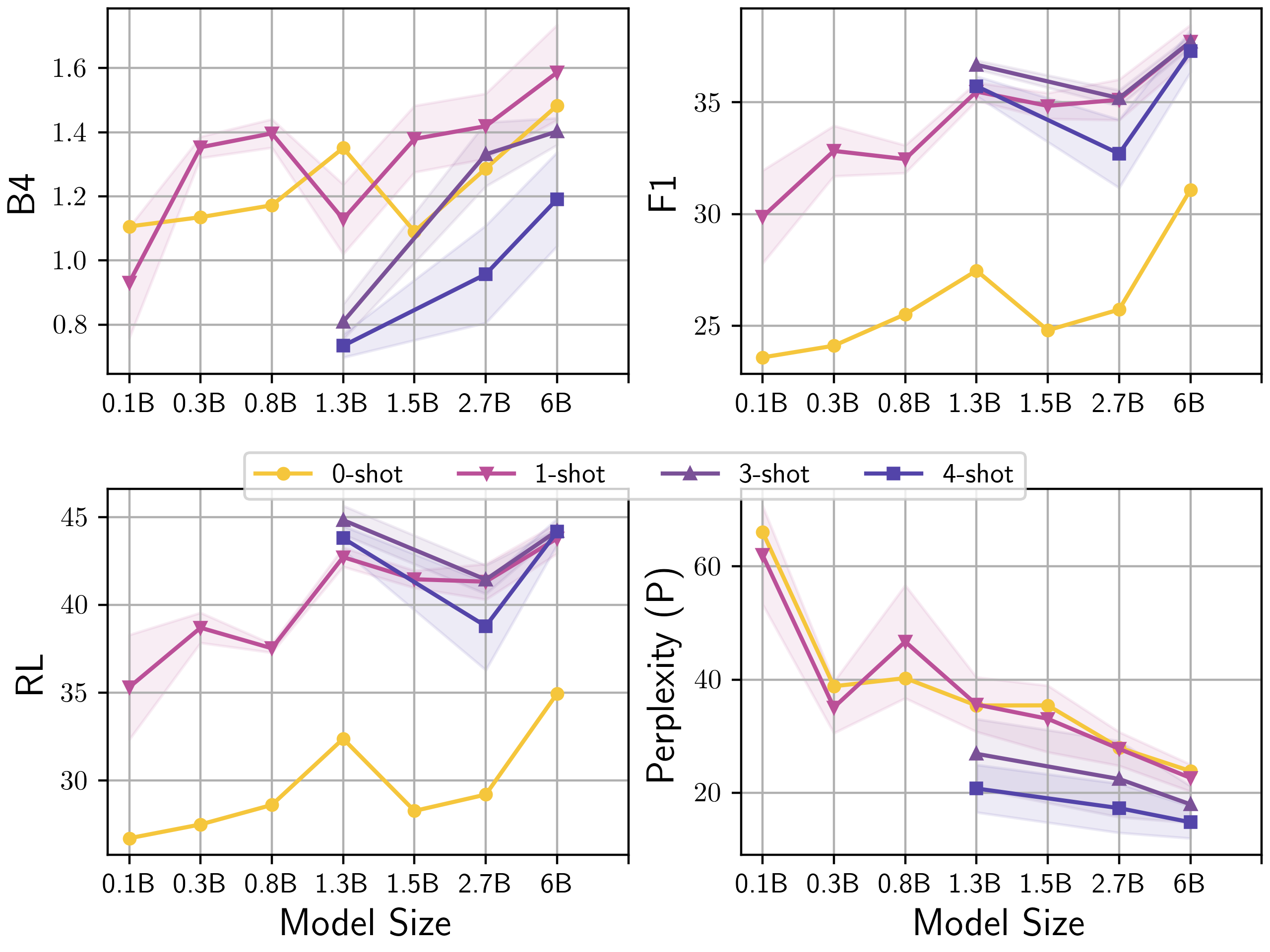}
    \caption{Model size vs metrics in the Wizard of Internet Parse (WiT-Parse) dataset.}
    \label{fig:wit_parse_size}
\end{figure}

\begin{figure}[t]
    \centering
    \includegraphics[width=\linewidth]{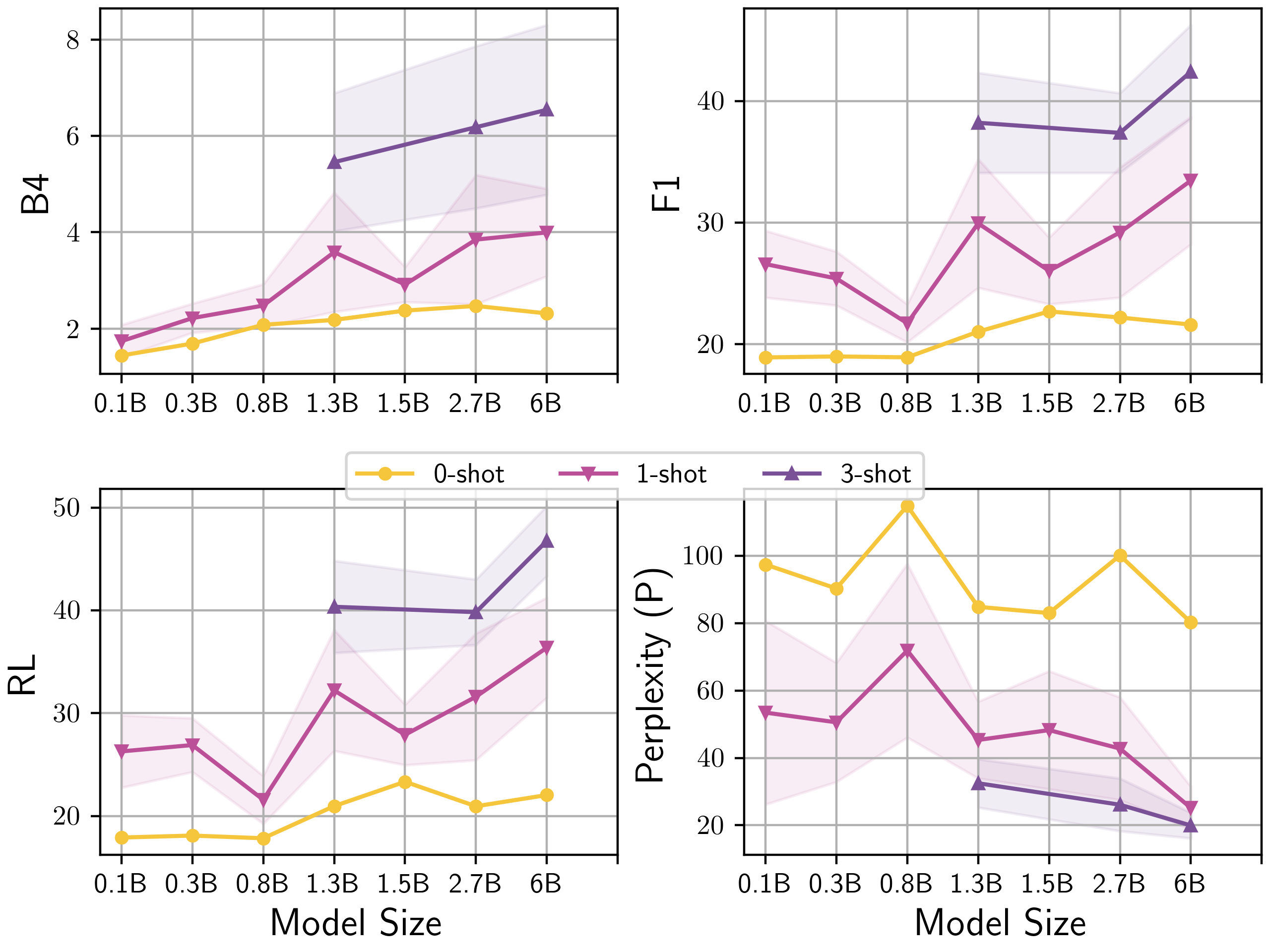}
    \caption{Model size vs metrics in the Multi-Session Chat Parse (MSC-Parse) dataset.}
    \label{fig:msc_parse_size}
\end{figure}

\begin{figure}[t]
    \centering
    \includegraphics[width=\linewidth]{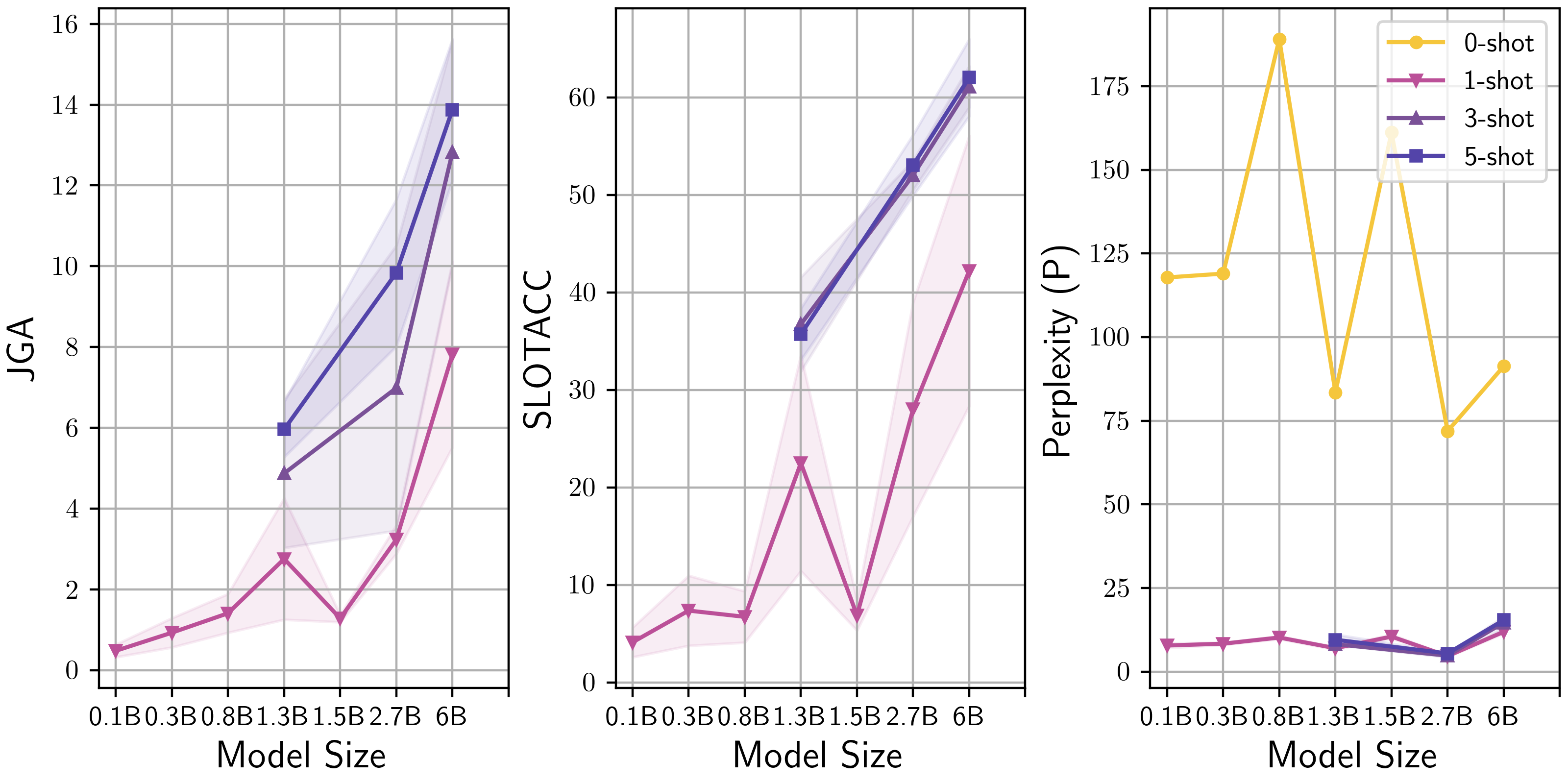}
    \caption{Model size vs metrics in the MWoZ-Single dataset.}
    \label{fig:mwoz_parse_size}
\end{figure}

\begin{figure}[t]
    \centering
    \includegraphics[width=\linewidth]{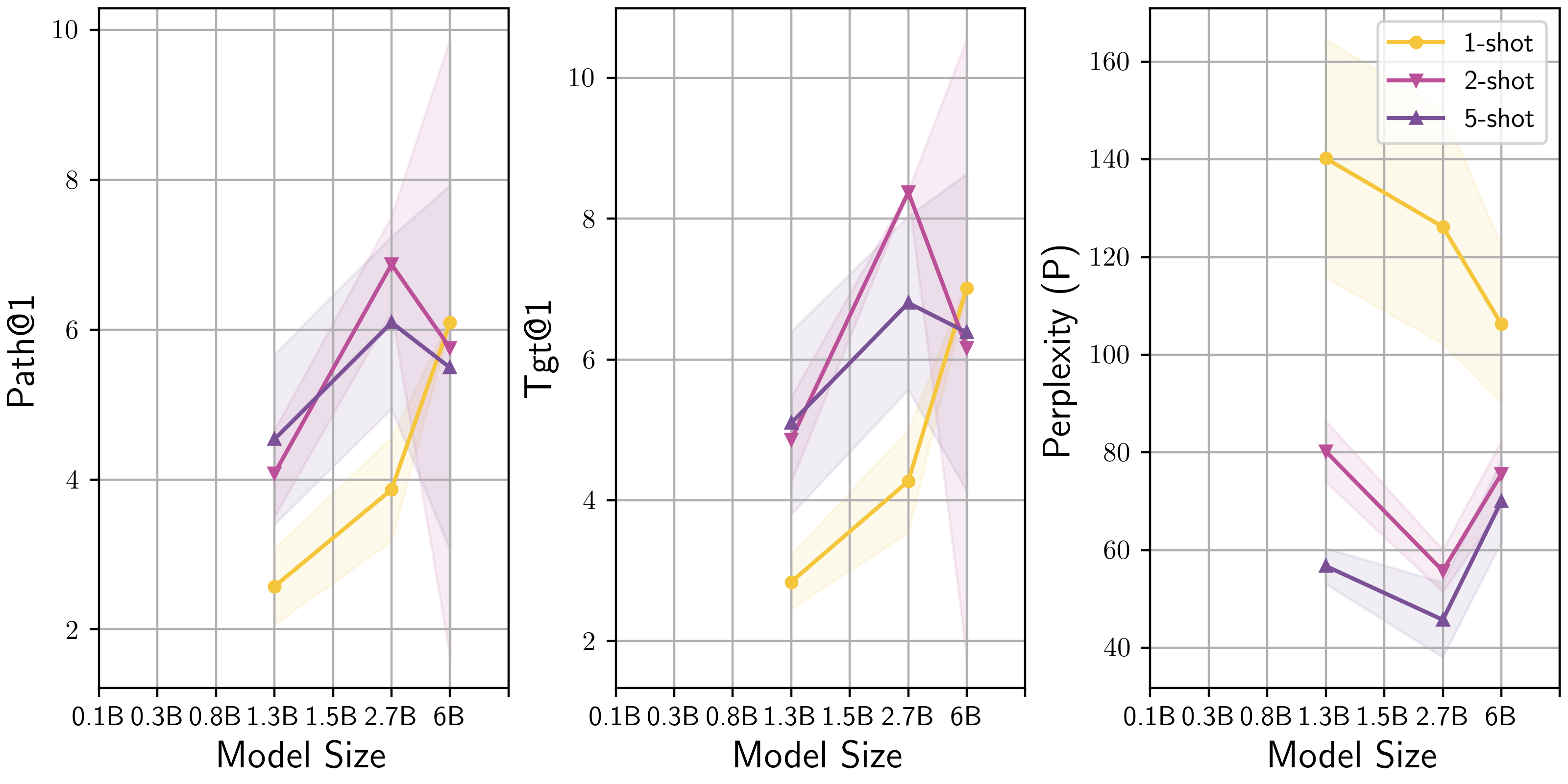}
    \caption{Model size vs metrics in the DialKG-Parse dataset.}
    \label{fig:dialkg_parse_size}
\end{figure}

\begin{figure}
\includegraphics{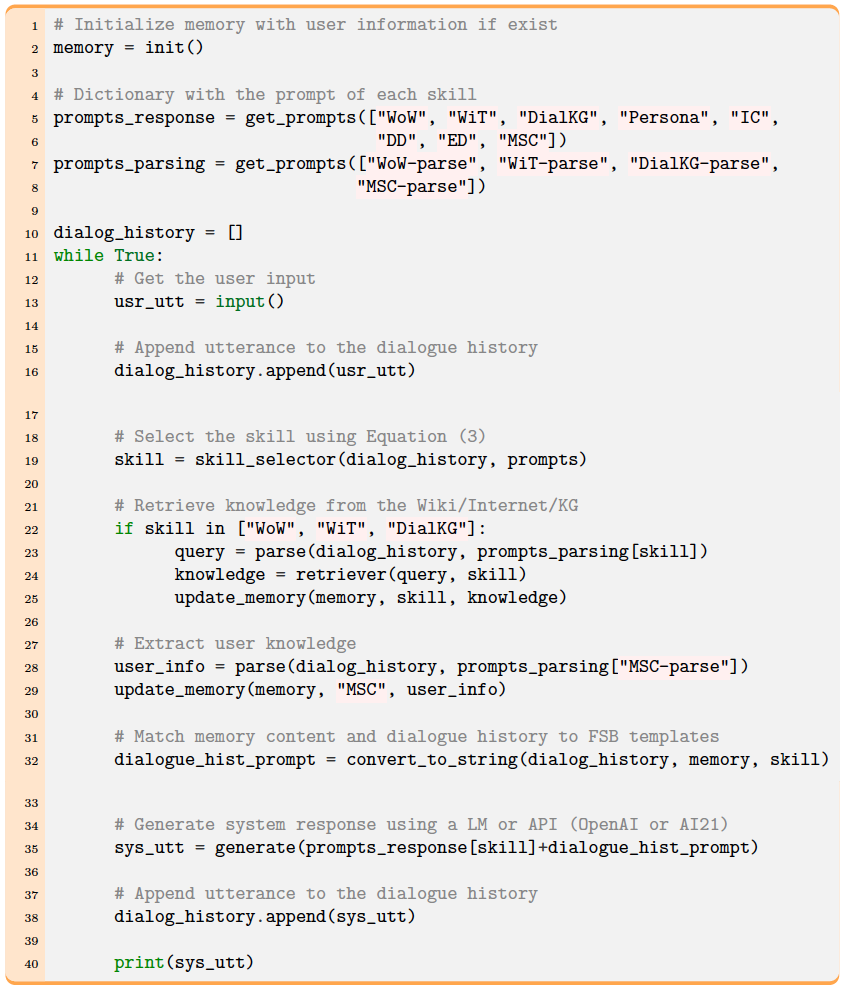}


      
      
      
      
      

      
      
\caption{Pseudo code for the FSB interaction algorithm. }\label{sec:FSB}
\end{figure}

\begin{figure}[t]
\includegraphics{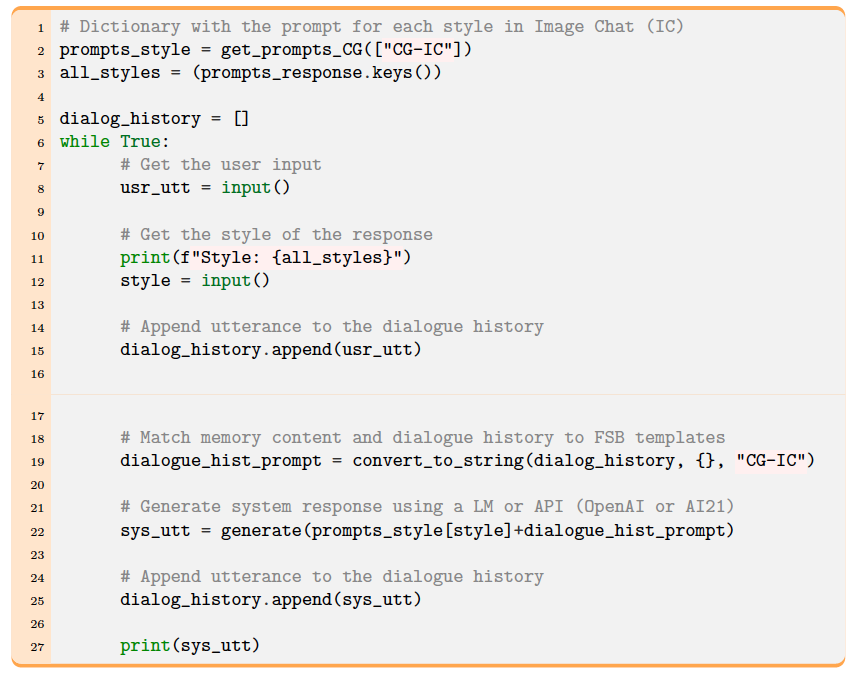}

      
      


      
      
\caption{Pseudo code for the controlled style generation.}\label{sec:FSB_cg}
\end{figure}

\end{document}